\documentclass{article} 
\usepackage{iclr2027_conference,times}

\usepackage{amsmath,amsfonts,bm}

\def\eqref#1{equation~\ref{#1}}

\def\1{\bm{1}}

\DeclareMathAlphabet{\mathsfit}{\encodingdefault}{\sfdefault}{m}{sl}
\SetMathAlphabet{\mathsfit}{bold}{\encodingdefault}{\sfdefault}{bx}{n}

\usepackage{url}
\usepackage{booktabs}
\usepackage[table]{xcolor}
\usepackage{graphicx}
\usepackage{amssymb}
\usepackage{xcolor}
\usepackage{colortbl}  
\usepackage{graphicx}  
\usepackage{booktabs}  
\usepackage{array}
\usepackage{multirow}
\usepackage{pifont}    
\usepackage{url}
\usepackage{tabularx}
\usepackage{enumitem}
\usepackage{mdframed}
\usepackage{hyperref}

\definecolor{questionBackground}{RGB}{255,244,229}
\definecolor{lightBlue}{RGB}{226,243,253}

\makeatletter
\newcommand{\theorylabel}[2]{\begingroup\phantomsection\def\@currentlabel{#2}\label{#1}\endgroup}
\makeatother
\newcommand{\theoryeqref}[1]{\hyperref[#1]{\textup{(\ref*{#1})}}}

\title{Random Is Almost Enough? Stable and Low-Cost Benchmark Compression via BestRand}

\author{
George Ling$^{1}$,
Shanshan Zhong$^{2}$,
Lijin Yang$^{1}$,
Hao Yang$^{1}$,
Nanxiang Li$^{1}$,
Zhongzhan Huang$^{1,\dagger}$ \\[6pt]
$^{1}$Bosch Research \quad
$^{2}$Carnegie Mellon University \quad
$^{\dagger}$Correspondence to: hru4sgh@bosch.com
}

\iclrfinalcopy

\begin{document}

\maketitle
\lhead{}

\addtocontents{toc}{\protect\setcounter{tocdepth}{-1}}

\begin{abstract}

Benchmark compression reduces the often prohibitive time and cost of model evaluation by evaluating models on a small benchmark subset that closely matches full-benchmark results.
To quantify estimation error, we derive an accurate analytical approximation to the error distribution when estimating full-benchmark scores from a subset, showing that most subsets yield small errors, while a few produce much larger errors, resulting in a long-tailed distribution.
This means random sampling alone can achieve surprisingly strong performance with high probability, but remains unstable due to rare large errors.
We further show that this instability is inherent to modern foundation models and persists even as more models are evaluated.
Based on these findings, we introduce BestRand, a stable and low-cost benchmark compression method that preserves the efficiency of random sampling while substantially reducing its high-error tail.
We validate BestRand on 70+ text, multimodal, and agent benchmarks, and further extend it to VLA evaluation.
Compared with competitive methods, BestRand reduces high-error risk by 50.2\% and estimation error by 16.4\% on average, while requiring only 51.3\% of the construction cost.
It further reduces OOD estimation error by 21.8\% on average across score-range, model-family, and temporal shifts.
Together, our theory provides practical guidance for understanding and predicting the accuracy and reliability of benchmark compression, while BestRand enables stable and accurate compression at low cost.
\looseness=-1
The code and data are publicly available on \href{https://github.com/George-Ling3/BestRand}{GitHub}.

\end{abstract}

\section{Introduction}
Evaluating large language models (LLMs) and vision-language models (VLMs) on benchmarks can require substantial time and compute \citep{polo2024tinybenchmarks}, as these benchmarks often contain many test examples and may also include redundant ones \citep{zhang2025redundancy}.
Benchmark compression reduces this burden by evaluating models on a small subset to approximate full-benchmark results \citep{vivek2024anchor}. 
Yet the accuracy and reliability of this approximation are not well characterized.  

\begin{figure}[t]\centering
\includegraphics[width=0.99\columnwidth]{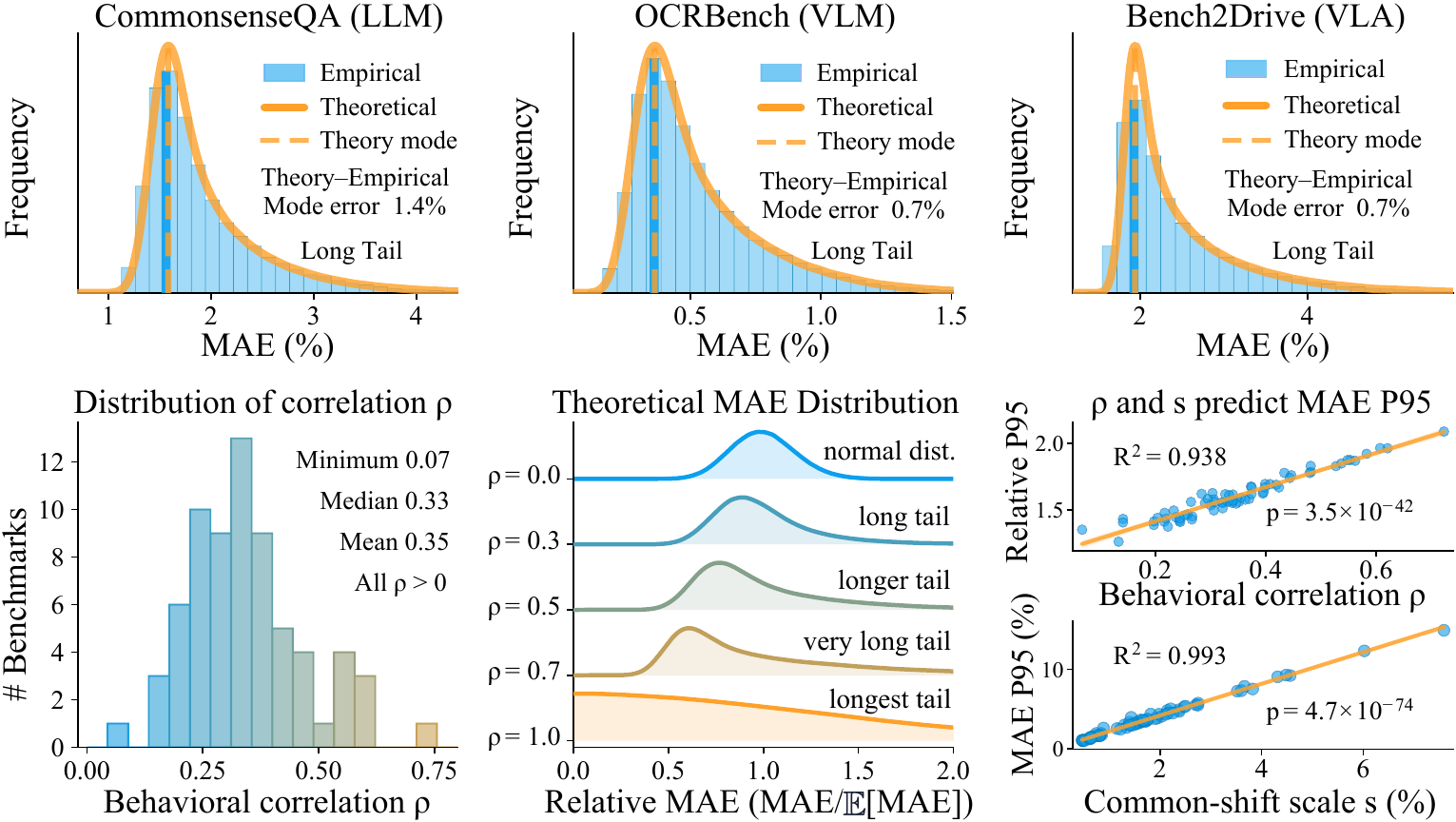}
\vspace{-5pt}
\caption{Empirical results are consistent with our theory. Top: empirical MAE distributions and modes closely match theoretical results on representative LLM, VLM, and VLA benchmarks. Bottom: $\rho$ is generally positive; higher $\rho$ theoretically yields longer tails, while $\rho$ and $s$ strongly predict empirical MAE P95, which captures long-tail risk. Additional validation is provided in Appendix~\ref{app:proofs}.}
\vspace{-5pt}
\label{fig:zzzz}\end{figure}

To quantify estimation error, we derive an accurate analytical approximation to the error distribution when using a subset to estimate full-benchmark scores \citep{truong2025reliable}. 
Our theoretical and empirical results show that most subsets yield small errors, while a few produce much larger errors, resulting in a long-tailed distribution.
This means random sampling alone can achieve surprisingly strong performance with high probability \citep{saranathan2025sublime}, but remains unstable due to rare large errors \citep{huang2025minilongbench}.
We further explain this high-probability effectiveness and show that the instability arises from the similarity among LLMs and VLMs and persists even as more models are evaluated.
Beyond explaining these phenomena, our theory quantifies the relationship between subset size and estimation error \citep{zhang2026benchmark}, guiding the balance between evaluation accuracy and cost \citep{perlitz2024efficient}.
Across all theoretical results, numerical experiments consistently agree with our theory.
Figure~\ref{fig:zzzz} and Appendix~\ref{app:proofs} provide extensive numerical validation.

Motivated by our theory, we introduce BestRand, a stable and low-cost benchmark compression method that preserves the efficiency of random sampling while substantially reducing its high-error tail.
BestRand generates multiple random subsets and uses a few probe models to search for the best candidate.
Our theory identifies a mechanism by which probe-guided selection can reduce common-shift variability, motivating BestRand's design. Experiments demonstrate its improved MAE stability.
We validate BestRand on 70+ text, multimodal, and agent benchmarks, and further extend it to VLA evaluation. 
Against competitive methods, BestRand reduces high-error risk by 50.2\% and estimation error by 16.4\% on average, while requiring only 51.3\% of the construction cost.
It further reduces OOD estimation error by 21.8\% on average across score-range, model-family, and temporal shifts.
Together, our theory provides a framework for understanding and guiding benchmark compression, while BestRand turns these insights into a stable and low-cost method.

We discuss related works in Appendix \ref{app:related-work} and summarize our contributions as follows:

\begin{itemize}[leftmargin=1.2em, labelsep=0.4em, itemsep=0.2em, topsep=0.2em]
    \item \textbf{Theory.} We derive an accurate analytical approximation to the error distribution for benchmark compression, explaining why random sampling is effective with high probability yet remains unstable. We further identify inherent similarity among foundation models as a primary source of this instability. Extensive numerical experiments are highly consistent with our theory.

    \item \textbf{Method.} We propose BestRand, a stable and low-cost benchmark compression method that improves random sampling through lightweight probe-guided search, with theoretical support for improved stability. We validate BestRand on 70+ text, multimodal, and agent benchmarks, as well as VLA evaluation, showing strong accuracy, stability, and OOD generalization at low cost.
\end{itemize}

\vspace{-6pt}

\section{Theory-inspired Method: BestRand}\label{sec:theory-method}
In this section, we combine theoretical analysis with empirical evidence to explain why random sampling often achieves strong compression performance yet remains unstable. Building on these insights, we introduce BestRand to mitigate this instability while preserving the efficiency.

\noindent\textbf{Preliminaries}.
Let $X\in[0,1]^{N\times M}$ be a fixed matrix for $N$ items and $M$ models, whose entry $X_{s,m}$ is the score of model $m$ on item $s$.
All randomness in our sampling analysis comes from drawing a subset $\mathcal S$ of $k$ items uniformly without replacement, where $1\leq k\leq N$. The full-benchmark score, subset estimate, and error of model $m$ are
\begin{equation}
\mu_m=(1/N)\sum\nolimits_{s=1}^{N}X_{s,m},\qquad
\hat\mu_m=(1/k)\sum\nolimits_{s\in\mathcal S}X_{s,m},\qquad
E_m=\hat\mu_m-\mu_m.
\end{equation}
Our primary quantity is the mean absolute error $\mathrm{MAE}=M^{-1}\sum\nolimits_{m=1}^{M}|E_m|$ of the error vector $E=(E_1,\dots,E_M)^\top$.
Subscripts $\mathcal S$ and $N$ indicate subset and full-benchmark averages.
We write $\mathrm{Var}$ for sampling variance, $\mathrm{Var}_N$ for empirical variance over items with denominator $N$, and $\Phi$ for the standard normal cumulative distribution function.

Define item difficulty as $g_s=M^{-1}\sum\nolimits_{m=1}^{M}X_{s,m}$, with larger scores indicating easier items.
Its full-benchmark mean and variance are $\bar g_N=N^{-1}\sum\nolimits_{s=1}^{N}g_s$ and $\mathrm{Var}_N(g)=N^{-1}\sum\nolimits_{s=1}^{N}(g_s-\bar g_N)^2$.
The common shift $\bar E=M^{-1}\sum\nolimits_{m=1}^{M}E_m=\bar g_{\mathcal S}-\bar g_N$, where $\bar g_{\mathcal S}=k^{-1}\sum\nolimits_{s\in\mathcal S}g_s$, measures whether the subset is easier or harder on average.
Its sampling standard deviation defines our central scale,
\begin{equation}
s:=\sqrt{\mathrm{Var}(\bar E)}=c(k)\sqrt{\mathrm{Var}_N(g)},\qquad
c(k)=\sqrt{(N-k)/[k(N-1)]}.
\end{equation}
The finite-population factor $c(k)$ grows as fewer items are retained and satisfies $c(N)=0$.

\noindent\theorylabel{assump:1}{1}\textbf{Assumption 1: }
Consider non-degenerate models whose scores vary across test items. For binary scores, this excludes models that answer all items incorrectly or all items correctly. Define the population score covariance matrix $C\in\mathbb R^{M\times M}$ and its relation to sampling errors by
\begin{equation}
C_{ml}=\frac{1}{N}\sum\nolimits_{s=1}^{N}(X_{s,m}-\mu_m)(X_{s,l}-\mu_l),\qquad
\mathrm{Cov}(E)=c(k)^2C.
\end{equation}
Assume $C_{mm}=\sigma^2>0$ and $C_{ml}=\rho\sigma^2$ for $m\ne l$, with $\rho\in[0,1)$.
Then the common-shift scale simplifies to $s=c(k)\,\sigma\sqrt{\rho+(1-\rho)/M}$.
Empirically, $\sigma^2$ and $\rho$ are read directly from $C$ as the mean diagonal entry and the mean normalized off-diagonal entry, $\sigma^2=M^{-1}\sum\nolimits_m C_{mm}$ and $\rho=[M(M-1)]^{-1}\sum\nolimits_{m\ne l}C_{ml}/\sigma^2$, so $\rho$ is the average pairwise correlation of the model errors.

\noindent\theorylabel{assump:2}{2}\textbf{Assumption 2: }
Let $x_s=(x_{s,1},\dots,x_{s,p})^\top\in\mathbb R^p$ contain the scores of $p$ probe models on item $s$, with means $\bar x_{\mathcal S}=k^{-1}\sum\nolimits_{s\in\mathcal S}x_s$ and $\bar x_N=N^{-1}\sum\nolimits_{s=1}^{N}x_s$.
Regress difficulty on probe scores by least squares with intercept $\alpha\in\mathbb R$, coefficients $\beta\in\mathbb R^p$, and residual $e_s$.
We assume $\mathrm{Var}_N(g)>0$ and positive explained variance,
\begin{equation}
\label{eq:probe-regression}
g_s=\alpha+\beta^\top x_s+e_s,\qquad
R^2:=\mathrm{Var}_N(\beta^\top x)/\mathrm{Var}_N(g)>0.
\end{equation}
Derivations are provided in Appendix~\ref{app:proofs}, and empirical checks across numerous benchmarks in Appendix~\ref{app:assumption-validation} support these assumptions as useful approximations rather than exact descriptions of the data.
The resulting theoretical predictions closely match empirical observations (e.g., Figure~\ref{fig:zzzz}), further supporting the value of these simplifications for tractable analysis.

\subsection{Why Random Sampling is almost enough for compression but unstable?}\label{sec:random-almost-enough}

In this Section, Theorem~\ref{thm:1} fixes the shape of the error distribution and its tail, and show why the random sampling is unstable. Theorem~\ref{thm:2} helps explain why random sampling may be almost enough in many cases: for large $M$, the mode of the MAE distribution lies close to the lower endpoint of its limiting distribution. Proofs are in Appendix~\ref{app:proofs}.

\medskip
\noindent\theorylabel{thm:1}{1}\textbf{Theorem 1.}
Under Assumption~\ref{assump:1}, when the effective sample size $\min\{k,N-k\}$ is large, the error vector is approximately zero-mean multivariate normal:
\begin{equation}
E\ \overset{\cdot}{\sim}\ \mathcal N\big(0,\,c(k)^2C\big),\qquad
\mathrm{MAE}=\frac{1}{M}\sum\nolimits_{m=1}^{M}|E_m|.
\end{equation}
Moreover, $\mathrm{MAE}\ge|\bar E|$ holds for every sampled subset, and the common shift $\bar E=M^{-1}\sum\nolimits_m E_m$ is approximately $\mathcal N(0,s^2)$.
Hence, for every threshold $t>0$,
\begin{equation}
\Pr(\mathrm{MAE}>t)\ \ge\ \Pr(|\bar E|>t)\ \approx\ 2\,\Phi\!\Big(-\tfrac{t}{s}\Big),
\end{equation}
where $s=\sqrt{\mathrm{Var}(\bar E)}$ is the common-shift scale defined above.

\medskip
\noindent\textbf{Why random sampling is unstable.}
Instability means that a single random draw occasionally lands far from the true scores, which is exactly the right tail of the MAE distribution. Theorem~\ref{thm:1} controls this tail through its lower bound $\Pr(\mathrm{MAE}>t)\ge 2\Phi(-t/s)$, whose only free quantity is the scale $s$. At any fixed threshold $t$, a larger $s$ pushes $-t/s$ toward zero, and because $\Phi$ is increasing it raises the probability of a large error. A larger $s$ therefore raises this lower bound on high-error risk, so it remains to ask what makes $s$ large.
Under Assumption~\ref{assump:1}, $s=c(k)\sigma\sqrt{\rho+(1-\rho)/M}$. For $M>1$, the term under the square root increases with the behavioral correlation $\rho$, since its derivative is $1-1/M>0$. Stronger correlation therefore increases $s$ and raises the lower bound on the probability of large errors. Increasing the number of evaluated models reduces the contribution $(1-\rho)/M$, but cannot eliminate the correlation-dependent component: for fixed $k<N$ and $\rho>0$, $s\to c(k)\sigma\sqrt{\rho}>0$ as $M\to\infty$. Thus, expanding the model panel alone cannot remove this source of instability.
Numerical experiments across diverse benchmarks in Figure~\ref{fig:zzzz} provide empirical support for this analysis based on Theorem~\ref{thm:1}.

For foundation models, such behavioral correlation is closely tied to how the models are built and trained. Language, vision-language, and action models often share similar or identical backbones and overlapping training corpora, while knowledge distillation can further align their behavior. These shared design and training choices can lead models to succeed and fail on the same items, sustaining a sizable $\rho$. In this sense, random sampling instability is rooted in the shared characteristics of foundation models, rather than merely in the finite number of models evaluated. The non-vanishing limit above describes a consequence of this intrinsic correlation: evaluating more similarly constructed models does not average away their shared errors.

\medskip
\noindent\theorylabel{thm:2}{2}\textbf{Theorem 2.}
Under Assumption~\ref{assump:1} and Theorem~\ref{thm:1} with fixed $c(k)>0$, $\sigma>0$, and $0<\rho<1$, the lower endpoint of the limiting $\mathrm{MAE}$ distribution as $M\to\infty$ is $y_0=\sqrt{2/\pi}\,c(k)\,\sigma\sqrt{1-\rho}$.
This is not a lower bound for finite $M$, where the MAE distribution can have probability mass below $y_0$.
 The mode $\hat y(M)$ of the $\mathrm{MAE}$ distribution approaches $y_0$ at the universal rate
\begin{equation}
\frac{\hat y(M)}{y_0}-1=\frac{C_0}{\sqrt M}+O\!\Big(\tfrac1M\Big),\quad
C_0=\sqrt{\tfrac\pi2-1}\,\arg\max_{u>0}\int_0^{\infty}\! v^{-1/2}\varphi(u-v)\,\mathrm dv\approx0.578,
\end{equation}
independent of $c(k),\sigma$ and $\rho$, with $\varphi$ the standard normal density.

\medskip
\noindent\textbf{Why random sampling is almost enough.}
Theorem~\ref{thm:2} gives an asymptotic description of the low-error region: for large $M$, the mode lies near the lower endpoint $y_0$ of the limiting distribution, with a leading relative gap of $C_0/\sqrt M$. This prediction is consistent with the empirical observation in Figure~\ref{fig:zzzz} and more experiments are shown in Appendix~\ref{app:thm2}, where the mode lies close to the lower end of the observed MAE distribution. Together, the asymptotic analysis and the observed concentration at small errors help explain why random sampling is often already effective. The long tail nevertheless leaves a small but non-negligible fraction of high-error draws, motivating BestRand to improve stability while preserving the low cost of random sampling. 


\subsection{How to build more stable and low-cost benchmark compression methods.}
Theorem~\ref{thm:2} leaves a clear target. The asymptotic analysis and empirical observations suggest that random sampling already favors low-error subsets, so a good compressor should preserve its low cost while reducing the risk of high-error draws. BestRand does this by drawing several random candidate subsets and selecting the one that a few cheap probe models judge most representative. It runs the target models on the selected subset alone and reuses the probe scores across all candidates, so its only overhead beyond random sampling is a one-time probe pass over a small pool.

\medskip
\noindent\textbf{The BestRand procedure.}
Given a benchmark of $N$ items, a target size $k$, $p$ probe models, a pool multiplier $n$, and a candidate count $m$, BestRand proceeds in four steps.
\begin{enumerate}[leftmargin=1.6em,labelsep=0.4em,itemsep=0.15em,topsep=0.2em]
\item Draw a pool $P$ of $\min(nk,N)$ items uniformly without replacement and score the $p$ probe models on $P$, recording $x_s\in\mathbb R^p$ for each item $s$.
\item Form the reference $H=|P|^{-1}\sum_{s\in P}x_s$, the average probe score over the pool.
\item Repeat $m$ times: draw a candidate $k$-subset $S_j$ from $P$ uniformly without replacement, and compute its probe mean $h_j=k^{-1}\sum_{s\in S_j}x_s$ and its imbalance $d_j=\|h_j-H\|^2$.
\item Return $S^{*}=S_{j^{*}}$ with $j^{*}=\arg\min_{j\le m}d_j$, and evaluate every target model on $S^{*}$ alone.
\end{enumerate}
The single probe pass costs $p\times|P|$ model-item evaluations, and the pool reference $H$ replaces the full-set probe mean so the probes need not run on the whole benchmark when $nk<N$. The $m$ candidate draws in step~3 reuse the cached probe scores and add no model calls, computing only averages of cached vectors, so they run in parallel as a single batched tensor operation on a GPU and make the search essentially instant. The selected subset $S^{*}$ is reused across all target models.

\medskip
\noindent\theorylabel{thm:3}{3}\textbf{Theorem 3.}
Under Assumptions~\ref{assump:1} and~\ref{assump:2}, least-squares orthogonality splits the common shift into a probe-explained part and a residual part,
\begin{equation}
\bar E=\bar E_\parallel+\bar E_\perp,\qquad
\bar E_\parallel=\beta^\top(\bar x_{\mathcal S}-\bar x_N),\quad
\bar E_\perp=\bar e_{\mathcal S}=\tfrac1k\!\sum_{s\in\mathcal S}e_s,
\end{equation}
which are uncorrelated with variances $\mathrm{Var}(\bar E_\parallel)=R^2 s^2$ and $\mathrm{Var}(\bar E_\perp)=(1-R^2)s^2$, so $R^2$ is the fraction of the common-shift variance that the probes explain. Draw $m$ candidates independently, each uniformly among the $k$-subsets of the full benchmark, and select the candidate with the smallest full-reference probe imbalance, breaking ties by a fixed index rule:
\begin{equation}
S^{*}=\arg\min_{j\le m}\big\|\bar x_{S_j}-\bar x_N\big\|,\qquad
D^{*}=\big\|\bar x_{S^{*}}-\bar x_N\big\|,
\end{equation}
Within a joint Gaussian model of probe and residual imbalances, the selected common-shift variance satisfies
\begin{equation}
s_{S^{*}}^2=\mathrm{Var}\big(\bar E_\parallel(S^{*})\big)+(1-R^2)\,s^2,\qquad
0\le\mathrm{Var}\big(\bar E_\parallel(S^{*})\big)\le\|\beta\|^2\,\mathbb E\big[D^{*2}\big],
\end{equation}
Thus, if $\mathbb E[D^{*2}]\to0$, the common-shift scale converges to $\sqrt{1-R^2}\,s$ within this model.
\begin{figure*}[t]
\centering
\includegraphics[width=\textwidth]{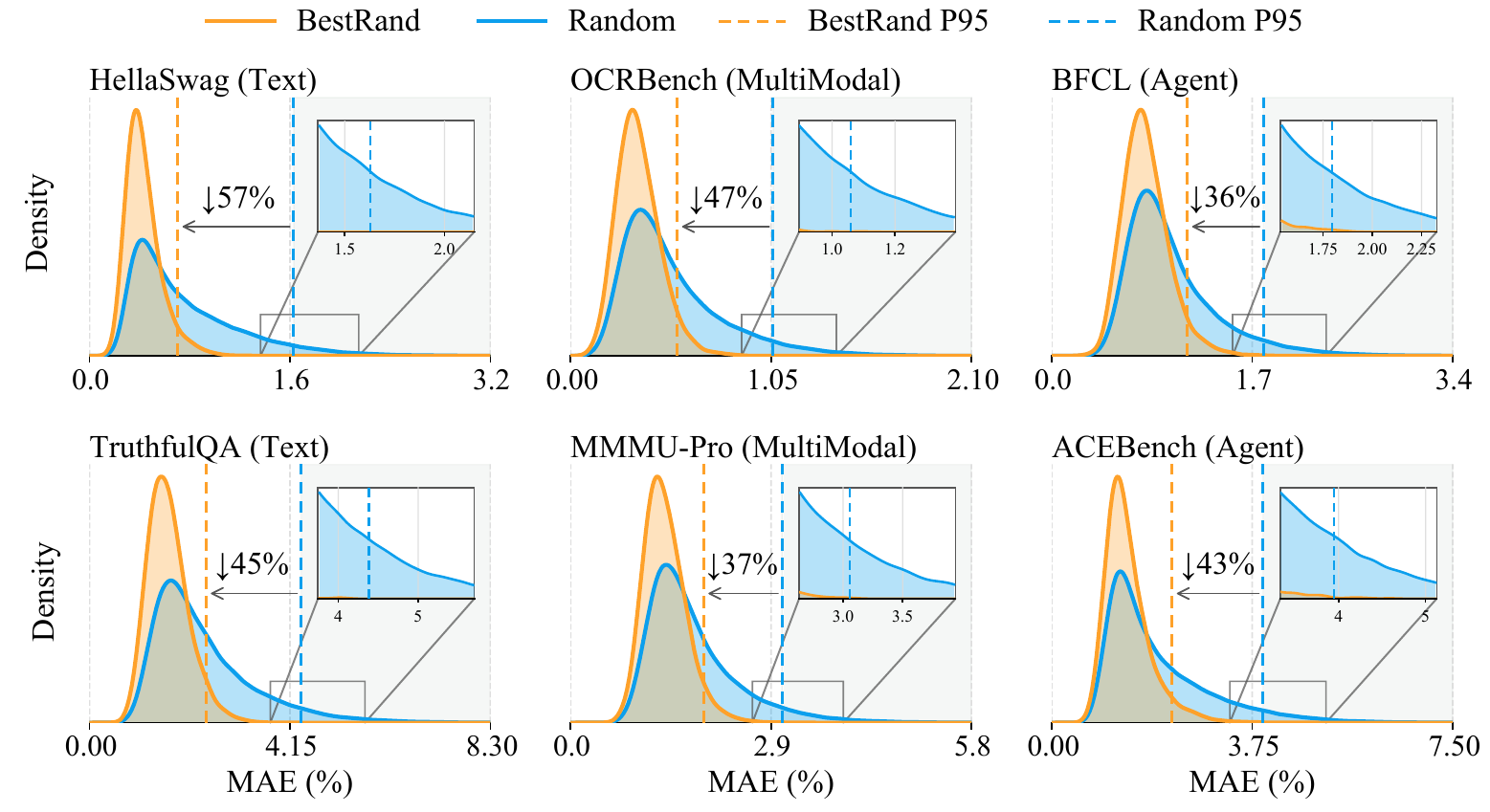}
\vspace{-20pt}
\caption{Error distributions of Random and BestRand on six representative benchmarks at 20\% retention. Random exhibits a pronounced right tail, whereas BestRand concentrates errors near zero and suppresses high-error outcomes. Results on additional benchmarks are shown in Figure~\ref{fig:long-tail-distribution-appendix}.}
\label{fig:long-tail-distribution-main}
\end{figure*}

\medskip
\noindent\textbf{Why BestRand is better than random sampling.}
Theorem~\ref{thm:3} \textbf{motivates} BestRand by identifying the probe-explained component of the common shift as a target for selection. By the Cauchy--Schwarz inequality, $|\bar E_\parallel|\le\|\beta\|\,\|\bar x_{\mathcal S}-\bar x_N\|$. Matching the full-benchmark probe mean therefore controls this component without knowing $\beta$ or evaluating the target models, motivating BestRand's probe-based selection criterion.
To implement this criterion at low cost, BestRand uses the pool mean $H$ to approximate $\bar x_N$ and selects the candidate that best matches $H$. The connection follows from $\|\bar x_{S^{*}}-\bar x_N\|\le\|\bar x_{S^{*}}-H\|+\|H-\bar x_N\|$. The first term is the imbalance minimized by the candidate search, while the second is the reference error controlled by the pool size. Choosing a sufficiently large $n$ makes $H$ close to $\bar x_N$ in mean square, with exact equality when the pool covers the full benchmark. Increasing $m$ then improves or preserves the best match to this reference without additional model evaluations. Thus, $n$ controls reference accuracy and $m$ controls search quality.

This design aims to reduce the common-shift scale. Within the selection model of Theorem~\ref{thm:3}, the probe-explained variance is bounded by $\|\beta\|^2\mathbb E[D^{*2}]$, while the residual variance remains $(1-R^2)s^2$. If $\mathbb E[D^{*2}]\to0$ and $s>0$, then
\begin{equation}
s_{S^{*}}^2 =\mathrm{Var}\big(\bar E_\parallel(S^{*})\big)+(1-R^2)\,s^2\ \longrightarrow\ (1-R^2)\,s^2 <\ s^2.
\end{equation}
A larger explained fraction $R^2$ yields a greater potential reduction in the shared source of high-error risk identified by Theorem~\ref{thm:1}. This limit describes the idealized benefit of probe matching, rather than a guaranteed contraction for finite-pool selection, where reference error and nonzero attainable imbalance remain.
Our numerical results are consistent with this analysis. Figure~\ref{fig:long-tail-distribution-main} shows that BestRand concentrates MAE near zero and suppresses the upper tail relative to random sampling across benchmarks, supporting the practical value of probe-guided selection for improving stability. 

\begin{table*}[t]
\centering
\caption{Compression accuracy across 70 benchmarks at six retention ratios. Results are grouped into 44 text, 16 multimodal, and 10 agent benchmarks. Each entry reports average MAE (\%) and the standard deviation across benchmarks. The final column averages all benchmarks and retention ratios. Lower is better. Best mean values are \textbf{bold} and second-best mean values are \underline{underlined}.}
\vspace{4pt}
\label{tab:category-main-results}
\footnotesize
\setlength{\tabcolsep}{2.8pt}
\renewcommand{\arraystretch}{1.08}
\newcommand{\std}[1]{\,{\scriptsize\color{black!70}#1}}
\newcommand{\catrow}[2]{\multicolumn{8}{l}{\textcolor{blue!45!black}{\textbf{#1}}\enspace\textcolor{black!70}{#2 benchmarks}}}
\resizebox{\textwidth}{!}{%
\begin{tabular}{lrrrrrrr}
\toprule
& \multicolumn{7}{c}{\textbf{Average MAE (\%) $\downarrow$}\enspace {\scriptsize mean $\pm$ std.}} \\
\cmidrule(lr){2-8}
\rowcolor{lightBlue}\textbf{Method} & \textbf{2\%} & \textbf{5\%} & \textbf{10\%} & \textbf{15\%} & \textbf{20\%} & \textbf{50\%} & \textbf{Mean} \\
\midrule
\catrow{Text}{44} \\
SparseEval (Linear) & 19.88\std{$\pm$9.48} & 21.08\std{$\pm$6.99} & 22.25\std{$\pm$7.72} & 21.18\std{$\pm$10.10} & 21.44\std{$\pm$9.60} & 20.04\std{$\pm$7.95} & 20.98\std{$\pm$8.67} \\
SparseEval (MLP) & 12.53\std{$\pm$5.85} & 11.94\std{$\pm$5.16} & 11.70\std{$\pm$5.38} & 11.49\std{$\pm$5.32} & 11.74\std{$\pm$5.51} & 11.80\std{$\pm$5.80} & 11.87\std{$\pm$5.46} \\
Anchor Points & 7.68\std{$\pm$3.24} & 7.23\std{$\pm$2.66} & 6.78\std{$\pm$2.73} & 6.32\std{$\pm$2.66} & 6.41\std{$\pm$2.73} & 4.72\std{$\pm$2.58} & 6.52\std{$\pm$2.90} \\
TinyBenchmarks & 7.45\std{$\pm$4.49} & 4.94\std{$\pm$3.34} & 3.26\std{$\pm$2.22} & 2.56\std{$\pm$1.82} & 2.26\std{$\pm$1.70} & 1.45\std{$\pm$1.55} & 3.65\std{$\pm$3.37} \\
ZipBench & \underline{6.48}\std{$\pm$3.97} & \underline{4.35}\std{$\pm$2.65} & \underline{3.03}\std{$\pm$1.97} & \underline{2.40}\std{$\pm$1.48} & \underline{2.02}\std{$\pm$1.20} & 1.10\std{$\pm$0.68} & \underline{3.23}\std{$\pm$2.85} \\
Random & 7.37\std{$\pm$5.18} & 4.60\std{$\pm$3.20} & 3.17\std{$\pm$2.19} & 2.51\std{$\pm$1.73} & 2.11\std{$\pm$1.45} & \underline{1.05}\std{$\pm$0.72} & 3.47\std{$\pm$3.46} \\
\rowcolor{lightBlue}\textbf{BestRand (ours)} & \textbf{6.27}\std{$\pm$4.08} & \textbf{3.92}\std{$\pm$2.49} & \textbf{2.67}\std{$\pm$1.69} & \textbf{2.09}\std{$\pm$1.31} & \textbf{1.73}\std{$\pm$1.07} & \textbf{0.85}\std{$\pm$0.52} & \textbf{2.92}\std{$\pm$2.80} \\
\midrule
\catrow{Multimodal}{16} \\
SparseEval (Linear) & 22.59\std{$\pm$10.58} & 21.92\std{$\pm$6.65} & 25.21\std{$\pm$14.38} & 25.97\std{$\pm$10.38} & 22.76\std{$\pm$6.94} & 21.02\std{$\pm$8.38} & 23.25\std{$\pm$9.81} \\
SparseEval (MLP) & 13.61\std{$\pm$6.37} & 14.59\std{$\pm$8.14} & 14.00\std{$\pm$7.11} & 14.12\std{$\pm$7.52} & 12.68\std{$\pm$7.05} & 13.00\std{$\pm$6.70} & 13.67\std{$\pm$7.01} \\
Anchor Points & 8.18\std{$\pm$4.13} & 6.83\std{$\pm$2.66} & 6.66\std{$\pm$2.23} & 5.84\std{$\pm$2.15} & 6.61\std{$\pm$2.46} & 5.07\std{$\pm$2.10} & 6.53\std{$\pm$2.81} \\
TinyBenchmarks & 7.58\std{$\pm$3.31} & 4.21\std{$\pm$2.11} & \underline{2.88}\std{$\pm$1.54} & \underline{2.27}\std{$\pm$1.09} & 2.11\std{$\pm$1.01} & 1.36\std{$\pm$0.62} & 3.40\std{$\pm$2.74} \\
ZipBench & \textbf{6.10}\std{$\pm$2.89} & \underline{4.15}\std{$\pm$2.24} & 3.28\std{$\pm$1.96} & 2.54\std{$\pm$1.66} & 1.99\std{$\pm$1.09} & 1.09\std{$\pm$0.54} & \underline{3.19}\std{$\pm$2.45} \\
Random & 6.89\std{$\pm$3.47} & 4.30\std{$\pm$2.15} & 2.94\std{$\pm$1.46} & 2.34\std{$\pm$1.16} & \underline{1.96}\std{$\pm$0.98} & \underline{0.98}\std{$\pm$0.49} & 3.24\std{$\pm$2.66} \\
\rowcolor{lightBlue}\textbf{BestRand (ours)} & \underline{6.16}\std{$\pm$3.21} & \textbf{3.83}\std{$\pm$2.00} & \textbf{2.61}\std{$\pm$1.36} & \textbf{2.06}\std{$\pm$1.08} & \textbf{1.71}\std{$\pm$0.90} & \textbf{0.84}\std{$\pm$0.45} & \textbf{2.87}\std{$\pm$2.43} \\
\midrule
\catrow{Agent}{10} \\
SparseEval (Linear) & 20.90\std{$\pm$6.89} & 25.68\std{$\pm$6.57} & 24.74\std{$\pm$6.99} & 22.33\std{$\pm$7.91} & 19.77\std{$\pm$4.42} & 22.79\std{$\pm$6.04} & 22.70\std{$\pm$6.60} \\
SparseEval (MLP) & 12.32\std{$\pm$3.41} & 12.76\std{$\pm$2.74} & 12.65\std{$\pm$3.37} & 11.65\std{$\pm$2.72} & 10.66\std{$\pm$2.93} & 10.40\std{$\pm$2.55} & 11.74\std{$\pm$2.99} \\
Anchor Points & 12.35\std{$\pm$5.84} & 9.56\std{$\pm$4.16} & 7.84\std{$\pm$2.65} & 7.25\std{$\pm$2.50} & 7.19\std{$\pm$2.35} & 5.34\std{$\pm$2.06} & 8.26\std{$\pm$4.04} \\
TinyBenchmarks & 13.10\std{$\pm$8.57} & \underline{7.91}\std{$\pm$5.73} & 5.96\std{$\pm$4.37} & 4.58\std{$\pm$2.80} & 3.80\std{$\pm$1.98} & 2.27\std{$\pm$1.28} & 6.27\std{$\pm$5.81} \\
ZipBench & \underline{12.17}\std{$\pm$7.83} & 8.07\std{$\pm$5.68} & 5.53\std{$\pm$3.82} & \underline{4.31}\std{$\pm$2.98} & 3.90\std{$\pm$2.74} & 2.08\std{$\pm$1.38} & \underline{6.01}\std{$\pm$5.51} \\
Random & 13.32\std{$\pm$8.77} & 8.22\std{$\pm$5.55} & \underline{5.48}\std{$\pm$3.56} & 4.38\std{$\pm$2.88} & \underline{3.65}\std{$\pm$2.36} & \underline{1.82}\std{$\pm$1.18} & 6.14\std{$\pm$5.91} \\
\rowcolor{lightBlue}\textbf{BestRand (ours)} & \textbf{11.94}\std{$\pm$8.25} & \textbf{7.30}\std{$\pm$5.14} & \textbf{4.82}\std{$\pm$3.25} & \textbf{3.82}\std{$\pm$2.61} & \textbf{3.14}\std{$\pm$2.13} & \textbf{1.56}\std{$\pm$1.06} & \textbf{5.43}\std{$\pm$5.44} \\
\bottomrule
\end{tabular}
}%
\end{table*}

\begin{figure*}[t]
\centering
\includegraphics[width=\textwidth]{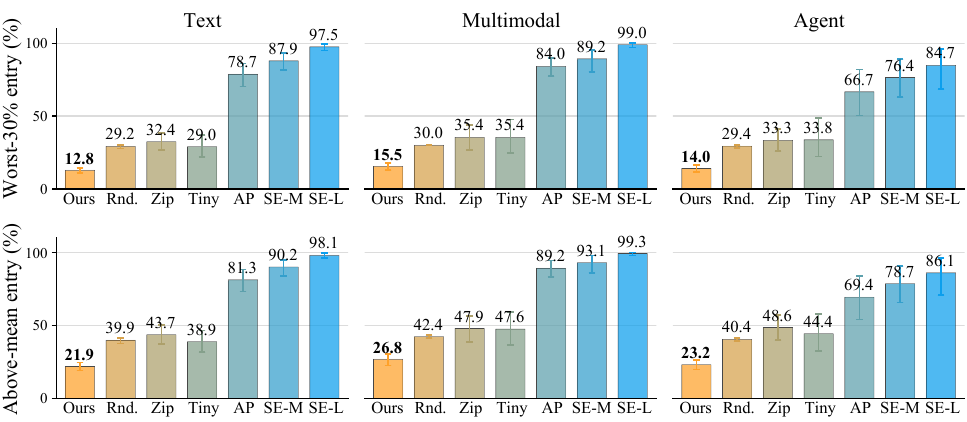}
\vspace{-15pt}
\caption{High-error risk across text, multimodal, and agent benchmarks. Worst-30\% entry and worse-than-average entry use Random's 70th-percentile and mean errors as thresholds, respectively. Whiskers show 95\% bootstrap confidence intervals. BestRand achieves the lowest risk in all settings. Rand, Zip, Tiny, AP, SE-M, and SE-L denote Random, ZipBench, TinyBenchmarks, Anchor Points, SparseEval MLP, and SparseEval Linear, respectively. Detailed results are shown in Figure~\ref{fig:tail-entry-appendix}.}
\label{fig:tail-entry-main}
\vspace{-6pt}
\end{figure*}

\section{Experiments}\label{sec:experiments}

We evaluate BestRand on 70 text, multimodal, and agent benchmarks, focusing on estimation accuracy and stability. We compare average errors across methods and compression budgets, then examine error distributions and high-error risk, with VLA results reported in the Appendix \ref{app:additional-analysis}.
We first introduce the benchmarks, then describe the baselines, evaluation protocol, and metrics.

\paragraph{Benchmarks and budgets.}
Our evaluation spans 70 benchmarks, comprising 44 text, 16 multimodal, and 10 agent benchmarks. These include widely used text benchmarks such as MMLU \citep{hendrycks2020measuring}, GSM8K \citep{cobbe2021training}, and HumanEval \citep{chen2021evaluating}, multimodal benchmarks such as MMMU \citep{yue2024mmmu}, MathVista \citep{lu2024mathvista}, and MME \citep{fu2026mme}, and agent benchmarks such as OSWorld \citep{xie2024osworld} and tau-bench \citep{yao2024tau}. To compare methods comprehensively across compression levels, we evaluate six retention ratios ranging from 2\% to 50\%. Details of all benchmarks are provided in Appendix~\ref{app:benchmark-details}.

\paragraph{Baselines and evaluation protocol.}
We compare against six representative baselines. 
\textbf{Random} uniformly samples a subset and averages its scores. \textbf{SparseEval} \citep{zhang2026sparseeval} uses a linear or MLP predictor to estimate full-benchmark scores from subset scores. \textbf{Anchor Points}, \textbf{TinyBenchmarks}, and \textbf{ZipBench} \citep{vivek2024anchor,polo2024tinybenchmarks, huang2026zipbenchlowcostframeworkcompressing} construct item representations and cluster them to select a representative subset.
All methods use the same model split and item budget. Random and BestRand use 10,000 seeds, while other methods use three. BestRand uses $p=5$, $m=128$, $n=4$, random probe selection, and $L_2$ matching. It probes only the candidate pool and thus incurs a lower construction cost.
Section~\ref{sec:analysis-ablation} further compares four additional baselines. Appendices~\ref{app:data-splits} and~\ref{app:baseline-details} provide details on the data splits and baselines, respectively.

\paragraph{Evaluation metrics.}
We use mean absolute error (MAE) as our primary metric. We also evaluate ranking fidelity using Spearman's rank correlation coefficient (SP). Since rank correlations are generally strong for leading methods in both our experiments and prior work \citep{yauney2026reliable}, we focus on MAE in the main results and report detailed SP results in Appendix~\ref{app:additional-results}, Tables~\ref{tab:representative-sp-text}--\ref{tab:representative-sp-agent}.

We then compare BestRand with existing methods, showing its high accuracy, low cost, and stability.

\paragraph{Main results.}
As shown in Table~\ref{tab:category-main-results}, Random performs strongly on average. BestRand consistently improves on this baseline, reducing both mean MAE and its variation across all settings. Against three competitive baselines, namely Random, ZipBench, and TinyBenchmarks, BestRand reduces MAE by 16.4\% on average, achieving the best overall performance. By probing only a candidate pool, BestRand uses just 8.0\% of data-dependent baselines' construction records at 2\% retention and 51.3\% on average. BestRand therefore achieves the lowest error at roughly half the construction cost. Beyond average performance, we next compare the stability and high-error risk of all methods.

\paragraph{Stability and high-error risk.}
Figure~\ref{fig:long-tail-distribution-main} shows that BestRand concentrates errors near zero and suppresses Random's upper tail, yielding more stable estimates.
Figure~\ref{fig:tail-entry-main} quantifies the risk of falling into the high-error region of Random's long-tailed distribution. We measure the probability that each method's error exceeds Random's 70th percentile or mean, corresponding to the worst 30\% of random subsets and worse-than-average random performance, respectively. Under both thresholds, BestRand achieves the lowest risk across benchmark categories and reduces high-error probability by 50.2\% versus three strong baselines. Additional results are reported in Appendix~\ref{app:additional-results}. Together, these results demonstrate that BestRand enables accurate and stable benchmark compression.

\begin{figure}[t]
\centering
\includegraphics[width=\columnwidth]{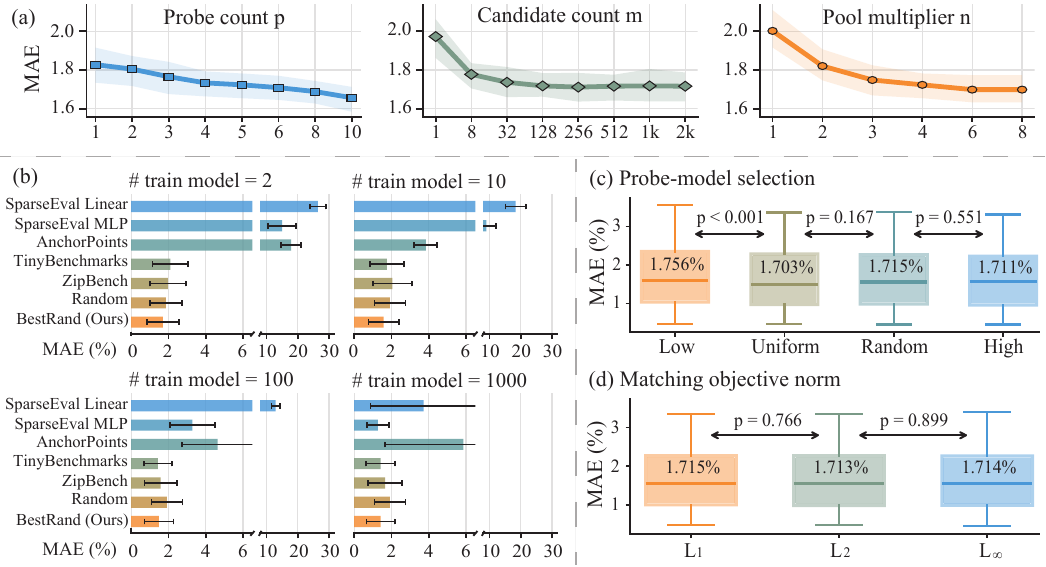}
\vspace{-18pt}
\caption{BestRand ablations at 20\% retention. (a) Effects of the number of probe models, the number of candidate subsets, and the candidate-pool size. (b) Performance comparison with more construction models. (c) Effect of probe-model selection. (d) Effect of the matching norm.}
\label{fig:BestRand-ablation-tradeoffs}
\end{figure}

\section{Analysis and ablation study}
\label{sec:analysis-ablation}

This section analyzes the key factors affecting BestRand and evaluates its effectiveness, efficiency, and robustness across practical settings. Additional analyses are provided in Appendix~\ref{app:additional-analysis}.

\begin{mdframed}[backgroundcolor=questionBackground,linecolor=gray!35,linewidth=0.5pt,innertopmargin=5pt,innerbottommargin=5pt]
\noindent\textbf{(1) Effects of Probe Count, Candidate Count, and Pool Multiplier on Performance.}
\end{mdframed}
\vspace{-5pt}
We first examine how the probe count $p$, candidate count $m$, and pool multiplier $n$ affect compression accuracy. As shown in Figure~\ref{fig:BestRand-ablation-tradeoffs}(a), more probes improve the reliability of candidate comparison, more candidates increase the chance of finding a representative subset, and a larger pool provides a more accurate reference and greater candidate diversity. Performance improves with all three quantities, but the gains from increasing $m$ and $n$ quickly diminish. We therefore use $p=5$, $m=128$, and $n=4$ by default. This setting achieves strong accuracy at low construction cost.

\vspace{3pt}
\begin{mdframed}[backgroundcolor=questionBackground,linecolor=gray!35,linewidth=0.5pt,innertopmargin=5pt,innerbottommargin=5pt]
\noindent\textbf{(2) Does BestRand Remain Competitive as More Construction Models Become Available?}
\end{mdframed}
\vspace{-5pt}
Because our main experiments use only five construction models, we next test whether BestRand remains competitive when more models are available. Figure~\ref{fig:BestRand-ablation-tradeoffs}(b) shows that BestRand performs best with only a few construction models. Learning-based methods improve as they gain more response records, but BestRand remains comparable to the strongest alternatives even with hundreds or thousands of models. Thus, BestRand offers the greatest gains when construction data are limited while retaining competitive accuracy when they are abundant. The low-data setting is also more practical because collecting complete benchmark responses from many models is often costly.

\vspace{3pt}
\begin{mdframed}[backgroundcolor=questionBackground,linecolor=gray!35,linewidth=0.5pt,innertopmargin=5pt,innerbottommargin=5pt]
\noindent\textbf{(3) Sensitivity to Probe-Model Selection and Matching Norm.}
\end{mdframed}
\vspace{-5pt}
We next assess the sensitivity of BestRand to probe selection and matching norm. We compare probes sampled uniformly across the score range, at random, or from its low and high ends. Using two-sided Wilcoxon signed-rank tests across benchmarks, Figure~\ref{fig:BestRand-ablation-tradeoffs}(c) shows no significant differences except between Uniform and Low, where Uniform performs better. This suggests that probe selection should avoid using only low-scoring models. Figure~\ref{fig:BestRand-ablation-tradeoffs}(d) shows similar results for $L_1$, $L_2$, and $L_\infty$, with no significant differences according to the Wilcoxon signed-rank test. BestRand is therefore robust to both choices, and we use random selection with $L_2$ matching by default.

\begin{figure}[t]
\centering
\includegraphics[width=\columnwidth]{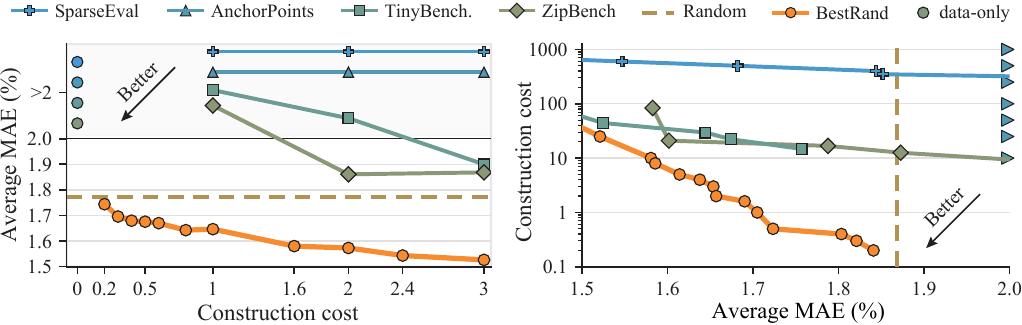}
\vspace{-18pt}
\caption{Low-cost compression at 20\% retention. (a) At the same construction cost, BestRand achieves substantially lower average MAE across 12 text benchmarks. Response-free methods appear at $\mathsf{Cost}=0$ but generally have much larger errors. BestRand remains accurate at $\mathsf{Cost}<1$, while learnable baselines require $\mathsf{Cost}\geq1$. (b) Over the displayed accuracy range, BestRand requires substantially less construction cost to reach the same average MAE on six text benchmarks.}
\label{fig:BestRand-low-cost}
\end{figure}

\begin{table*}[t]
\centering
\caption{OOD performance under score-range, model-family, and temporal shifts. Arrows in the column headers point from the construction group to the test group. BestRand achieves the lowest average MAE in every setting. The final column averages the three OOD types equally.}
\label{tab:ablation-ood-20pct}
\vspace{2pt}
\footnotesize
\setlength{\tabcolsep}{3.5pt}
\renewcommand{\arraystretch}{1.08}
\newcommand{\oodstd}[1]{\,{\scriptsize\color{black!70}#1}}
\resizebox{\textwidth}{!}{%
\begin{tabular}{@{}lrrrrcr@{}}
\toprule
& \multicolumn{3}{c}{\textbf{Score Range OOD}} & \multicolumn{1}{c}{\textbf{Family OOD}} & \multicolumn{1}{c}{\textbf{Temporal OOD}} & \\
\cmidrule(lr){2-4}\cmidrule(lr){5-5}\cmidrule(lr){6-6}
\rowcolor{lightBlue}
\textbf{Method} & {\fontsize{7.5pt}{8.5pt}\selectfont\textbf{Mid$\to$Tails}} & {\fontsize{7.5pt}{8.5pt}\selectfont\textbf{Low$\to$High}} & {\fontsize{7.5pt}{8.5pt}\selectfont\textbf{High$\to$Low}} & {\fontsize{7.5pt}{8.5pt}\selectfont\textbf{Held-out Families}} & \multicolumn{1}{c}{{\fontsize{7.5pt}{8.5pt}\selectfont\textbf{Past$\to$Future}}} & \textbf{Mean} \\
\midrule
SparseEval (Linear) & 18.50\oodstd{$\pm$3.24} & 21.50\oodstd{$\pm$5.74} & 18.31\oodstd{$\pm$4.58} & 22.45\oodstd{$\pm$3.63} & 24.32\oodstd{$\pm$6.65} & 22.07\oodstd{$\pm$2.98} \\
SparseEval (MLP) & 16.13\oodstd{$\pm$7.04} & 16.43\oodstd{$\pm$7.57} & 21.59\oodstd{$\pm$9.64} & 11.87\oodstd{$\pm$5.89} & 11.71\oodstd{$\pm$6.05} & 13.88\oodstd{$\pm$6.12} \\
Anchor Points & 4.94\oodstd{$\pm$1.62} & 5.63\oodstd{$\pm$3.03} & 6.96\oodstd{$\pm$4.52} & 7.26\oodstd{$\pm$1.93} & 8.23\oodstd{$\pm$2.64} & 7.11\oodstd{$\pm$1.57} \\
TinyBenchmarks & 2.34\oodstd{$\pm$1.17} & 2.46\oodstd{$\pm$1.35} & 3.03\oodstd{$\pm$1.93} & 2.20\oodstd{$\pm$0.93} & 2.27\oodstd{$\pm$1.01} & 2.36\oodstd{$\pm$0.89} \\
ZipBench & 2.03\oodstd{$\pm$0.94} & 2.43\oodstd{$\pm$1.62} & 2.30\oodstd{$\pm$1.06} & 2.16\oodstd{$\pm$1.28} & 2.15\oodstd{$\pm$1.28} & 2.19\oodstd{$\pm$1.21} \\
\addlinespace[1pt]
Random & 1.81\oodstd{$\pm$0.88} & 2.02\oodstd{$\pm$1.06} & 1.78\oodstd{$\pm$0.88} & 1.97\oodstd{$\pm$0.95} & 2.01\oodstd{$\pm$1.00} & 1.95\oodstd{$\pm$0.95} \\
\rowcolor{lightBlue}\textbf{BestRand (ours)} & \textbf{1.60}\oodstd{$\pm$0.83} & \textbf{1.74}\oodstd{$\pm$1.01} & \textbf{1.61}\oodstd{$\pm$0.81} & \textbf{1.70}\oodstd{$\pm$0.89} & \textbf{1.73}\oodstd{$\pm$0.91} & \textbf{1.69}\oodstd{$\pm$0.88} \\
\bottomrule
\end{tabular}%
}
\end{table*}

\vspace{3pt}
\begin{mdframed}[backgroundcolor=questionBackground,linecolor=gray!35,linewidth=0.5pt,innertopmargin=5pt,innerbottommargin=5pt]
\noindent\textbf{(4) Does BestRand Provide a Better Accuracy--Cost Trade-off than Existing Methods?}
\end{mdframed}
\vspace{-5pt}
Methods differ in how many historical model responses they use. Response-free methods require no historical model-item scores but generally yield much higher MAE. Among response-based methods, the same number of construction models can also incur different costs because BestRand evaluates each model only on its candidate pool, whereas several baselines require full-benchmark responses. We therefore define normalized construction cost as $\mathsf{Cost}=R/N$, where $R$ is the total number of observed model-item scores and $N$ is the benchmark size. Thus, $\mathsf{Cost}=1$ corresponds to one full evaluation of one model. Under this common budget, Figure~\ref{fig:BestRand-low-cost} shows that BestRand achieves lower MAE at the same cost and requires less cost to achieve the same MAE.

\vspace{3pt}
\begin{mdframed}[backgroundcolor=questionBackground,linecolor=gray!35,linewidth=0.5pt,innertopmargin=5pt,innerbottommargin=5pt]
\noindent\textbf{(5) OOD Generalization across Score-Range, Model-Family, and Temporal Shifts.}
\end{mdframed}
\vspace{-5pt}
Table~\ref{tab:ablation-ood-20pct} evaluates BestRand under three types of OOD shifts. Score-range OOD covers Mid$\to$Tails, which uses middle-scoring models for construction and low- and high-scoring models for testing, as well as Low$\to$High and High$\to$Low. Family OOD uses disjoint model families for construction and testing. Temporal OOD uses older models for construction and newer models for testing. Methods that learn from historical performance records perform worse than Random in every setting, showing that fitting these records can hurt under distribution shifts. In contrast, BestRand achieves the lowest error in all five settings and reduces average OOD error by 21.8\% relative to competitive baselines.

\section{Conclusion}
In this paper, we developed a theoretical framework that characterizes the error distribution of benchmark compression and provides a principled basis for understanding and predicting its accuracy and reliability.
Extensive empirical results closely match our theoretical results. 
Guided by our theory, we introduced BestRand, which improves the stability of random sampling at low construction cost. 
Across 70+ text, multimodal, and agent benchmarks, together with VLA evaluation, BestRand improved accuracy and stability over strong baselines and remained robust to distribution shifts. 
More broadly, our findings suggest that improving benchmark compression may not require complex selection, but rather understanding when and why simple sampling fails and correcting it at low cost.

\subsection*{AI use statement}
AI assistants were used in this work exclusively for language polishing and code implementation. All research ideas and scientific conclusions are original contributions of the authors. The authors reviewed all LLM-assisted outputs and take full responsibility for the final content of this paper.

\subsection*{Reproducibility statement}
To support reproducibility, we document all key methodological, experimental, and theoretical details. Section~\ref{sec:theory-method} presents the complete formulation and procedure of BestRand. Sections~\ref{sec:experiments} and~\ref{sec:analysis-ablation} specify the evaluation protocol, metrics, default settings, ablations, construction cost, and OOD evaluation. Appendix~\ref{app:experiment-details} gives the model splits, benchmark sources and metrics, baseline details, and hyperparameters. Appendices~\ref{app:additional-results} and~\ref{app:additional-analysis} report detailed results and describe the protocols for the VLA, multi-metric, and data-only experiments. Appendix~\ref{app:proofs} contains the full proofs, derivations, assumption checks, and numerical validation. We will make all code and data publicly available.

\bibliography{iclr2027_conference}
\bibliographystyle{iclr2027_conference}

\newpage

\appendix

\addtocontents{toc}{\protect\setcounter{tocdepth}{2}}
\renewcommand{\contentsname}{Appendix Contents}
\tableofcontents
\clearpage

\section{Additional Experiment Details}
\label{app:experiment-details}

\subsection{Details of Data Splits}
\label{app:data-splits}
We strictly separate construction and test models to evaluate whether a compressed benchmark transfers to unseen models. For each benchmark, we select five construction models and reserve a disjoint set of models for testing. At each retention ratio, all methods use the same construction and test split. Methods may use only responses from the construction models for item selection or estimator fitting, while the test models are used exclusively for evaluation. Response-free methods use no model responses during construction and are evaluated on the same test models.

\subsection{Details of Benchmarks Considered}
\label{app:benchmark-details}

Section~\ref{sec:experiments} evaluates BestRand on 70 benchmarks spanning text, multimodal, and agent tasks, while Appendix~\ref{app:bestrand-vla-effectiveness} extends the evaluation to two VLA benchmarks. Tables~\ref{tab:benchmark-inventory} and~\ref{tab:benchmark-inventory-mm-agent} list all benchmarks and report their names, references, public links, numbers of items and test models, the metrics used in our experiments, and whether they report multiple metrics. For benchmarks with multiple metrics, we select one representative metric for the main experiments. Appendix~\ref{app:bestrand-multi-metric-effectiveness} further evaluates BestRand in multi-metric settings and discusses its extension to jointly consider multiple metrics.

\begingroup
\footnotesize
\urlstyle{same}
\setlength{\tabcolsep}{2.8pt}
\renewcommand{\arraystretch}{1.25}
\newcommand{\benchSingleYes}{\textcolor{green!45!black}{\ding{51}}}
\newcommand{\benchSingleNo}{\textcolor{red!70!black}{\ding{55}}}
\newcommand{\hug}{\raisebox{-.2ex}{\includegraphics[height=2.1ex]{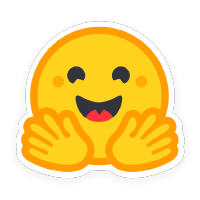}}\,}
\newcommand{\github}{\raisebox{-.2ex}{\includegraphics[height=2ex]{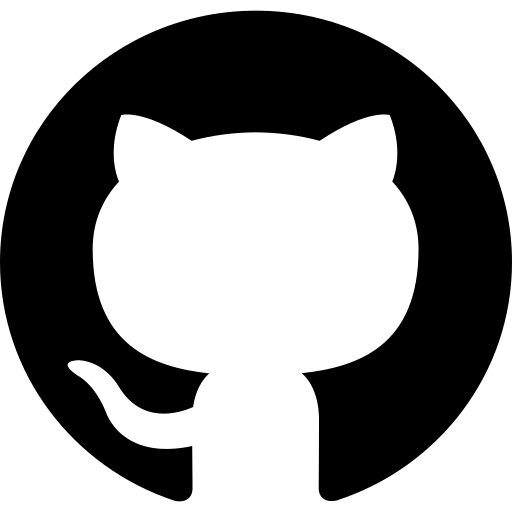}}\,}
\newcommand{\weblinkicon}{\raisebox{.1ex}{\scriptsize$\nearrow$}\,}

\begin{table}[t]
\centering
\caption{Details of the 16 multimodal and 10 agent benchmarks used in the main experiments and the two VLA benchmarks used in the extension. Metric (Single?) reports the metric used in our experiments and whether it is the benchmark's only official metric.}
\vspace{3pt}
\label{tab:benchmark-inventory-mm-agent}
\resizebox{\linewidth}{!}{%
\begin{tabular}{@{}cllrc@{\hspace{10pt}}l@{}}
\toprule
\textbf{Type} & \multicolumn{1}{c}{\hspace*{-25pt}\textbf{Benchmark}} & \textbf{Public Link} & \textbf{\# Items} & \textbf{\# Test Models} & \textbf{Metric (Single?)} \\
\midrule
\multirow[c]{16}{*}{\rotatebox[origin=c]{90}{\textbf{Multimodal Benchmarks}}} & CountBenchQA~\citep{beyer2024paligemma} & \hug\href{https://huggingface.co/datasets/vikhyatk/CountBenchQA}{{\scriptsize\nolinkurl{HF/CountBenchQA}}} & 491 & 11 & Accuracy (\benchSingleYes) \\
 & DynaMath~\citep{zou2025dynamath} & \hug\href{https://huggingface.co/datasets/DynaMath/DynaMath_Sample}{{\scriptsize\nolinkurl{HF/DynaMath_Sample}}} & 5,010 & 20 & Accuracy (\benchSingleNo) \\
 & LogicVista~\citep{xiao2024logicvista} & \github\href{https://github.com/Yijia-Xiao/LogicVista}{{\scriptsize\nolinkurl{GH/LogicVista}}} & 448 & 20 & Accuracy (\benchSingleYes) \\
 & MathVerse~\citep{zhang2024mathverse} & \hug\href{https://huggingface.co/datasets/AI4Math/MathVerse}{{\scriptsize\nolinkurl{HF/MathVerse}}} & 788 & 20 & Accuracy (\benchSingleYes) \\
 & MathVista~\citep{lu2024mathvista} & \hug\href{https://huggingface.co/datasets/AI4Math/MathVista}{{\scriptsize\nolinkurl{HF/MathVista}}} & 1,000 & 10 & Accuracy (\benchSingleYes) \\
 & MME~\citep{fu2026mme} & \github\href{https://github.com/BradyFU/Awesome-Multimodal-Large-Language-Models/tree/Evaluation}{{\scriptsize\nolinkurl{GH/MME}}} & 2,374 & 16 & MME Score (\benchSingleYes) \\
 & MMMU~\citep{yue2024mmmu} & \hug\href{https://huggingface.co/datasets/MMMU/MMMU}{{\scriptsize\nolinkurl{HF/MMMU}}} & 1,050 & 8 & Accuracy (\benchSingleYes) \\
 & MMMU-Pro~\citep{yue2025mmmu} & \hug\href{https://huggingface.co/datasets/MMMU/MMMU_Pro}{{\scriptsize\nolinkurl{HF/MMMU_Pro}}} & 1,730 & 20 & Accuracy (\benchSingleYes) \\
 & MMStar~\citep{chen2024we} & \hug\href{https://huggingface.co/datasets/Lin-Chen/MMStar}{{\scriptsize\nolinkurl{HF/MMStar}}} & 1,500 & 7 & Accuracy (\benchSingleYes) \\
 & MMVP~\citep{tong2024eyes} & \hug\href{https://huggingface.co/datasets/MMVP/MMVP}{{\scriptsize\nolinkurl{HF/MMVP}}} & 300 & 7 & Accuracy (\benchSingleYes) \\
 & OCRBench v2~\citep{fu2026ocrbench} & \hug\href{https://huggingface.co/datasets/morpheushoc/OCRBenchv2}{{\scriptsize\nolinkurl{HF/OCRBenchv2}}} & 10,000 & 8 & OCR Score (\benchSingleYes) \\
 & OmniDocBench~\citep{ouyang2025omnidocbench} & \github\href{https://github.com/opendatalab/OmniDocBench/tree/v1_5}{{\scriptsize\nolinkurl{GH/OmniDocBench@v1.5}}} & 1,355 & 6 & NED (\benchSingleNo) \\
 & RealWorldQA~\citep{ai2024grok} & \hug\href{https://huggingface.co/datasets/xai-org/RealworldQA}{{\scriptsize\nolinkurl{HF/RealworldQA}}} & 765 & 7 & Accuracy (\benchSingleYes) \\
 & SpatialEval~\citep{wang2024picture} & \hug\href{https://huggingface.co/datasets/MilaWang/SpatialEval}{{\scriptsize\nolinkurl{HF/SpatialEval}}} & 4,635 & 9 & Accuracy (\benchSingleYes) \\
 & VisualPuzzles~\citep{song2025visualpuzzles} & \hug\href{https://huggingface.co/datasets/neulab/VisualPuzzles}{{\scriptsize\nolinkurl{HF/VisualPuzzles}}} & 1,168 & 5 & Exact Match (\benchSingleYes) \\
 & VisualWebBench~\citep{liu2024visualwebbench} & \github\href{https://github.com/VisualWebBench/VisualWebBench}{{\scriptsize\nolinkurl{GH/VisualWebBench}}} & 1,534 & 5 & Accuracy (\benchSingleNo) \\
\midrule
\multirow[c]{10}{*}{\rotatebox[origin=c]{90}{\textbf{Agent Benchmarks}}} & ACEBench~\citep{chen2025acebench} & \github\href{https://github.com/chenchen0103/ACEBench}{{\scriptsize\nolinkurl{GH/ACEBench}}} & 1,023 & 20 & Score (\benchSingleNo) \\
 & BFCL~\citep{patil2025berkeley} & \github\href{https://github.com/ShishirPatil/gorilla/tree/main/berkeley-function-call-leaderboard}{{\scriptsize\nolinkurl{GH/BFCL}}} & 5,106 & 16 & Accuracy (\benchSingleNo) \\
 & ComplexFuncBench~\citep{zhong2025complexfuncbench} & \hug\href{https://huggingface.co/datasets/THUDM/ComplexFuncBench}{{\scriptsize\nolinkurl{HF/ComplexFuncBench}}} & 1,000 & 20 & Success Rate (\benchSingleNo) \\
 & DrafterBench~\citep{li2025drafterbench} & \hug\href{https://huggingface.co/datasets/Eason666/DrafterBench}{{\scriptsize\nolinkurl{HF/DrafterBench}}} & 1,920 & 20 & Score (\benchSingleNo) \\
 & MultiChallenge~\citep{deshpande2025multichallenge} & \hug\href{https://huggingface.co/datasets/ScaleAI/MultiChallenge}{{\scriptsize\nolinkurl{HF/MultiChallenge}}} & 273 & 20 & Accuracy (\benchSingleYes) \\
 & NexusBench~\citep{nexusbench} & \github\href{https://github.com/nexusflowai/NexusBench}{{\scriptsize\nolinkurl{GH/NexusBench}}} & 1,118 & 14 & Accuracy (\benchSingleYes) \\
 & OSWorld~\citep{xie2024osworld} & \github\href{https://github.com/xlang-ai/OSWorld}{{\scriptsize\nolinkurl{GH/OSWorld}}} & 361 & 20 & Success Rate (\benchSingleYes) \\
 & tau-bench~\citep{yao2024tau} & \github\href{https://github.com/sierra-research/tau-bench}{{\scriptsize\nolinkurl{GH/tau-bench}}} & 165 & 20 & Pass Rate (\benchSingleNo) \\
 & Terminal-Bench~\citep{merrill2026terminal} & \hug\href{https://huggingface.co/datasets/harborframework/terminal-bench-2.0}{{\scriptsize\nolinkurl{HF/terminal-bench-2.0}}} & 89 & 20 & Reward (\benchSingleYes) \\
 & Toolathlon~\citep{li2026tool} & \github\href{https://github.com/hkust-nlp/Toolathlon}{{\scriptsize\nolinkurl{GH/Toolathlon}}} & 108 & 20 & Success Rate (\benchSingleYes) \\
\midrule
\multirow[c]{2}{*}{\rotatebox[origin=c]{90}{\textbf{VLA}}} & Bench2Drive~\citep{jia2024bench2drive} & \github\href{https://github.com/Thinklab-SJTU/Bench2Drive}{{\scriptsize\nolinkurl{GH/Bench2Drive}}} & 220 & 50 & Success Rate (\benchSingleNo) \\
 & Fail2Drive~\citep{gerstenecker2026fail2drive} & \github\href{https://github.com/autonomousvision/fail2drive}{{\scriptsize\nolinkurl{GH/Fail2Drive}}} & 200 & 37 & Success Rate (\benchSingleNo) \\
\bottomrule
\end{tabular}%
}
\end{table}

In Section~\ref{sec:analysis-ablation}, we examine how performance changes as more construction models become available, using evaluation records from up to 1,000 models on ARC-Challenge, BBH, GPQA, GSM8K, TruthfulQA, HellaSwag, IFEval, MATH, MMLU-Pro, MMLU, MuSR, and WinoGrande provided by \citet{huang2025routereval}.

\begin{table}[t]
\centering
\caption{Details of the 44 text benchmarks used in the main experiments. Each benchmark uses five construction models and the remaining models for testing. Metric (Single?) reports the metric used in our experiments and whether it is the benchmark's only official metric.}
\vspace{3pt}
\label{tab:benchmark-inventory}
\resizebox{\linewidth}{!}{%
\begin{tabular}{@{}cllrc@{\hspace{10pt}}l@{}}
\toprule
\textbf{Type} & \multicolumn{1}{c}{\hspace*{-25pt}\textbf{Benchmark}} & \textbf{Public Link} & \textbf{\# Items} & \textbf{\# Test Models} & \textbf{Metric (Single?)} \\
\midrule
\multirow[c]{44}{*}{\rotatebox[origin=c]{90}{\textbf{Text Benchmarks}}} & ACPBench Hard~\citep{kokel2026acpbench} & \hug\href{https://huggingface.co/datasets/ibm-research/acp_bench}{{\scriptsize\nolinkurl{HF/acp_bench}}} & 1,040 & 20 & Accuracy (\benchSingleYes) \\
 & AGC-Bench~\citep{beaty2026agc} & \hug\href{https://huggingface.co/agcbench-2026}{{\scriptsize\nolinkurl{HF/agcbench-2026}}} & 100 & 20 & Accuracy (\benchSingleNo) \\
 & AIME~\citep{maa_invitational} & \weblinkicon\href{https://maa.org/maa-invitational-competitions/}{{\scriptsize\nolinkurl{MAA/AIME}}} & 60 & 20 & Accuracy (\benchSingleYes) \\
 & ARC-Challenge~\citep{clark2018think} & \hug\href{https://huggingface.co/datasets/allenai/ai2_arc}{{\scriptsize\nolinkurl{HF/ai2_arc}}} & 1,172 & 20 & Accuracy (\benchSingleYes) \\
 & ARC-Easy~\citep{clark2018think} & \hug\href{https://huggingface.co/datasets/allenai/ai2_arc}{{\scriptsize\nolinkurl{HF/ai2_arc}}} & 2,376 & 20 & Accuracy (\benchSingleYes) \\
 & BBEH~\citep{kazemi2025big} & \github\href{https://github.com/google-deepmind/bbeh}{{\scriptsize\nolinkurl{GH/bbeh}}} & 4,520 & 20 & Accuracy (\benchSingleYes) \\
 & BBH~\citep{suzgun2023challenging} & \github\href{https://github.com/suzgunmirac/BIG-Bench-Hard}{{\scriptsize\nolinkurl{GH/BIG-Bench-Hard}}} & 5,761 & 20 & Accuracy (\benchSingleYes) \\
 & BEIR NFCorpus~\citep{thakur2021beir} & \hug\href{https://huggingface.co/datasets/BeIR/nfcorpus}{{\scriptsize\nolinkurl{HF/nfcorpus}}} & 323 & 17 & nDCG@10 (\benchSingleNo) \\
 & BigCodeBench~\citep{zhuo2025bigcodebench} & \hug\href{https://huggingface.co/datasets/bigcode/bigcodebench}{{\scriptsize\nolinkurl{HF/bigcodebench}}} & 1,140 & 20 & Pass@1 (\benchSingleYes) \\
 & BoolQ~\citep{clark2019boolq} & \hug\href{https://huggingface.co/datasets/google/boolq}{{\scriptsize\nolinkurl{HF/boolq}}} & 3,270 & 20 & Accuracy (\benchSingleYes) \\
 & C3~\citep{sun2020investigating} & \github\href{https://github.com/CLUEbenchmark/CLUE}{{\scriptsize\nolinkurl{GH/CLUE}}} & 1,825 & 20 & Accuracy (\benchSingleYes) \\
 & CommonsenseQA~\citep{talmor2019commonsenseqa} & \hug\href{https://huggingface.co/datasets/tau/commonsense_qa}{{\scriptsize\nolinkurl{HF/commonsense_qa}}} & 1,221 & 20 & Accuracy (\benchSingleYes) \\
 & CyberSecEval 2~\citep{bhatt2024cyberseceval} & \github\href{https://github.com/meta-llama/PurpleLlama/tree/main/CybersecurityBenchmarks}{{\scriptsize\nolinkurl{GH/CybersecurityBenchmarks}}} & 1,336 & 5 & Success Rate (\benchSingleNo) \\
 & Finance Fund.~\citep{krumdick2024bizbench} & \hug\href{https://huggingface.co/collections/michaelkrumdickkensho/financefundamentals}{{\scriptsize\nolinkurl{HF/FinanceFundamentals}}} & 600 & 20 & Accuracy (\benchSingleYes) \\
 & FLASK~\citep{ye2024flask} & \github\href{https://github.com/kaistAI/FLASK}{{\scriptsize\nolinkurl{GH/FLASK}}} & 1,740 & 5 & Score (\benchSingleNo) \\
 & GPQA~\citep{rein2023gpqa} & \hug\href{https://huggingface.co/datasets/Idavidrein/gpqa}{{\scriptsize\nolinkurl{HF/gpqa}}} & 546 & 20 & Accuracy (\benchSingleYes) \\
 & GSM-Symbolic~\citep{mirzadeh2025gsm} & \github\href{https://github.com/apple/ml-gsm-symbolic}{{\scriptsize\nolinkurl{GH/ml-gsm-symbolic}}} & 7,500 & 20 & Accuracy (\benchSingleYes) \\
 & GSM8K~\citep{cobbe2021training} & \hug\href{https://huggingface.co/datasets/openai/gsm8k}{{\scriptsize\nolinkurl{HF/gsm8k}}} & 1,319 & 20 & Accuracy (\benchSingleYes) \\
 & HellaSwag~\citep{zellers2019hellaswag} & \hug\href{https://huggingface.co/datasets/Rowan/hellaswag}{{\scriptsize\nolinkurl{HF/hellaswag}}} & 10,042 & 20 & Accuracy (\benchSingleYes) \\
 & HumanEval~\citep{chen2021evaluating} & \github\href{https://github.com/openai/human-eval}{{\scriptsize\nolinkurl{GH/human-eval}}} & 164 & 20 & Pass@1 (\benchSingleYes) \\
 & IFEval~\citep{zhou2023instruction} & \hug\href{https://huggingface.co/datasets/google/IFEval}{{\scriptsize\nolinkurl{HF/IFEval}}} & 541 & 20 & Accuracy (\benchSingleNo) \\
 & InFoBench~\citep{qin2024infobench} & \hug\href{https://huggingface.co/datasets/kqsong/InFoBench}{{\scriptsize\nolinkurl{HF/InFoBench}}} & 500 & 14 & DRFR (\benchSingleYes) \\
 & JudgeBench~\citep{tan2025judgebench} & \hug\href{https://huggingface.co/datasets/ScalerLab/JudgeBench}{{\scriptsize\nolinkurl{HF/JudgeBench}}} & 350 & 20 & Judge Accuracy (\benchSingleYes) \\
 & LIFBench~\citep{wu2025lifbench} & \github\href{https://github.com/SheldonWu0327/LIFBench-2024}{{\scriptsize\nolinkurl{GH/LIFBench-2024}}} & 2,802 & 15 & Total Score (\benchSingleNo) \\
 & LiveCodeBench~\citep{jain2025livecodebench} & \hug\href{https://huggingface.co/datasets/livecodebench/code_generation_lite}{{\scriptsize\nolinkurl{HF/code_generation_lite}}} & 880 & 17 & Pass@1 (\benchSingleYes) \\
 & LogiQA~\citep{liu2020logiqa} & \github\href{https://github.com/lgw863/LogiQA-dataset}{{\scriptsize\nolinkurl{GH/LogiQA-dataset}}} & 1,302 & 20 & Exact Match (\benchSingleYes) \\
 & LongBench v2~\citep{bai2025longbench} & \hug\href{https://huggingface.co/datasets/zai-org/LongBench-v2}{{\scriptsize\nolinkurl{HF/LongBench-v2}}} & 503 & 12 & Accuracy (\benchSingleYes) \\
 & MATH lv5~\citep{hendrycks2021measuring} & \hug\href{https://huggingface.co/datasets/EleutherAI/hendrycks_math}{{\scriptsize\nolinkurl{HF/hendrycks_math}}} & 1,324 & 20 & Accuracy (\benchSingleYes) \\
 & MedQA~\citep{jin2021disease} & \github\href{https://github.com/jind11/MedQA}{{\scriptsize\nolinkurl{GH/MedQA}}} & 1,273 & 20 & Accuracy (\benchSingleYes) \\
 & MMLU~\citep{hendrycks2020measuring} & \hug\href{https://huggingface.co/datasets/cais/mmlu}{{\scriptsize\nolinkurl{HF/mmlu}}} & 14,042 & 20 & Accuracy (\benchSingleYes) \\
 & MMLU-Pro~\citep{wang2024mmlu} & \hug\href{https://huggingface.co/datasets/TIGER-Lab/MMLU-Pro}{{\scriptsize\nolinkurl{HF/MMLU-Pro}}} & 12,032 & 20 & Accuracy (\benchSingleYes) \\
 & MuSR~\citep{sprague2024musr} & \hug\href{https://huggingface.co/datasets/TAUR-Lab/MuSR}{{\scriptsize\nolinkurl{HF/MuSR}}} & 756 & 20 & Accuracy (\benchSingleYes) \\
 & OpenBookQA~\citep{mihaylov2018can} & \hug\href{https://huggingface.co/datasets/allenai/openbookqa}{{\scriptsize\nolinkurl{HF/openbookqa}}} & 500 & 20 & Accuracy (\benchSingleYes) \\
 & PAWS~\citep{zhang2019paws} & \hug\href{https://huggingface.co/datasets/google-research-datasets/paws}{{\scriptsize\nolinkurl{HF/paws}}} & 8,000 & 20 & Accuracy (\benchSingleYes) \\
 & PlanBench~\citep{valmeekam2023planbench} & \github\href{https://github.com/karthikv792/LLMs-Planning}{{\scriptsize\nolinkurl{GH/LLMs-Planning}}} & 500 & 17 & Accuracy (\benchSingleYes) \\
 & Pre-Flight~\citep{brooker2026pre} & \github\href{https://github.com/UKGovernmentBEIS/inspect_evals/tree/main/src/inspect_evals/pre_flight}{{\scriptsize\nolinkurl{GH/pre_flight}}} & 300 & 20 & Accuracy (\benchSingleYes) \\
 & RACE-H~\citep{lai2017race} & \hug\href{https://huggingface.co/datasets/ehovy/race}{{\scriptsize\nolinkurl{HF/race}}} & 3,498 & 20 & Accuracy (\benchSingleYes) \\
 & RepoBench~\citep{liu2024repobench} & \github\href{https://github.com/Leolty/repobench}{{\scriptsize\nolinkurl{GH/repobench}}} & 937 & 7 & CodeBLEU (\benchSingleNo) \\
 & SciBench~\citep{wang2023scibench} & \github\href{https://github.com/mandyyyyii/scibench}{{\scriptsize\nolinkurl{GH/scibench}}} & 583 & 19 & Accuracy (\benchSingleYes) \\
 & SciCode~\citep{tian2024scicode} & \github\href{https://github.com/scicode-bench/SciCode}{{\scriptsize\nolinkurl{GH/SciCode}}} & 288 & 16 & Accuracy (\benchSingleYes) \\
 & SimpleQA~\citep{wei2024measuring} & \github\href{https://github.com/openai/simple-evals}{{\scriptsize\nolinkurl{GH/simple-evals}}} & 4,326 & 13 & Correctness (\benchSingleNo) \\
 & TheoremQA~\citep{chen2023theoremqa} & \hug\href{https://huggingface.co/datasets/TIGER-Lab/TheoremQA}{{\scriptsize\nolinkurl{HF/TheoremQA}}} & 800 & 20 & Accuracy (\benchSingleYes) \\
 & TruthfulQA~\citep{lin2022truthfulqa} & \github\href{https://github.com/sylinrl/TruthfulQA}{{\scriptsize\nolinkurl{GH/TruthfulQA}}} & 817 & 20 & Truthfulness (\benchSingleNo) \\
 & WinoGrande~\citep{sakaguchi2021winogrande} & \hug\href{https://huggingface.co/datasets/allenai/winogrande}{{\scriptsize\nolinkurl{HF/winogrande}}} & 1,267 & 20 & Accuracy (\benchSingleYes) \\
\bottomrule
\end{tabular}%
}
\end{table}
\clearpage

\endgroup

\subsection{Details of Baselines Considered}
\label{app:baseline-details}

We group the baselines by whether benchmark construction uses historical model responses. Response-based methods use these records to identify representative items and, when needed, learn how to estimate full-benchmark scores from selected-item responses. Random and the data-only methods use no historical performance records and are collectively termed response-free methods.

Response-based baselines differ in their item representations and score estimators. \textbf{SparseEval} \citep{zhang2026sparseeval} represents each item by its response vector across construction models. Starting from a clustered or random subset, it iteratively replaces items to reduce full-score prediction error on the construction models. A linear model or MLP then maps a new model's selected-item responses to its estimated full score. \textbf{Anchor Points} \citep{vivek2024anchor} derives item similarities from correlations between response patterns and applies $k$-medoids clustering. It selects each cluster medoid and estimates performance using a cluster-size-weighted average of the selected-item scores. \textbf{TinyBenchmarks} \citep{polo2024tinybenchmarks} fits an item response theory model to learn item difficulty and discrimination, clusters the resulting item representations, and selects one representative per cluster. For a new model, it infers ability from the selected responses and predicts performance on unobserved items. We use its gp-IRT estimator, which combines these predictions with the weighted observed-item score. \textbf{ZipBench} \citep{huang2026zipbenchlowcostframeworkcompressing} is designed for settings with limited construction budgets. It constructs synthetic response records by majority voting within a group of construction models and retains records spanning the observed performance range. It then fits an item response theory model to the augmented records, clusters the learned item representations, and estimates the full score using a cluster-size-weighted average of the selected-item scores.

Among the response-free baselines, \textbf{Random} uniformly samples a subset of items, whereas the four data-only methods select subsets using prompt-derived representations. All five methods estimate each model's full-benchmark score by averaging its scores on the selected items. Facility Location, MIS, and Coverage use Sentence-BERT embeddings \citep{reimers2019sentence}, whereas Agreement uses multilingual BERT representations \citep{devlin2019bert}. \textbf{Facility Location} \citep{yao2026coresets} greedily selects items that maximize semantic coverage, ensuring that each benchmark item is close to a selected item. \textbf{MIS} \citep{kjorvezir2026consistent} connects semantically similar prompts in a graph and greedily selects an independent set to reduce redundancy. \textbf{Agreement} \citep{purificato2026select} applies an eight-head attention module to the BERT representations, ranks prompts by the variance of their attention weights, and retains those with the greatest disagreement. \textbf{Coverage} \citep{purificato2026select} begins with a random prompt and repeatedly selects the prompt farthest from the most recently chosen one, encouraging coverage of distinct regions in the embedding space.

\subsection{Implementation and Hyperparameters}
\label{app:implementation-hyperparameters}
\paragraph{Main experiment.}
BestRand uses five probe models, 128 candidate subsets, a pool multiplier of four, and $L_2$ matching. We evaluate Random and BestRand over 10,000 trials to obtain stable estimates. We evaluate other baselines with three seeds to obtain reliable results. Anchor Points uses FasterPAM for $k$-medoids clustering. TinyBenchmarks and ZipBench use multidimensional two-parameter logistic IRT models with 10 and 5 dimensions, respectively. Both IRT models are trained for 2{,}000 epochs with an initial learning rate of $0.1$ and a decay factor of $0.9999$. ZipBench adds six synthetic response profiles. SparseEval performs 10 subset-refinement rounds and trains each proxy for 2{,}000 epochs. Its MLP has hidden widths of 128, 64, and 32 with dropout $0.1$. Both the linear and MLP variants use a learning rate of $5\times10^{-4}$ and train the final predictor for up to 20{,}000 epochs, with early stopping checked every 500 epochs using a patience of 30.

\paragraph{Data-only experiment.}
We evaluate data-only methods in Section~\ref{sec:analysis-ablation} and Appendix~\ref{app:data-only-methods-effectiveness} as response-free baselines. We evaluate the four methods at 20\% retention on the 12 text benchmarks used in the cost analysis. The primary setting uses 384-dimensional Sentence-BERT representations, while RoBERTa, RoBERTa's last-layer representation, and Longformer are used only for embedding sensitivity. Agreement uses \texttt{bert-base-multilingual-uncased}, a maximum sequence length of 64, eight attention heads, and a batch size of 32. MIS estimates its similarity threshold from 10{,}000 randomly sampled prompt pairs and searches percentiles 80 through 100 until the item budget can be met. Facility Location chooses its neighborhood scale from $\{0.1,0.5,1,5,10,100\}$ by three-fold validation based only on geometric coverage. 

\paragraph{OOD experiment.}
In Section~\ref{sec:analysis-ablation}, we evaluate OOD generalization at 20\% retention under score-range, model-family, and temporal shifts. For score-range OOD, we rank models by their full-benchmark scores, randomly sample construction models from the lowest, highest, or middle 75\%, and reserve the complementary 25\% for testing. Full scores are used only to define the OOD splits, and no test-model responses are used for item selection or estimator fitting. Model-family OOD uses disjoint families for construction and testing, while temporal OOD uses only models released earlier than every test model for construction. We use five splits with different random seeds for each shift setting to reduce sensitivity to any single model split. For each split, we evaluate Random and BestRand with 100 seeds and all other baselines with three seeds. We average MAE over seeds and splits within each benchmark, then report the mean and standard deviation across benchmarks.

\paragraph{Additional baseline experiment.}
In Appendix~\ref{app:broader-baseline-comparison}, we evaluate five additional baselines on 12 benchmarks using three random seeds for each stochastic method and a single run for the deterministic DatBench. SubLIME uses BERT representations and a rank correlation predictor to select from 1{,}000 candidate subsets after transport filtering. MINCE estimates score drift from 10{,}000 Monte Carlo subsets per retention ratio. EssenceBench uses BGE-M3 representations, redundancy thresholds of $0.9$, three genetic search rounds, and a regularized predictor with 25 spline terms. FlashEval retains the top 100{,}000 of three million candidates according to Kendall's $\tau_b$. DatBench ranks binary items by point-biserial correlation and uses Pearson correlation for nonbinary scores. All methods use identical data splits and item budgets, and test responses are used only for final evaluation.

\paragraph{Construction information requirements.}
Table~\ref{tab:baseline-inventory} summarizes the construction information required by BestRand and 15 baselines. Among response-based methods, BestRand has relatively low information requirements because its probe models score only a candidate pool rather than the full benchmark. This partial coverage reduces the required construction data and construction cost.

\begingroup
\footnotesize
\setlength{\tabcolsep}{3.5pt}
\renewcommand{\arraystretch}{1.16}
\newcommand{\baselineYes}{\textcolor{green!45!black}{\ding{51}}}
\newcommand{\baselineNo}{\textcolor{black!32}{\textemdash}}
\newcommand{\bestcell}[1]{\cellcolor{lightBlue}#1}
\begin{table}[t]
\centering
\caption{Comparison of the construction information required by BestRand and 15 diverse baselines. Text and Embed. denote direct prompt features and pretrained text embeddings. Scores denotes historical scores for individual models and items, All items indicates whether these scores must cover the entire benchmark, and Full score denotes the corresponding full benchmark aggregate. Among response-based methods, BestRand has lower information requirements because it uses scores from only a subset of items, reducing construction cost while retaining strong empirical performance.}
\vspace{5pt}
\label{tab:baseline-inventory}
\begin{tabular}{@{}clccccc@{}}
\toprule
& & \multicolumn{5}{c}{\textbf{Construction Information}} \\
\cmidrule(lr){3-7}
\textbf{Type} & \hspace{30pt}\textbf{Method} & \textbf{Text} & \textbf{Embed.} & \textbf{Scores} & \textbf{All items} & \textbf{Full score} \\
\midrule
\multirow[c]{5}{*}{\rotatebox[origin=c]{90}{\textbf{Response-free}}}
 & Random & \baselineNo & \baselineNo & \baselineNo & \baselineNo & \baselineNo \\
 & Facility Location~\citep{yao2026coresets} & \baselineNo & \baselineYes & \baselineNo & \baselineNo & \baselineNo \\
 & MIS~\citep{kjorvezir2026consistent} & \baselineNo & \baselineYes & \baselineNo & \baselineNo & \baselineNo \\
 & Agreement~\citep{purificato2026select} & \baselineNo & \baselineYes & \baselineNo & \baselineNo & \baselineNo \\
 & Coverage~\citep{purificato2026select} & \baselineNo & \baselineYes & \baselineNo & \baselineNo & \baselineNo \\
\midrule
\multirow[c]{11}{*}{\rotatebox[origin=c]{90}{\textbf{Response-based}}}
 & SparseEval (Linear)~\citep{zhang2026sparseeval} & \baselineNo & \baselineNo & \baselineYes & \baselineYes & \baselineYes \\
 & SparseEval (MLP)~\citep{zhang2026sparseeval} & \baselineNo & \baselineNo & \baselineYes & \baselineYes & \baselineYes \\
 & Anchor Points~\citep{vivek2024anchor} & \baselineNo & \baselineNo & \baselineYes & \baselineYes & \baselineNo \\
 & TinyBenchmarks~\citep{polo2024tinybenchmarks} & \baselineNo & \baselineNo & \baselineYes & \baselineYes & \baselineNo \\
 & ZipBench~\citep{huang2026zipbenchlowcostframeworkcompressing} & \baselineNo & \baselineNo & \baselineYes & \baselineYes & \baselineYes \\
 & SubLIME~\citep{saranathan2025sublime} & \baselineYes & \baselineYes & \baselineYes & \baselineYes & \baselineYes \\
 & MINCE~\citep{das2026mince} & \baselineNo & \baselineNo & \baselineYes & \baselineYes & \baselineYes \\
 & EssenceBench~\citep{wang2026rethinking} & \baselineNo & \baselineYes & \baselineYes & \baselineYes & \baselineYes \\
 & FlashEval~\citep{zhao2024flasheval} & \baselineNo & \baselineNo & \baselineYes & \baselineYes & \baselineYes \\
 & DatBench~\citep{joshi2026datbench} & \baselineNo & \baselineNo & \baselineYes & \baselineYes & \baselineYes \\
 & \bestcell{\textbf{BestRand (ours)}} & \bestcell{\baselineNo} & \bestcell{\baselineNo} & \bestcell{\baselineYes} & \bestcell{\baselineNo} & \bestcell{\baselineNo} \\
\bottomrule
\end{tabular}
\end{table}
\endgroup

\clearpage

\section{Additional Visualizations}
\label{app:additional-detailed-result-visualizations}

\begin{figure*}[h]
\includegraphics[width=\textwidth,height=0.82\textheight,keepaspectratio]{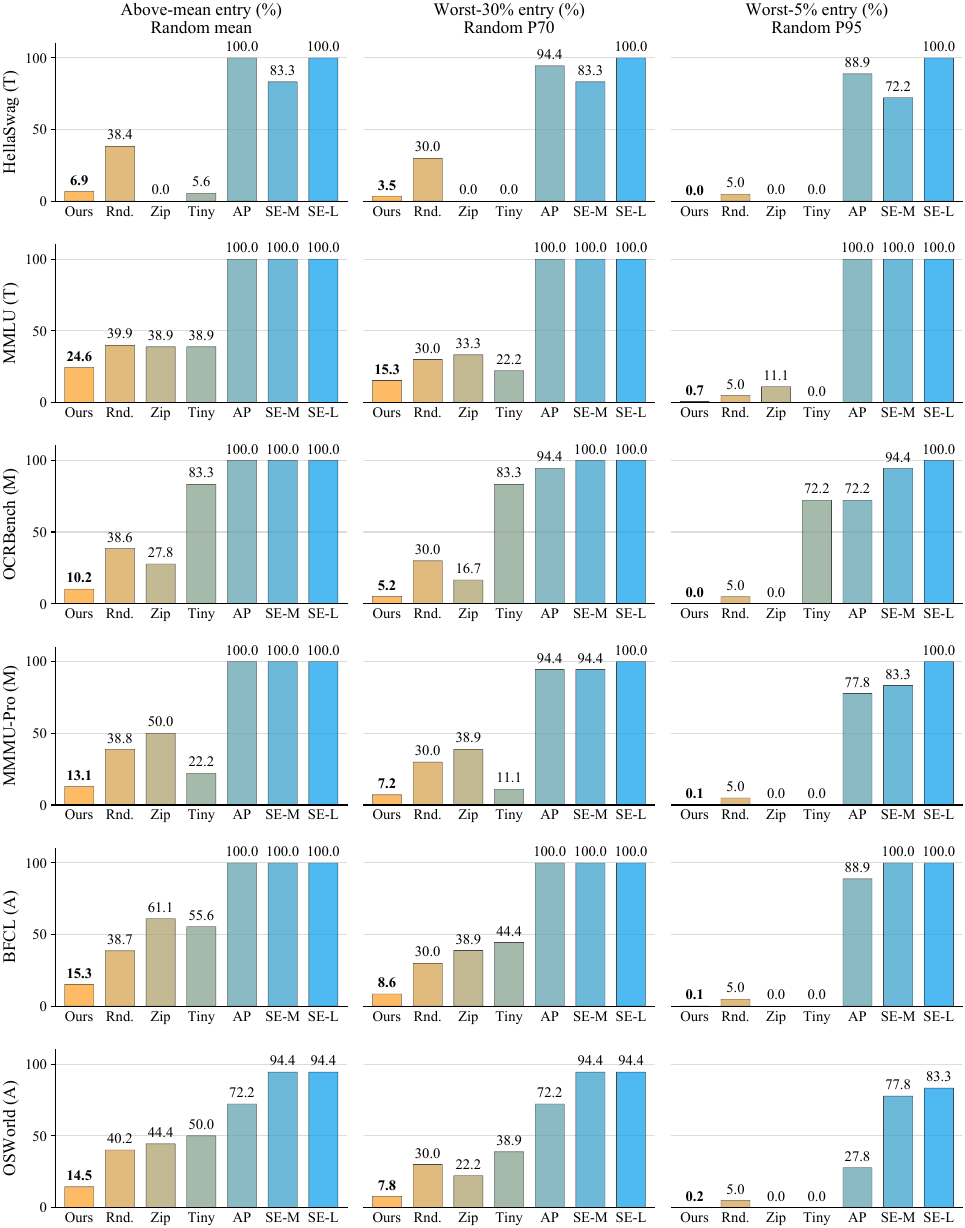}
\caption{High-error risk across six representative text, multimodal, and agent benchmarks. Labels T, M, and A denote the text, multimodal, and agent benchmark categories, respectively. Columns use Random's empirical mean, 70th percentile, and 95th percentile MAE as thresholds for the worse-than-average, worst 30\% tail, and worst 5\% tail region. Results are averaged equally across six retention ratios. Random and BestRand use 10{,}000 trials per ratio, while the other baselines use three trials. Rand, Zip, Tiny, AP, SE-M, and SE-L denote Random, ZipBench, TinyBenchmarks, Anchor Points, SparseEval MLP, and SparseEval Linear, respectively.}
\label{fig:tail-entry-appendix}
\end{figure*}

\begin{figure*}[h]
\includegraphics[width=\textwidth,height=0.82\textheight,keepaspectratio]{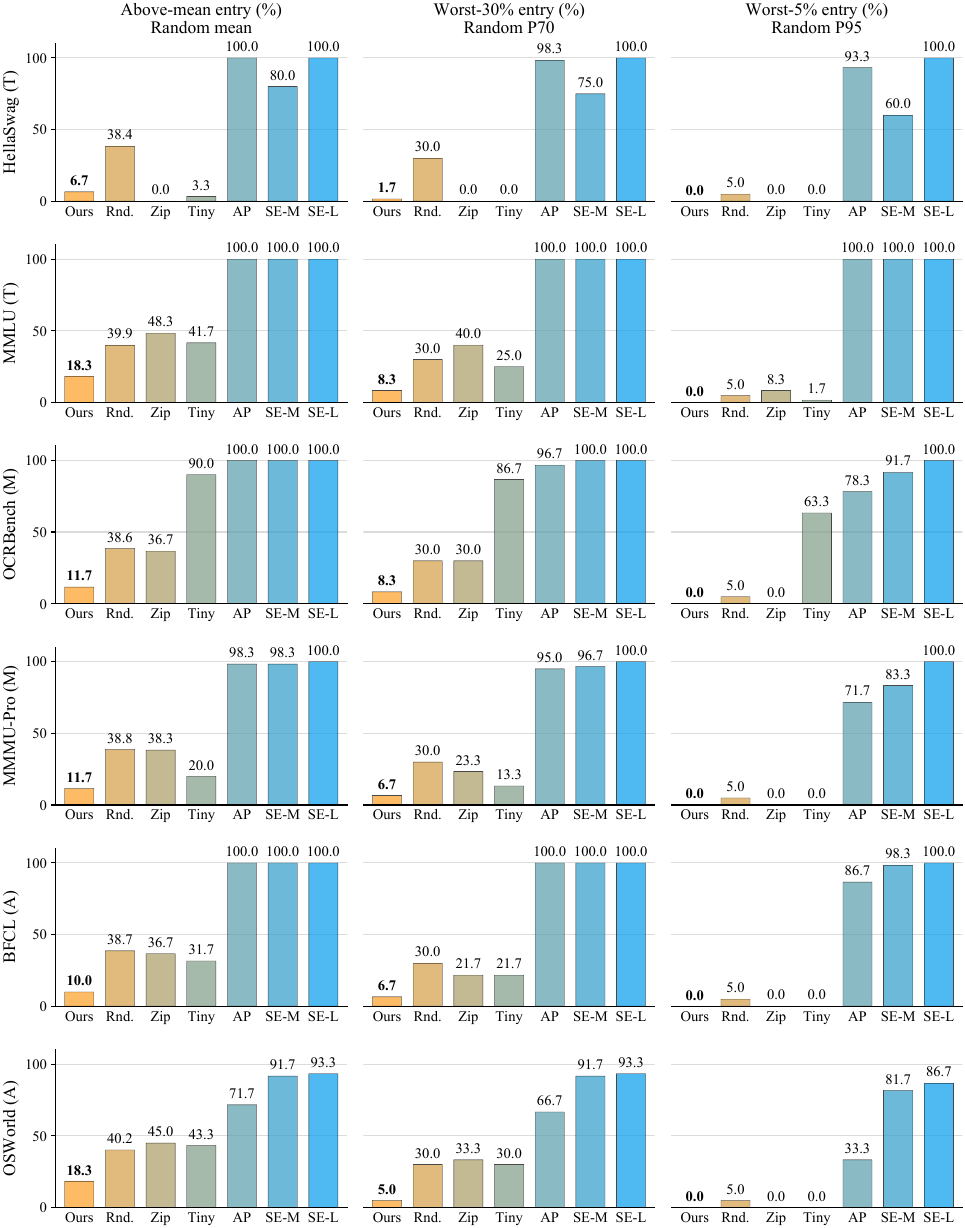}
\caption{We conduct an additional experiment on the same six representative benchmarks as in Figure~\ref{fig:tail-entry-appendix}. Random uses the same 10{,}000 trials, while BestRand and all other methods are evaluated with ten seeds per retention ratio. The high error risk patterns remain similar across the two experiments. Averaged over the worse-than-average and worst-30\% metrics, BestRand reduces high error risk relative to Random, ZipBench, and TinyBenchmarks by 68.06\% on the six benchmarks in Figure~\ref{fig:tail-entry-appendix} and by 71.06\% in this new experiment. This close agreement confirms that the main conclusion is robust to the number of seeds and that BestRand consistently reduces high error risk.}
\label{fig:tail-entry-appendix-2}
\end{figure*}

\begin{figure*}[h]
\centering
\includegraphics[width=\textwidth]{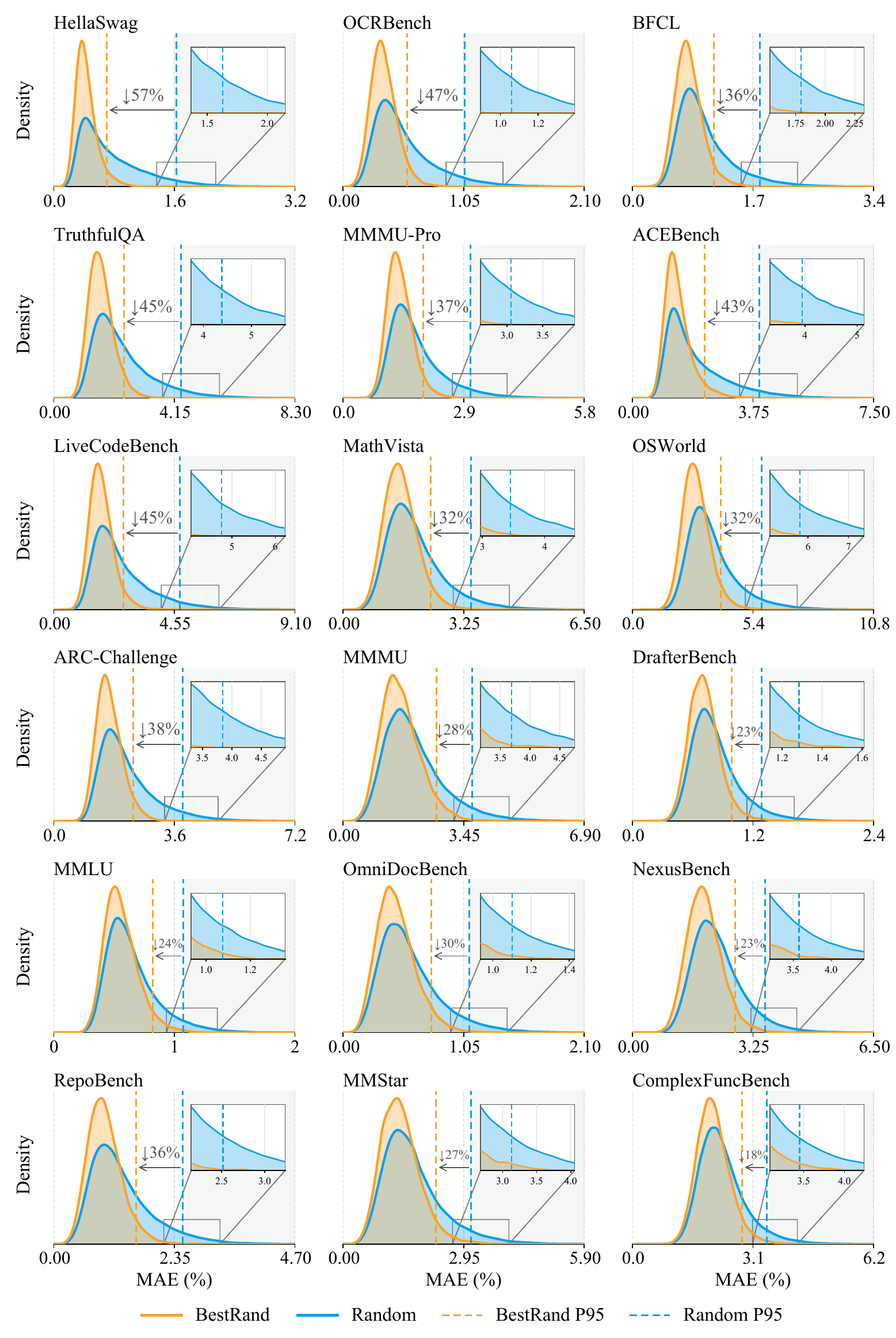}
\caption{MAE distributions of Random and BestRand across 18 text, multimodal, and agent benchmarks. BestRand consistently concentrates errors near zero and suppresses high-error outcomes.}
\label{fig:long-tail-distribution-appendix}
\end{figure*}

\begin{figure*}[h]
\centering
\includegraphics[width=\textwidth]{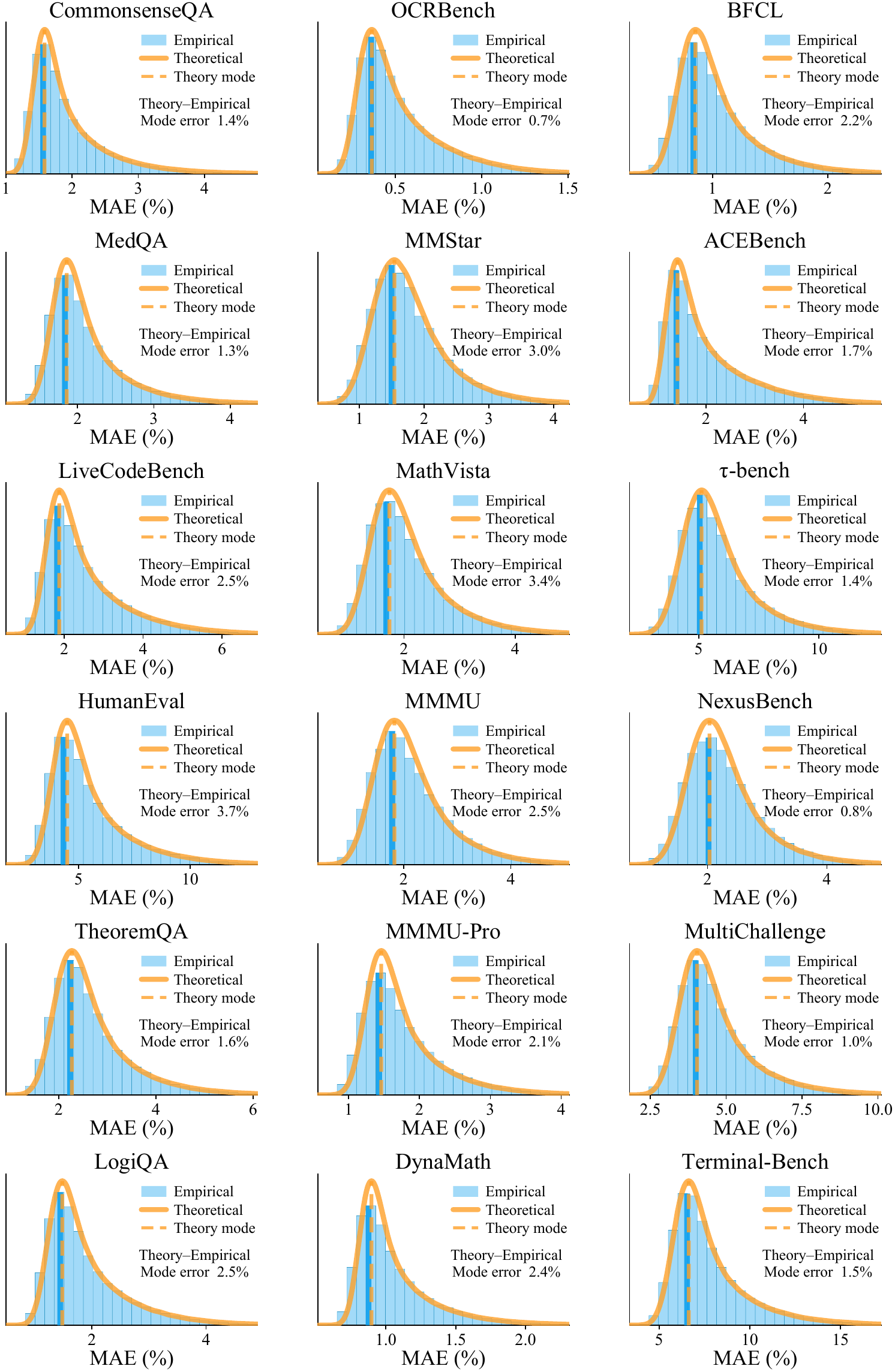}
\caption{Empirical MAE distributions closely match theoretical predictions across representative text, multimodal, and agent benchmarks, including the predicted shapes and modes.}
\label{fig:theory-validation-appendix}
\end{figure*}

\clearpage

\section{Additional Analysis}
\label{app:additional-analysis}

Section~\ref{sec:analysis-ablation} analyzes the key design choices of BestRand and evaluates its effectiveness, efficiency, robustness, and OOD generalization. This section complements those results with additional experiments on VLA benchmarks, multi-metric evaluation, and data-only compression methods.

\subsection{Effectiveness of BestRand on VLA Benchmarks}
\label{app:bestrand-vla-effectiveness}

Section~\ref{sec:experiments} reports results on text, multimodal, and agent benchmarks. Here, we provide additional results on VLA benchmarks, where evaluation is especially costly because each item requires an interactive task rollout rather than a single model inference. We evaluate BestRand on Bench2Drive \citep{jia2024bench2drive} and Fail2Drive \citep{gerstenecker2026fail2drive}. Both experiments construct subsets from five training records and evaluate them on held-out records at removal ratios of 50\% and 65\%. To evaluate the direct transfer of the original method, BestRand-SR selects routes using success rate alone. The next subsection extends BestRand to jointly optimize multiple metrics.

Tables~\ref{tab:bench2drive-five-model-results} and~\ref{tab:fail2drive-compression-results} show that BestRand-SR is the strongest single-metric method on both benchmarks. Compared with Random, it reduces Avg. Error from 1.99 to 1.89 at 50\% removal and from 2.84 to 2.46 at 65\% removal on Bench2Drive. On Fail2Drive, it reduces the error from 2.56 to 2.14 and from 3.38 to 2.88, respectively. It also improves success-rate estimation at both compression levels on both benchmarks. These results demonstrate that lightweight probe matching transfers effectively to closed-loop VLA evaluation across benchmarks and evaluation splits.

\begin{table*}[t]
\centering
\caption{Bench2Drive compression results using five training models. Within each run, all train-dependent methods use the same SR-stratified model sample. We report percentage-point MAE for SR and Smoothness, score MAE for DS, and mean relative error in percent for Efficiency. Avg. Error averages these four errors. Lower is better. Standard deviations appear in smaller type. The best and second-best means in each removal-ratio block are \textbf{bolded} and \underline{underlined}, respectively.}
\vspace{4pt}
\label{tab:bench2drive-five-model-results}
\footnotesize
\setlength{\tabcolsep}{3pt}
\renewcommand{\arraystretch}{1.08}
\newcommand{\btdstd}[1]{\,{\scriptsize\color{black!70}#1}}
\newcommand{\btdgroup}[1]{\multicolumn{6}{l}{\textcolor{blue!45!black}{\textbf{#1}}}}
\begin{tabularx}{\textwidth}{@{}l*{5}{>{\raggedleft\arraybackslash}X}@{}}
\toprule
& \multicolumn{5}{c}{\textbf{Prediction error $\downarrow$}\enspace {\scriptsize mean $\pm$ std.}} \\
\cmidrule(lr){2-6}
\textbf{Method} & \textbf{SR} & \textbf{DS} & \textbf{Eff. MRE} & \textbf{Smooth.} & \textbf{Avg. Error} \\
\midrule
\btdgroup{Bench2Drive, removal ratio $=50\%$} \\
SparseEval (Linear) & 16.14\btdstd{$\pm$0.24} & 1.22\btdstd{$\pm$0.16} & 2.73\btdstd{$\pm$1.73} & \underline{1.38}\btdstd{$\pm$0.31} & 5.37\btdstd{$\pm$0.49} \\
SparseEval (MLP) & 3.50\btdstd{$\pm$0.15} & 1.78\btdstd{$\pm$0.19} & 2.31\btdstd{$\pm$2.13} & 1.85\btdstd{$\pm$0.73} & 2.36\btdstd{$\pm$0.37} \\
Anchor Points & 5.21\btdstd{$\pm$1.48} & 4.55\btdstd{$\pm$1.62} & 2.40\btdstd{$\pm$1.71} & 3.44\btdstd{$\pm$2.15} & 3.90\btdstd{$\pm$1.38} \\
TinyBenchmarks & 2.15\btdstd{$\pm$0.12} & 2.38\btdstd{$\pm$0.50} & 3.80\btdstd{$\pm$3.18} & 2.02\btdstd{$\pm$0.62} & 2.58\btdstd{$\pm$0.83} \\
ZipBench & 2.45\btdstd{$\pm$0.26} & 3.18\btdstd{$\pm$0.31} & \textbf{1.64}\btdstd{$\pm$0.33} & 1.92\btdstd{$\pm$0.47} & 2.30\btdstd{$\pm$0.08} \\
\addlinespace[1pt]
Random & 2.34\btdstd{$\pm$0.64} & 1.14\btdstd{$\pm$0.24} & 2.74\btdstd{$\pm$1.67} & 1.72\btdstd{$\pm$0.73} & 1.99\btdstd{$\pm$0.49} \\
\rowcolor{lightBlue}\textbf{BestRand-SR (ours)} & \textbf{1.94}\btdstd{$\pm$0.29} & \underline{1.03}\btdstd{$\pm$0.15} & 3.05\btdstd{$\pm$1.96} & 1.55\btdstd{$\pm$0.54} & \underline{1.89}\btdstd{$\pm$0.51} \\
\rowcolor{questionBackground}\textbf{BestRand-4Metric (ours)} & \underline{1.95}\btdstd{$\pm$0.32} & \textbf{1.00}\btdstd{$\pm$0.17} & \underline{1.69}\btdstd{$\pm$0.71} & \textbf{1.26}\btdstd{$\pm$0.25} & \textbf{1.47}\btdstd{$\pm$0.21} \\
\midrule
\btdgroup{Bench2Drive, removal ratio $=65\%$} \\
SparseEval (Linear) & 20.49\btdstd{$\pm$1.87} & 2.18\btdstd{$\pm$0.29} & 3.19\btdstd{$\pm$1.36} & 1.74\btdstd{$\pm$0.53} & 6.90\btdstd{$\pm$0.37} \\
SparseEval (MLP) & 2.85\btdstd{$\pm$0.40} & 2.61\btdstd{$\pm$1.21} & 3.88\btdstd{$\pm$1.32} & \textbf{1.40}\btdstd{$\pm$0.10} & 2.68\btdstd{$\pm$0.54} \\
Anchor Points & 5.80\btdstd{$\pm$1.58} & 4.11\btdstd{$\pm$0.95} & \underline{2.33}\btdstd{$\pm$1.02} & 3.78\btdstd{$\pm$2.63} & 4.00\btdstd{$\pm$0.95} \\
TinyBenchmarks & 3.01\btdstd{$\pm$0.05} & 2.86\btdstd{$\pm$0.68} & 4.67\btdstd{$\pm$3.47} & 2.10\btdstd{$\pm$0.24} & 3.16\btdstd{$\pm$0.89} \\
ZipBench & 3.12\btdstd{$\pm$0.50} & 3.91\btdstd{$\pm$0.83} & 2.54\btdstd{$\pm$0.92} & 2.63\btdstd{$\pm$0.25} & 3.05\btdstd{$\pm$0.41} \\
\addlinespace[1pt]
Random & 3.38\btdstd{$\pm$1.20} & 1.66\btdstd{$\pm$0.54} & 3.99\btdstd{$\pm$2.74} & 2.33\btdstd{$\pm$0.98} & 2.84\btdstd{$\pm$0.92} \\
\rowcolor{lightBlue}\textbf{BestRand-SR (ours)} & \underline{2.63}\btdstd{$\pm$0.44} & \underline{1.42}\btdstd{$\pm$0.26} & 3.78\btdstd{$\pm$2.28} & 2.03\btdstd{$\pm$0.74} & \underline{2.46}\btdstd{$\pm$0.63} \\
\rowcolor{questionBackground}\textbf{BestRand-4Metric (ours)} & \textbf{2.63}\btdstd{$\pm$0.44} & \textbf{1.36}\btdstd{$\pm$0.25} & \textbf{2.30}\btdstd{$\pm$0.95} & \underline{1.70}\btdstd{$\pm$0.36} & \textbf{2.00}\btdstd{$\pm$0.31} \\
\bottomrule
\end{tabularx}
\end{table*}

\subsection{Effectiveness of BestRand on Multi-Metric Benchmarks}
\label{app:bestrand-multi-metric-effectiveness}

Appendix~\ref{app:benchmark-details} summarizes all benchmarks used in our study. The main experiments use one metric per benchmark, selecting a representative metric when multiple metrics are available. Here, we examine whether BestRand can extend to multi-metric settings and preserve several metrics simultaneously. We normalize each metric separately and jointly match their probe profiles. BestRand-4Metric uses success rate, driving score, efficiency, and smoothness on Bench2Drive. BestRand-2Metric uses success rate and driving score on Fail2Drive, with the harmonic mean used only for evaluation. This extension changes only the selection objective and retains the same subset size.

Joint metric matching achieves the lowest overall error on both benchmarks. Compared with BestRand-SR, BestRand-4Metric reduces Avg. Error from 1.89 to 1.47 at 50\% removal and from 2.46 to 2.00 at 65\% removal on Bench2Drive. The largest gains are in efficiency and smoothness, while success-rate accuracy is maintained. On Fail2Drive, BestRand-2Metric reduces Avg. Error from 2.14 to 2.09 and from 2.88 to 2.83. Although smaller, these gains are consistent across compression levels and maintain balanced in-distribution and generalization performance. These results show that one compressed subset can accurately preserve multiple benchmark metrics.

\begin{table*}[t]
\centering
\caption{Fail2Drive paired-route compression results using five uniformly sampled training records. The same selected route pairs are used for ID and generalization evaluation. Random and BestRand use 100 seeds, while fitted baselines use three. We report test-record MAE as mean $\pm$ standard deviation. Lower is better. Standard deviations appear in smaller type. The best and second-best means in each removal-ratio block are \textbf{bolded} and \underline{underlined}, respectively.}
\vspace{4pt}
\label{tab:fail2drive-compression-results}
\footnotesize
\setlength{\tabcolsep}{2pt}
\renewcommand{\arraystretch}{1.08}
\newcommand{\fstd}[1]{\hspace{0.25pt}{\scriptsize\color{black!70}#1}}
\newcommand{\fgroup}[1]{\multicolumn{8}{l}{\textcolor{blue!45!black}{\textbf{#1}}}}
\begin{tabular*}{\textwidth}{@{\extracolsep{\fill}}lrrrrrrr@{}}
\toprule
& \multicolumn{7}{c}{\textbf{Prediction error $\downarrow$}\enspace {\scriptsize mean $\pm$ std.}} \\
\cmidrule(lr){2-8}
\textbf{Method} & \textbf{ID SR} & \textbf{ID DS} & \textbf{ID HM} & \textbf{Gen SR} & \textbf{Gen DS} & \textbf{Gen HM} & \mbox{\textbf{Avg. Error}} \\
\midrule
\fgroup{Fail2Drive, removal ratio $=50\%$} \\
SparseEval (Linear) & 17.40\fstd{$\pm$8.19} & 4.63\fstd{$\pm$0.62} & 10.44\fstd{$\pm$7.40} & 18.48\fstd{$\pm$2.77} & 5.56\fstd{$\pm$0.78} & 13.17\fstd{$\pm$4.26} & 11.62\fstd{$\pm$3.38} \\
SparseEval (MLP) & 4.56\fstd{$\pm$0.67} & 3.53\fstd{$\pm$0.95} & 3.07\fstd{$\pm$0.66} & 6.71\fstd{$\pm$2.20} & 3.86\fstd{$\pm$0.55} & 3.17\fstd{$\pm$0.81} & 4.15\fstd{$\pm$0.36} \\
Anchor Points & 2.79\fstd{$\pm$0.09} & 2.51\fstd{$\pm$0.63} & 2.09\fstd{$\pm$0.43} & 5.33\fstd{$\pm$1.30} & 3.99\fstd{$\pm$0.42} & 3.88\fstd{$\pm$1.18} & 3.43\fstd{$\pm$0.39} \\
TinyBenchmarks & 2.52\fstd{$\pm$0.23} & 1.81\fstd{$\pm$0.49} & 1.89\fstd{$\pm$0.28} & 3.49\fstd{$\pm$0.93} & \textbf{1.63}\fstd{$\pm$0.58} & 2.57\fstd{$\pm$0.56} & 2.32\fstd{$\pm$0.23} \\
ZipBench & 2.43\fstd{$\pm$0.35} & 3.06\fstd{$\pm$1.17} & 2.17\fstd{$\pm$0.60} & \textbf{2.87}\fstd{$\pm$0.37} & 2.51\fstd{$\pm$0.89} & \textbf{2.42}\fstd{$\pm$0.29} & 2.58\fstd{$\pm$0.39} \\
\addlinespace[1pt]
Random & 2.69\fstd{$\pm$0.71} & 1.66\fstd{$\pm$0.58} & 2.14\fstd{$\pm$0.63} & 3.80\fstd{$\pm$1.06} & 2.02\fstd{$\pm$0.61} & 3.04\fstd{$\pm$0.88} & 2.56\fstd{$\pm$0.56} \\
\rowcolor{lightBlue}\textbf{BestRand-SR (ours)} & \underline{2.34}\fstd{$\pm$0.50} & \underline{1.41}\fstd{$\pm$0.40} & \underline{1.80}\fstd{$\pm$0.41} & \underline{3.08}\fstd{$\pm$0.58} & 1.74\fstd{$\pm$0.43} & \underline{2.46}\fstd{$\pm$0.47} & \underline{2.14}\fstd{$\pm$0.32} \\
\rowcolor{questionBackground}\textbf{BestRand-2Metric (ours)} & \textbf{2.30}\fstd{$\pm$0.41} & \textbf{1.27}\fstd{$\pm$0.23} & \textbf{1.72}\fstd{$\pm$0.29} & 3.14\fstd{$\pm$0.64} & \underline{1.65}\fstd{$\pm$0.38} & 2.48\fstd{$\pm$0.51} & \textbf{2.09}\fstd{$\pm$0.29} \\
\midrule
\fgroup{Fail2Drive, removal ratio $=65\%$} \\
SparseEval (Linear) & 19.99\fstd{$\pm$9.27} & 5.24\fstd{$\pm$2.08} & 13.30\fstd{$\pm$8.77} & 18.93\fstd{$\pm$2.93} & 4.59\fstd{$\pm$0.77} & 13.01\fstd{$\pm$5.05} & 12.51\fstd{$\pm$3.89} \\
SparseEval (MLP) & 5.62\fstd{$\pm$1.43} & 4.58\fstd{$\pm$1.24} & 4.19\fstd{$\pm$0.75} & 5.48\fstd{$\pm$1.22} & 3.66\fstd{$\pm$0.51} & \textbf{3.24}\fstd{$\pm$0.23} & 4.46\fstd{$\pm$0.11} \\
Anchor Points & 3.60\fstd{$\pm$0.15} & 4.27\fstd{$\pm$0.95} & 3.37\fstd{$\pm$0.27} & 5.00\fstd{$\pm$1.75} & 4.34\fstd{$\pm$1.08} & 3.32\fstd{$\pm$0.47} & 3.98\fstd{$\pm$0.43} \\
TinyBenchmarks & 3.39\fstd{$\pm$0.45} & 2.79\fstd{$\pm$1.17} & 2.81\fstd{$\pm$0.86} & 5.50\fstd{$\pm$0.90} & 2.62\fstd{$\pm$0.65} & 4.24\fstd{$\pm$0.83} & 3.56\fstd{$\pm$0.71} \\
ZipBench & \underline{3.20}\fstd{$\pm$0.75} & 3.32\fstd{$\pm$1.64} & 2.70\fstd{$\pm$1.12} & 4.54\fstd{$\pm$0.59} & 3.25\fstd{$\pm$0.71} & 3.41\fstd{$\pm$0.39} & 3.40\fstd{$\pm$0.68} \\
\addlinespace[1pt]
Random & 3.76\fstd{$\pm$1.04} & 2.22\fstd{$\pm$0.80} & 2.94\fstd{$\pm$0.89} & 4.88\fstd{$\pm$1.19} & 2.60\fstd{$\pm$0.71} & 3.89\fstd{$\pm$1.01} & 3.38\fstd{$\pm$0.71} \\
\rowcolor{lightBlue}\textbf{BestRand-SR (ours)} & 3.20\fstd{$\pm$0.55} & \underline{1.82}\fstd{$\pm$0.42} & \underline{2.42}\fstd{$\pm$0.45} & \underline{4.23}\fstd{$\pm$0.89} & \underline{2.26}\fstd{$\pm$0.48} & 3.33\fstd{$\pm$0.71} & \underline{2.88}\fstd{$\pm$0.37} \\
\rowcolor{questionBackground}\textbf{BestRand-2Metric (ours)} & \textbf{3.18}\fstd{$\pm$0.51} & \textbf{1.75}\fstd{$\pm$0.34} & \textbf{2.39}\fstd{$\pm$0.39} & \textbf{4.20}\fstd{$\pm$0.82} & \textbf{2.19}\fstd{$\pm$0.46} & \underline{3.31}\fstd{$\pm$0.68} & \textbf{2.83}\fstd{$\pm$0.36} \\
\bottomrule
\end{tabular*}
\end{table*}

\subsection{Can Data-Only Methods Achieve Effective Benchmark Compression?}
\label{app:data-only-methods-effectiveness}

Some benchmark compression methods require no historical model responses and select items solely from representations of the benchmark data. This makes subset construction inexpensive and practical when performance records are unavailable. We study four such methods. Facility Location \citep{yao2026coresets} selects items that cover the prompt embedding space, while MIS \citep{kjorvezir2026consistent} constructs a similarity graph and chooses a nonredundant independent set. Agreement selects prompts with the greatest variation in attention weights, while Coverage iteratively selects distant prompts to span diverse regions of the representation space \citep{purificato2026select}. Appendix~\ref{app:baseline-details} describes these methods, and Appendix~\ref{app:implementation-hyperparameters} provides their implementation settings.

\begin{figure*}[t]
\centering
\includegraphics[width=\textwidth,keepaspectratio]{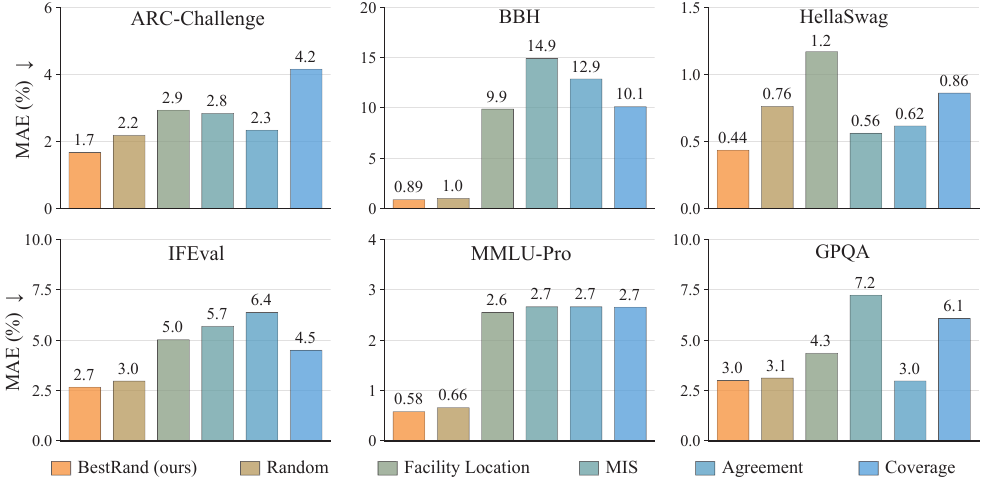}
\vspace{-18pt}
\caption{Performance of data-only compression methods at 20\% retention on six representative benchmarks. Facility Location, MIS, Agreement, and Coverage select items from prompt-derived representations without historical model responses. Random also requires no historical responses, while BestRand evaluates a few probe models only on its candidate pool. Lower MAE is better.}
\label{fig:data-only}
\end{figure*}

Figure~\ref{fig:data-only} compares the four data-only methods with Random and BestRand at 20\% retention. Although inexpensive to construct, the data-only subsets often yield substantially higher MAE than Random and BestRand, with performance varying markedly across benchmarks. BestRand achieves the lowest MAE on five of the six benchmarks and is nearly tied for the lowest on GPQA, yielding the strongest overall accuracy. These results show that a small initial investment in probe-model inference clearly improves accuracy over methods that rely only on benchmark data.

\section{Additional Statements}
\label{app:additional-statements}

\subsection{Limitations of BestRand}
\label{app:limitations-of-bestrand}

BestRand provides an efficient approximation to full evaluation, but several limitations remain.

\paragraph{Nonzero estimation error.}
Although BestRand outperforms many existing benchmark compression methods at lower construction cost, it cannot guarantee zero estimation error. This is a common limitation of benchmark compression, but the compression ratio can be adjusted to balance evaluation efficiency and reliability. BestRand is most useful for rapid model iteration and comparison, but compressed benchmarks cannot fully replace comprehensive evaluation. We therefore recommend reporting full-benchmark results for final evaluations when feasible.

\paragraph{Probe inference cost.}
BestRand still requires probe models to run full inference on some benchmark items, so its construction cost is not zero. However, this cost is small in practice and brings clear accuracy gains. BestRand substantially outperforms data-only methods that use no model responses. Moreover, only a few probe models are needed, and they score only the candidate pool, which can be much smaller than the full benchmark at low retention ratios. BestRand therefore achieves better accuracy at a lower construction cost than existing response-based methods.

\subsection{Broader Impact of BestRand}
\label{app:broader-impact}

BestRand provides stable, accurate, and low-cost benchmark compression, reducing evaluation costs while preserving reliable results. Its efficient construction process also keeps the cost of creating compressed benchmarks low. Together, these benefits can reduce compute and energy use, lower associated carbon emissions, and support faster experimental iteration. By making reliable evaluation more affordable, BestRand can also lower barriers for smaller laboratories and researchers with limited computing resources, enabling broader participation in model development and evaluation.

\subsection{Computational Resources Statement}
\label{app:computational-resources-statement}
Our experiments use historical model performance records collected through prior research, so their original computational cost cannot be reliably estimated. All benchmark compression experiments can run on a single GPU. BestRand's subset selection requires only a CPU and completes in under 0.1 seconds with parallel execution. We will release the code and data to support reproducibility.

\section{Additional Experimental Results}
\label{app:additional-results}

The main experiments in Section~\ref{sec:experiments} evaluate BestRand on 70 text, multimodal, and agent benchmarks and report results averaged across benchmarks. This section complements these aggregate findings with detailed results on representative benchmarks. We first report benchmark-level results for Spearman's rank correlation coefficient (SP) and mean absolute error (MAE). Finally, we provide additional visualizations that complement and extend the figures in the main text.

\subsection{Benchmark-Wise Results for Spearman Rank Correlation}
\label{app:benchmark-wise-spearman-results}
Table~\ref{tab:category-main-results-sp} summarizes Spearman rank correlation (SP) across all 70 benchmarks, while Tables~\ref{tab:representative-sp-text}, \ref{tab:representative-sp-multimodal}, and~\ref{tab:representative-sp-agent} report results for six representative benchmarks in each category. SP quantifies agreement between model rankings derived from compressed and full benchmarks. Higher values indicate better rank preservation, with $1$ denoting identical rankings, but do not guarantee small score errors. Leading methods achieve high SP at moderate and high retention ratios. Averaged across TinyBenchmarks, ZipBench, Random, and BestRand over all benchmarks, SP rises from $0.930$ at 20\% retention to $0.969$ at 50\%. This strong rank preservation, also observed in prior work~\citep{yauney2026reliable}, motivates our use of MAE as the primary metric. Full SP results are reported here.

\subsection{Benchmark-Wise Detailed Results for MAE}
\label{app:benchmark-wise-mae-results}

Tables~\ref{tab:representative-mae-text}, \ref{tab:representative-mae-multimodal}, and~\ref{tab:representative-mae-agent} provide detailed MAE results for six representative text, multimodal, and agent benchmarks, respectively, using the same 18 benchmarks as the SP tables.

\begin{table*}[t]
\centering
\caption{Spearman rank correlation across 70 benchmarks at 5--50\% retention. Values are mean $\pm$ sample standard deviation across benchmarks, and the Mean column pools all displayed settings. Higher is better. Best means are \textbf{bold} and second-best means are \underline{underlined}.}
\vspace{4pt}
\label{tab:category-main-results-sp}
\footnotesize
\setlength{\tabcolsep}{2.8pt}
\renewcommand{\arraystretch}{1.08}
\newcommand{\spmacrostd}[1]{\,{\scriptsize\color{black!70}#1}}
\newcommand{\spcatrow}[2]{\multicolumn{7}{l}{\textcolor{blue!45!black}{\textbf{#1}}\enspace\textcolor{black!70}{#2 benchmarks}}}
\resizebox{\textwidth}{!}{%
\begin{tabular}{lrrrrrr}
\toprule
& \multicolumn{6}{c}{\textbf{Average Spearman correlation $\uparrow$}\enspace {\scriptsize mean $\pm$ std.}} \\
\cmidrule(lr){2-7}
\rowcolor{lightBlue}\textbf{Method} & \textbf{5\%} & \textbf{10\%} & \textbf{15\%} & \textbf{20\%} & \textbf{50\%} & \textbf{Mean} \\
\midrule
\spcatrow{Text}{44} \\
SparseEval (Linear) & 0.381\spmacrostd{$\pm$0.238} & 0.364\spmacrostd{$\pm$0.252} & 0.384\spmacrostd{$\pm$0.242} & 0.347\spmacrostd{$\pm$0.271} & 0.407\spmacrostd{$\pm$0.265} & 0.377\spmacrostd{$\pm$0.252} \\
SparseEval (MLP) & 0.466\spmacrostd{$\pm$0.288} & 0.514\spmacrostd{$\pm$0.263} & 0.586\spmacrostd{$\pm$0.263} & 0.588\spmacrostd{$\pm$0.303} & 0.638\spmacrostd{$\pm$0.308} & 0.558\spmacrostd{$\pm$0.290} \\
Anchor Points & 0.756\spmacrostd{$\pm$0.190} & 0.791\spmacrostd{$\pm$0.165} & 0.801\spmacrostd{$\pm$0.165} & 0.800\spmacrostd{$\pm$0.172} & 0.862\spmacrostd{$\pm$0.146} & 0.802\spmacrostd{$\pm$0.170} \\
TinyBenchmarks & 0.841\spmacrostd{$\pm$0.181} & 0.883\spmacrostd{$\pm$0.152} & 0.900\spmacrostd{$\pm$0.164} & 0.932\spmacrostd{$\pm$0.100} & 0.966\spmacrostd{$\pm$0.053} & 0.904\spmacrostd{$\pm$0.143} \\
ZipBench & 0.837\spmacrostd{$\pm$0.193} & 0.890\spmacrostd{$\pm$0.151} & 0.901\spmacrostd{$\pm$0.163} & 0.927\spmacrostd{$\pm$0.118} & 0.967\spmacrostd{$\pm$0.059} & 0.904\spmacrostd{$\pm$0.149} \\
Random & \underline{0.855}\spmacrostd{$\pm$0.170} & \underline{0.903}\spmacrostd{$\pm$0.129} & \underline{0.926}\spmacrostd{$\pm$0.103} & \underline{0.941}\spmacrostd{$\pm$0.084} & \underline{0.978}\spmacrostd{$\pm$0.031} & \underline{0.921}\spmacrostd{$\pm$0.120} \\
\rowcolor{lightBlue}\textbf{BestRand (ours)} & \textbf{0.857}\spmacrostd{$\pm$0.167} & \textbf{0.904}\spmacrostd{$\pm$0.129} & \textbf{0.927}\spmacrostd{$\pm$0.101} & \textbf{0.942}\spmacrostd{$\pm$0.083} & \textbf{0.978}\spmacrostd{$\pm$0.031} & \textbf{0.922}\spmacrostd{$\pm$0.118} \\
\midrule
\spcatrow{Multimodal}{16} \\
SparseEval (Linear) & 0.327\spmacrostd{$\pm$0.319} & 0.252\spmacrostd{$\pm$0.329} & 0.256\spmacrostd{$\pm$0.228} & 0.254\spmacrostd{$\pm$0.319} & 0.274\spmacrostd{$\pm$0.237} & 0.272\spmacrostd{$\pm$0.284} \\
SparseEval (MLP) & 0.363\spmacrostd{$\pm$0.377} & 0.576\spmacrostd{$\pm$0.225} & 0.506\spmacrostd{$\pm$0.267} & 0.588\spmacrostd{$\pm$0.280} & 0.607\spmacrostd{$\pm$0.324} & 0.528\spmacrostd{$\pm$0.305} \\
Anchor Points & 0.772\spmacrostd{$\pm$0.193} & 0.749\spmacrostd{$\pm$0.186} & 0.757\spmacrostd{$\pm$0.206} & 0.774\spmacrostd{$\pm$0.183} & 0.842\spmacrostd{$\pm$0.160} & 0.779\spmacrostd{$\pm$0.185} \\
TinyBenchmarks & 0.818\spmacrostd{$\pm$0.202} & 0.885\spmacrostd{$\pm$0.142} & 0.922\spmacrostd{$\pm$0.081} & 0.930\spmacrostd{$\pm$0.089} & 0.961\spmacrostd{$\pm$0.044} & 0.903\spmacrostd{$\pm$0.131} \\
ZipBench & 0.795\spmacrostd{$\pm$0.311} & 0.868\spmacrostd{$\pm$0.181} & 0.917\spmacrostd{$\pm$0.114} & 0.940\spmacrostd{$\pm$0.055} & 0.963\spmacrostd{$\pm$0.051} & 0.897\spmacrostd{$\pm$0.178} \\
Random & \textbf{0.850}\spmacrostd{$\pm$0.145} & \textbf{0.903}\spmacrostd{$\pm$0.106} & \underline{0.927}\spmacrostd{$\pm$0.084} & \underline{0.941}\spmacrostd{$\pm$0.070} & \underline{0.976}\spmacrostd{$\pm$0.032} & \textbf{0.919}\spmacrostd{$\pm$0.102} \\
\rowcolor{lightBlue}\textbf{BestRand (ours)} & \underline{0.840}\spmacrostd{$\pm$0.168} & \underline{0.902}\spmacrostd{$\pm$0.106} & \textbf{0.927}\spmacrostd{$\pm$0.084} & \textbf{0.941}\spmacrostd{$\pm$0.069} & \textbf{0.976}\spmacrostd{$\pm$0.033} & \underline{0.917}\spmacrostd{$\pm$0.110} \\
\midrule
\spcatrow{Agent}{10} \\
SparseEval (Linear) & 0.275\spmacrostd{$\pm$0.157} & 0.243\spmacrostd{$\pm$0.226} & 0.326\spmacrostd{$\pm$0.221} & 0.339\spmacrostd{$\pm$0.212} & 0.254\spmacrostd{$\pm$0.260} & 0.288\spmacrostd{$\pm$0.212} \\
SparseEval (MLP) & 0.230\spmacrostd{$\pm$0.292} & 0.330\spmacrostd{$\pm$0.331} & 0.316\spmacrostd{$\pm$0.274} & 0.454\spmacrostd{$\pm$0.243} & 0.554\spmacrostd{$\pm$0.292} & 0.377\spmacrostd{$\pm$0.299} \\
Anchor Points & 0.678\spmacrostd{$\pm$0.176} & 0.726\spmacrostd{$\pm$0.121} & 0.750\spmacrostd{$\pm$0.127} & 0.724\spmacrostd{$\pm$0.160} & 0.860\spmacrostd{$\pm$0.101} & 0.748\spmacrostd{$\pm$0.147} \\
TinyBenchmarks & 0.727\spmacrostd{$\pm$0.179} & 0.815\spmacrostd{$\pm$0.134} & 0.851\spmacrostd{$\pm$0.116} & 0.878\spmacrostd{$\pm$0.101} & 0.943\spmacrostd{$\pm$0.049} & 0.843\spmacrostd{$\pm$0.138} \\
ZipBench & 0.752\spmacrostd{$\pm$0.161} & 0.829\spmacrostd{$\pm$0.124} & \underline{0.880}\spmacrostd{$\pm$0.094} & 0.891\spmacrostd{$\pm$0.088} & 0.935\spmacrostd{$\pm$0.054} & 0.858\spmacrostd{$\pm$0.123} \\
Random & \textbf{0.759}\spmacrostd{$\pm$0.166} & \underline{0.842}\spmacrostd{$\pm$0.118} & 0.880\spmacrostd{$\pm$0.093} & \underline{0.905}\spmacrostd{$\pm$0.075} & \underline{0.965}\spmacrostd{$\pm$0.028} & \underline{0.870}\spmacrostd{$\pm$0.123} \\
\rowcolor{lightBlue}\textbf{BestRand (ours)} & \underline{0.755}\spmacrostd{$\pm$0.173} & \textbf{0.843}\spmacrostd{$\pm$0.118} & \textbf{0.881}\spmacrostd{$\pm$0.093} & \textbf{0.905}\spmacrostd{$\pm$0.075} & \textbf{0.965}\spmacrostd{$\pm$0.028} & \textbf{0.870}\spmacrostd{$\pm$0.126} \\
\bottomrule
\end{tabular}
}%
\end{table*}

\begin{table*}[t]
\centering
\caption{Text benchmark Spearman correlations at 5--50\% retention. Values are mean $\pm$ sample standard deviation; higher is better. Best means are \textbf{bold} and second-best means are \underline{underlined}.}
\vspace{4pt}
\label{tab:representative-sp-text}
\footnotesize
\setlength{\tabcolsep}{2.8pt}
\renewcommand{\arraystretch}{1.00}
\newcommand{\spstd}[1]{\,{\scriptsize\color{black!70}#1}}
\newcommand{\benchrow}[1]{\multicolumn{7}{l}{\textcolor{blue!45!black}{\textbf{#1}}}}
\begin{tabular}{lrrrrrr}
\toprule
& \multicolumn{6}{c}{\textbf{Average Spearman correlation $\uparrow$}\enspace {\scriptsize mean $\pm$ std.}} \\
\cmidrule(lr){2-7}
\rowcolor{lightBlue}\textbf{Method} & \textbf{5\%} & \textbf{10\%} & \textbf{15\%} & \textbf{20\%} & \textbf{50\%} & \textbf{Mean} \\
\midrule
\benchrow{TruthfulQA} \\
SparseEval (Linear) & -0.002\spstd{$\pm$0.420} & 0.627\spstd{$\pm$0.227} & 0.672\spstd{$\pm$0.174} & 0.697\spstd{$\pm$0.109} & 0.729\spstd{$\pm$0.071} & 0.545\spstd{$\pm$0.347} \\
SparseEval (MLP) & -0.145\spstd{$\pm$0.776} & 0.255\spstd{$\pm$0.951} & 0.415\spstd{$\pm$0.554} & 0.522\spstd{$\pm$0.364} & 0.684\spstd{$\pm$0.303} & 0.346\spstd{$\pm$0.614} \\
Anchor Points & 0.884\spstd{$\pm$0.010} & \textbf{0.974}\spstd{$\pm$0.010} & 0.905\spstd{$\pm$0.054} & 0.974\spstd{$\pm$0.006} & 0.974\spstd{$\pm$0.006} & 0.942\spstd{$\pm$0.046} \\
TinyBenchmarks & 0.896\spstd{$\pm$0.075} & 0.925\spstd{$\pm$0.042} & 0.959\spstd{$\pm$0.014} & 0.970\spstd{$\pm$0.009} & 0.981\spstd{$\pm$0.013} & 0.946\spstd{$\pm$0.047} \\
ZipBench & \textbf{0.946}\spstd{$\pm$0.045} & 0.958\spstd{$\pm$0.034} & 0.970\spstd{$\pm$0.008} & 0.958\spstd{$\pm$0.012} & 0.990\spstd{$\pm$0.005} & 0.965\spstd{$\pm$0.027} \\
Random & 0.930\spstd{$\pm$0.036} & 0.959\spstd{$\pm$0.021} & \underline{0.970}\spstd{$\pm$0.015} & \underline{0.977}\spstd{$\pm$0.012} & \underline{0.992}\spstd{$\pm$0.004} & \underline{0.966}\spstd{$\pm$0.029} \\
\rowcolor{lightBlue}\textbf{BestRand (ours)} & \underline{0.933}\spstd{$\pm$0.033} & \underline{0.961}\spstd{$\pm$0.019} & \textbf{0.972}\spstd{$\pm$0.014} & \textbf{0.979}\spstd{$\pm$0.011} & \textbf{0.992}\spstd{$\pm$0.004} & \textbf{0.967}\spstd{$\pm$0.027} \\
\midrule
\benchrow{BoolQ} \\
SparseEval (Linear) & 0.263\spstd{$\pm$0.145} & 0.519\spstd{$\pm$0.197} & 0.497\spstd{$\pm$0.220} & 0.509\spstd{$\pm$0.325} & 0.278\spstd{$\pm$0.321} & 0.413\spstd{$\pm$0.245} \\
SparseEval (MLP) & 0.197\spstd{$\pm$0.203} & 0.294\spstd{$\pm$0.195} & 0.333\spstd{$\pm$0.512} & 0.489\spstd{$\pm$0.371} & 0.196\spstd{$\pm$0.633} & 0.302\spstd{$\pm$0.372} \\
Anchor Points & 0.407\spstd{$\pm$0.178} & 0.573\spstd{$\pm$0.218} & 0.784\spstd{$\pm$0.043} & 0.612\spstd{$\pm$0.356} & 0.790\spstd{$\pm$0.021} & 0.633\spstd{$\pm$0.228} \\
TinyBenchmarks & 0.903\spstd{$\pm$0.023} & 0.959\spstd{$\pm$0.014} & 0.966\spstd{$\pm$0.022} & 0.976\spstd{$\pm$0.012} & 0.993\spstd{$\pm$0.006} & 0.959\spstd{$\pm$0.034} \\
ZipBench & 0.855\spstd{$\pm$0.005} & 0.933\spstd{$\pm$0.032} & 0.976\spstd{$\pm$0.003} & 0.978\spstd{$\pm$0.007} & 0.986\spstd{$\pm$0.003} & 0.946\spstd{$\pm$0.052} \\
Random & \underline{0.934}\spstd{$\pm$0.032} & \textbf{0.965}\spstd{$\pm$0.017} & \underline{0.976}\spstd{$\pm$0.011} & \underline{0.982}\spstd{$\pm$0.009} & \underline{0.994}\spstd{$\pm$0.003} & \underline{0.970}\spstd{$\pm$0.027} \\
\rowcolor{lightBlue}\textbf{BestRand (ours)} & \textbf{0.934}\spstd{$\pm$0.032} & \underline{0.965}\spstd{$\pm$0.016} & \textbf{0.977}\spstd{$\pm$0.011} & \textbf{0.983}\spstd{$\pm$0.009} & \textbf{0.994}\spstd{$\pm$0.003} & \textbf{0.970}\spstd{$\pm$0.027} \\
\midrule
\benchrow{IFEval} \\
SparseEval (Linear) & 0.564\spstd{$\pm$0.140} & 0.670\spstd{$\pm$0.125} & 0.601\spstd{$\pm$0.188} & 0.667\spstd{$\pm$0.094} & 0.232\spstd{$\pm$0.559} & 0.547\spstd{$\pm$0.290} \\
SparseEval (MLP) & 0.387\spstd{$\pm$0.251} & 0.793\spstd{$\pm$0.116} & 0.812\spstd{$\pm$0.118} & 0.871\spstd{$\pm$0.089} & 0.889\spstd{$\pm$0.105} & 0.751\spstd{$\pm$0.229} \\
Anchor Points & 0.891\spstd{$\pm$0.018} & 0.912\spstd{$\pm$0.035} & 0.900\spstd{$\pm$0.050} & 0.888\spstd{$\pm$0.081} & 0.954\spstd{$\pm$0.011} & 0.909\spstd{$\pm$0.046} \\
TinyBenchmarks & 0.923\spstd{$\pm$0.020} & 0.949\spstd{$\pm$0.033} & 0.978\spstd{$\pm$0.003} & 0.980\spstd{$\pm$0.002} & 0.992\spstd{$\pm$0.003} & 0.965\spstd{$\pm$0.030} \\
ZipBench & 0.861\spstd{$\pm$0.029} & 0.949\spstd{$\pm$0.035} & 0.971\spstd{$\pm$0.015} & 0.982\spstd{$\pm$0.003} & 0.987\spstd{$\pm$0.007} & 0.950\spstd{$\pm$0.051} \\
Random & \underline{0.942}\spstd{$\pm$0.031} & \underline{0.970}\spstd{$\pm$0.016} & \underline{0.980}\spstd{$\pm$0.010} & \textbf{0.985}\spstd{$\pm$0.008} & \textbf{0.993}\spstd{$\pm$0.004} & \underline{0.974}\spstd{$\pm$0.024} \\
\rowcolor{lightBlue}\textbf{BestRand (ours)} & \textbf{0.944}\spstd{$\pm$0.028} & \textbf{0.970}\spstd{$\pm$0.015} & \textbf{0.980}\spstd{$\pm$0.010} & \underline{0.985}\spstd{$\pm$0.008} & \underline{0.993}\spstd{$\pm$0.004} & \textbf{0.974}\spstd{$\pm$0.023} \\
\midrule
\benchrow{Finance-Fundamentals} \\
SparseEval (Linear) & 0.409\spstd{$\pm$0.320} & 0.021\spstd{$\pm$0.510} & 0.303\spstd{$\pm$0.187} & 0.264\spstd{$\pm$0.154} & 0.040\spstd{$\pm$0.649} & 0.208\spstd{$\pm$0.381} \\
SparseEval (MLP) & 0.267\spstd{$\pm$0.242} & 0.369\spstd{$\pm$0.259} & 0.564\spstd{$\pm$0.136} & 0.645\spstd{$\pm$0.040} & 0.753\spstd{$\pm$0.091} & 0.520\spstd{$\pm$0.237} \\
Anchor Points & 0.757\spstd{$\pm$0.067} & 0.830\spstd{$\pm$0.050} & 0.822\spstd{$\pm$0.032} & 0.834\spstd{$\pm$0.072} & 0.940\spstd{$\pm$0.018} & 0.836\spstd{$\pm$0.075} \\
TinyBenchmarks & 0.796\spstd{$\pm$0.028} & 0.838\spstd{$\pm$0.028} & 0.923\spstd{$\pm$0.037} & 0.946\spstd{$\pm$0.019} & 0.973\spstd{$\pm$0.005} & 0.895\spstd{$\pm$0.073} \\
ZipBench & 0.819\spstd{$\pm$0.038} & 0.807\spstd{$\pm$0.087} & 0.900\spstd{$\pm$0.041} & 0.923\spstd{$\pm$0.014} & 0.972\spstd{$\pm$0.016} & 0.884\spstd{$\pm$0.076} \\
Random & \underline{0.822}\spstd{$\pm$0.084} & \underline{0.901}\spstd{$\pm$0.043} & \textbf{0.933}\spstd{$\pm$0.029} & \textbf{0.949}\spstd{$\pm$0.022} & \textbf{0.984}\spstd{$\pm$0.008} & \underline{0.918}\spstd{$\pm$0.071} \\
\rowcolor{lightBlue}\textbf{BestRand (ours)} & \textbf{0.825}\spstd{$\pm$0.080} & \textbf{0.903}\spstd{$\pm$0.041} & \underline{0.932}\spstd{$\pm$0.028} & \underline{0.949}\spstd{$\pm$0.022} & \underline{0.984}\spstd{$\pm$0.008} & \textbf{0.919}\spstd{$\pm$0.069} \\
\midrule
\benchrow{LIFBench} \\
SparseEval (Linear) & 0.657\spstd{$\pm$0.186} & 0.261\spstd{$\pm$0.635} & 0.210\spstd{$\pm$0.374} & 0.377\spstd{$\pm$0.375} & 0.288\spstd{$\pm$0.398} & 0.359\spstd{$\pm$0.390} \\
SparseEval (MLP) & 0.792\spstd{$\pm$0.051} & 0.827\spstd{$\pm$0.034} & 0.852\spstd{$\pm$0.030} & 0.807\spstd{$\pm$0.055} & 0.848\spstd{$\pm$0.021} & 0.825\spstd{$\pm$0.042} \\
Anchor Points & 0.926\spstd{$\pm$0.039} & 0.965\spstd{$\pm$0.039} & 0.962\spstd{$\pm$0.014} & 0.962\spstd{$\pm$0.018} & 0.986\spstd{$\pm$0.009} & 0.960\spstd{$\pm$0.030} \\
TinyBenchmarks & 0.932\spstd{$\pm$0.040} & 0.952\spstd{$\pm$0.020} & 0.952\spstd{$\pm$0.009} & 0.950\spstd{$\pm$0.028} & 0.958\spstd{$\pm$0.002} & 0.949\spstd{$\pm$0.022} \\
ZipBench & \underline{0.964}\spstd{$\pm$0.009} & \textbf{0.986}\spstd{$\pm$0.007} & \textbf{0.987}\spstd{$\pm$0.002} & \textbf{0.992}\spstd{$\pm$0.008} & 0.992\spstd{$\pm$0.004} & \textbf{0.984}\spstd{$\pm$0.012} \\
Random & 0.961\spstd{$\pm$0.024} & 0.976\spstd{$\pm$0.016} & 0.982\spstd{$\pm$0.012} & 0.985\spstd{$\pm$0.010} & \underline{0.992}\spstd{$\pm$0.006} & 0.979\spstd{$\pm$0.018} \\
\rowcolor{lightBlue}\textbf{BestRand (ours)} & \textbf{0.966}\spstd{$\pm$0.020} & \underline{0.979}\spstd{$\pm$0.014} & \underline{0.984}\spstd{$\pm$0.010} & \underline{0.987}\spstd{$\pm$0.009} & \textbf{0.993}\spstd{$\pm$0.006} & \underline{0.981}\spstd{$\pm$0.016} \\
\midrule
\benchrow{GPQA} \\
SparseEval (Linear) & 0.391\spstd{$\pm$0.138} & 0.187\spstd{$\pm$0.427} & 0.297\spstd{$\pm$0.364} & 0.113\spstd{$\pm$0.143} & -0.096\spstd{$\pm$0.366} & 0.178\spstd{$\pm$0.315} \\
SparseEval (MLP) & -0.103\spstd{$\pm$0.248} & -0.112\spstd{$\pm$0.361} & -0.078\spstd{$\pm$0.457} & 0.151\spstd{$\pm$0.025} & 0.608\spstd{$\pm$0.141} & 0.093\spstd{$\pm$0.376} \\
Anchor Points & 0.396\spstd{$\pm$0.351} & 0.268\spstd{$\pm$0.343} & 0.367\spstd{$\pm$0.109} & 0.397\spstd{$\pm$0.070} & 0.441\spstd{$\pm$0.099} & 0.374\spstd{$\pm$0.204} \\
TinyBenchmarks & 0.231\spstd{$\pm$0.121} & 0.474\spstd{$\pm$0.104} & \textbf{0.668}\spstd{$\pm$0.083} & \underline{0.722}\spstd{$\pm$0.070} & 0.819\spstd{$\pm$0.132} & 0.583\spstd{$\pm$0.233} \\
ZipBench & 0.092\spstd{$\pm$0.316} & \textbf{0.598}\spstd{$\pm$0.183} & 0.537\spstd{$\pm$0.048} & 0.531\spstd{$\pm$0.143} & 0.766\spstd{$\pm$0.112} & 0.505\spstd{$\pm$0.278} \\
Random & \underline{0.430}\spstd{$\pm$0.212} & 0.573\spstd{$\pm$0.171} & 0.658\spstd{$\pm$0.142} & 0.717\spstd{$\pm$0.120} & \underline{0.884}\spstd{$\pm$0.055} & \underline{0.652}\spstd{$\pm$0.212} \\
\rowcolor{lightBlue}\textbf{BestRand (ours)} & \textbf{0.445}\spstd{$\pm$0.196} & \underline{0.576}\spstd{$\pm$0.159} & \underline{0.662}\spstd{$\pm$0.131} & \textbf{0.725}\spstd{$\pm$0.111} & \textbf{0.886}\spstd{$\pm$0.052} & \textbf{0.659}\spstd{$\pm$0.202} \\
\bottomrule
\end{tabular}
\end{table*}

\clearpage

\begin{table*}[t]
\centering
\caption{Multimodal benchmark Spearman correlations at 5--50\% retention. Values are mean $\pm$ sample standard deviation; higher is better. Best means are \textbf{bold} and second-best are \underline{underlined}.}
\vspace{4pt}
\label{tab:representative-sp-multimodal}
\footnotesize
\setlength{\tabcolsep}{2.8pt}
\renewcommand{\arraystretch}{1.00}
\newcommand{\spstd}[1]{\,{\scriptsize\color{black!70}#1}}
\newcommand{\benchrow}[1]{\multicolumn{7}{l}{\textcolor{blue!45!black}{\textbf{#1}}}}
\begin{tabular}{lrrrrrr}
\toprule
& \multicolumn{6}{c}{\textbf{Average Spearman correlation $\uparrow$}\enspace {\scriptsize mean $\pm$ std.}} \\
\cmidrule(lr){2-7}
\rowcolor{lightBlue}\textbf{Method} & \textbf{5\%} & \textbf{10\%} & \textbf{15\%} & \textbf{20\%} & \textbf{50\%} & \textbf{Mean} \\
\midrule
\benchrow{MathVerse} \\
SparseEval (Linear) & 0.199\spstd{$\pm$0.372} & -0.318\spstd{$\pm$0.355} & -0.306\spstd{$\pm$0.128} & 0.043\spstd{$\pm$0.435} & 0.107\spstd{$\pm$0.411} & -0.055\spstd{$\pm$0.375} \\
SparseEval (MLP) & -0.177\spstd{$\pm$0.440} & 0.450\spstd{$\pm$0.423} & 0.618\spstd{$\pm$0.160} & 0.747\spstd{$\pm$0.048} & 0.445\spstd{$\pm$0.065} & 0.417\spstd{$\pm$0.407} \\
Anchor Points & 0.771\spstd{$\pm$0.142} & 0.783\spstd{$\pm$0.071} & 0.802\spstd{$\pm$0.114} & 0.846\spstd{$\pm$0.053} & 0.958\spstd{$\pm$0.010} & 0.832\spstd{$\pm$0.104} \\
TinyBenchmarks & 0.853\spstd{$\pm$0.054} & 0.936\spstd{$\pm$0.026} & 0.946\spstd{$\pm$0.016} & 0.962\spstd{$\pm$0.020} & \textbf{0.989}\spstd{$\pm$0.004} & 0.937\spstd{$\pm$0.053} \\
ZipBench & 0.897\spstd{$\pm$0.055} & 0.893\spstd{$\pm$0.027} & 0.942\spstd{$\pm$0.029} & 0.957\spstd{$\pm$0.009} & 0.987\spstd{$\pm$0.007} & 0.935\spstd{$\pm$0.045} \\
Random & \underline{0.907}\spstd{$\pm$0.043} & \textbf{0.948}\spstd{$\pm$0.022} & \textbf{0.963}\spstd{$\pm$0.016} & \underline{0.971}\spstd{$\pm$0.013} & \underline{0.988}\spstd{$\pm$0.006} & \underline{0.955}\spstd{$\pm$0.036} \\
\rowcolor{lightBlue}\textbf{BestRand (ours)} & \textbf{0.907}\spstd{$\pm$0.041} & \underline{0.948}\spstd{$\pm$0.022} & \underline{0.963}\spstd{$\pm$0.016} & \textbf{0.971}\spstd{$\pm$0.013} & 0.988\spstd{$\pm$0.006} & \textbf{0.955}\spstd{$\pm$0.036} \\
\midrule
\benchrow{OmniDocBench} \\
SparseEval (Linear) & -0.448\spstd{$\pm$0.315} & 0.105\spstd{$\pm$0.662} & 0.162\spstd{$\pm$0.493} & -0.181\spstd{$\pm$0.616} & -0.029\spstd{$\pm$0.674} & -0.078\spstd{$\pm$0.531} \\
SparseEval (MLP) & 0.543\spstd{$\pm$0.605} & 0.467\spstd{$\pm$0.436} & 0.619\spstd{$\pm$0.293} & 0.848\spstd{$\pm$0.066} & 0.581\spstd{$\pm$0.175} & 0.611\spstd{$\pm$0.338} \\
Anchor Points & \textbf{0.886}\spstd{$\pm$0.099} & 0.829\spstd{$\pm$0.000} & 0.829\spstd{$\pm$0.151} & 0.905\spstd{$\pm$0.066} & 0.905\spstd{$\pm$0.119} & 0.870\spstd{$\pm$0.093} \\
TinyBenchmarks & 0.638\spstd{$\pm$0.175} & 0.810\spstd{$\pm$0.144} & 0.733\spstd{$\pm$0.175} & 0.771\spstd{$\pm$0.099} & 0.886\spstd{$\pm$0.099} & 0.768\spstd{$\pm$0.147} \\
ZipBench & 0.829\spstd{$\pm$0.099} & \textbf{0.943}\spstd{$\pm$0.057} & 0.886\spstd{$\pm$0.057} & 0.924\spstd{$\pm$0.033} & 0.924\spstd{$\pm$0.087} & 0.901\spstd{$\pm$0.073} \\
Random & 0.831\spstd{$\pm$0.152} & 0.895\spstd{$\pm$0.091} & \underline{0.920}\spstd{$\pm$0.074} & \underline{0.936}\spstd{$\pm$0.064} & \underline{0.976}\spstd{$\pm$0.033} & \underline{0.912}\spstd{$\pm$0.103} \\
\rowcolor{lightBlue}\textbf{BestRand (ours)} & \underline{0.836}\spstd{$\pm$0.143} & \underline{0.898}\spstd{$\pm$0.089} & \textbf{0.922}\spstd{$\pm$0.073} & \textbf{0.937}\spstd{$\pm$0.064} & \textbf{0.976}\spstd{$\pm$0.032} & \textbf{0.914}\spstd{$\pm$0.099} \\
\midrule
\benchrow{SpatialEval} \\
SparseEval (Linear) & 0.776\spstd{$\pm$0.160} & 0.479\spstd{$\pm$0.464} & 0.462\spstd{$\pm$0.655} & 0.490\spstd{$\pm$0.665} & 0.445\spstd{$\pm$0.714} & 0.531\spstd{$\pm$0.498} \\
SparseEval (MLP) & 0.681\spstd{$\pm$0.218} & 0.843\spstd{$\pm$0.229} & 0.899\spstd{$\pm$0.044} & 0.888\spstd{$\pm$0.035} & 0.759\spstd{$\pm$0.245} & 0.814\spstd{$\pm$0.175} \\
Anchor Points & 0.978\spstd{$\pm$0.026} & 0.978\spstd{$\pm$0.026} & 0.989\spstd{$\pm$0.010} & 0.978\spstd{$\pm$0.026} & 0.989\spstd{$\pm$0.019} & 0.982\spstd{$\pm$0.020} \\
TinyBenchmarks & \textbf{0.989}\spstd{$\pm$0.010} & 0.961\spstd{$\pm$0.019} & 0.983\spstd{$\pm$0.029} & \textbf{1.000}\spstd{$\pm$0.000} & \textbf{1.000}\spstd{$\pm$0.000} & 0.987\spstd{$\pm$0.020} \\
ZipBench & \underline{0.983}\spstd{$\pm$0.017} & 0.961\spstd{$\pm$0.035} & 0.978\spstd{$\pm$0.026} & 0.978\spstd{$\pm$0.026} & \textbf{1.000}\spstd{$\pm$0.000} & 0.980\spstd{$\pm$0.024} \\
Random & 0.979\spstd{$\pm$0.023} & \underline{0.987}\spstd{$\pm$0.017} & \underline{0.991}\spstd{$\pm$0.014} & 0.993\spstd{$\pm$0.012} & 0.998\spstd{$\pm$0.005} & \textbf{0.990}\spstd{$\pm$0.017} \\
\rowcolor{lightBlue}\textbf{BestRand (ours)} & 0.974\spstd{$\pm$0.026} & \textbf{0.988}\spstd{$\pm$0.017} & \textbf{0.991}\spstd{$\pm$0.014} & \underline{0.993}\spstd{$\pm$0.012} & \underline{0.998}\spstd{$\pm$0.005} & \underline{0.989}\spstd{$\pm$0.018} \\
\midrule
\benchrow{LogicVista} \\
SparseEval (Linear) & -0.038\spstd{$\pm$0.282} & -0.008\spstd{$\pm$0.448} & 0.084\spstd{$\pm$0.312} & 0.168\spstd{$\pm$0.158} & 0.367\spstd{$\pm$0.200} & 0.115\spstd{$\pm$0.293} \\
SparseEval (MLP) & 0.175\spstd{$\pm$0.533} & 0.387\spstd{$\pm$0.289} & 0.412\spstd{$\pm$0.491} & 0.382\spstd{$\pm$0.514} & 0.869\spstd{$\pm$0.028} & 0.445\spstd{$\pm$0.425} \\
Anchor Points & \textbf{0.795}\spstd{$\pm$0.070} & 0.664\spstd{$\pm$0.042} & 0.829\spstd{$\pm$0.085} & 0.691\spstd{$\pm$0.247} & 0.866\spstd{$\pm$0.035} & 0.769\spstd{$\pm$0.132} \\
TinyBenchmarks & 0.729\spstd{$\pm$0.046} & \textbf{0.868}\spstd{$\pm$0.020} & 0.884\spstd{$\pm$0.034} & 0.902\spstd{$\pm$0.039} & 0.939\spstd{$\pm$0.021} & 0.865\spstd{$\pm$0.080} \\
ZipBench & 0.739\spstd{$\pm$0.154} & 0.777\spstd{$\pm$0.087} & 0.862\spstd{$\pm$0.028} & 0.882\spstd{$\pm$0.041} & 0.968\spstd{$\pm$0.020} & 0.846\spstd{$\pm$0.109} \\
Random & 0.763\spstd{$\pm$0.122} & 0.866\spstd{$\pm$0.065} & \textbf{0.905}\spstd{$\pm$0.043} & \underline{0.927}\spstd{$\pm$0.033} & \textbf{0.974}\spstd{$\pm$0.013} & \underline{0.887}\spstd{$\pm$0.098} \\
\rowcolor{lightBlue}\textbf{BestRand (ours)} & \underline{0.766}\spstd{$\pm$0.111} & \underline{0.867}\spstd{$\pm$0.060} & \underline{0.905}\spstd{$\pm$0.041} & \textbf{0.927}\spstd{$\pm$0.031} & \underline{0.973}\spstd{$\pm$0.013} & \textbf{0.888}\spstd{$\pm$0.093} \\
\midrule
\benchrow{RealWorldQA} \\
SparseEval (Linear) & 0.115\spstd{$\pm$0.580} & 0.200\spstd{$\pm$0.429} & 0.309\spstd{$\pm$0.073} & 0.333\spstd{$\pm$0.200} & 0.079\spstd{$\pm$0.431} & 0.207\spstd{$\pm$0.344} \\
SparseEval (MLP) & -0.358\spstd{$\pm$0.138} & 0.115\spstd{$\pm$0.606} & 0.285\spstd{$\pm$0.738} & 0.188\spstd{$\pm$0.207} & -0.248\spstd{$\pm$0.421} & -0.004\spstd{$\pm$0.483} \\
Anchor Points & 0.382\spstd{$\pm$0.601} & 0.721\spstd{$\pm$0.255} & 0.236\spstd{$\pm$0.419} & 0.398\spstd{$\pm$0.803} & 0.572\spstd{$\pm$0.418} & 0.462\spstd{$\pm$0.483} \\
TinyBenchmarks & 0.248\spstd{$\pm$0.704} & 0.673\spstd{$\pm$0.109} & \textbf{0.842}\spstd{$\pm$0.164} & \textbf{0.891}\spstd{$\pm$0.109} & 0.927\spstd{$\pm$0.063} & 0.716\spstd{$\pm$0.381} \\
ZipBench & -0.297\spstd{$\pm$0.480} & 0.624\spstd{$\pm$0.309} & \underline{0.818}\spstd{$\pm$0.063} & \underline{0.855}\spstd{$\pm$0.221} & 0.891\spstd{$\pm$0.096} & 0.578\spstd{$\pm$0.519} \\
Random & \textbf{0.598}\spstd{$\pm$0.304} & \underline{0.732}\spstd{$\pm$0.222} & 0.801\spstd{$\pm$0.177} & 0.846\spstd{$\pm$0.147} & \textbf{0.955}\spstd{$\pm$0.061} & \textbf{0.786}\spstd{$\pm$0.232} \\
\rowcolor{lightBlue}\textbf{BestRand (ours)} & \underline{0.543}\spstd{$\pm$0.328} & \textbf{0.735}\spstd{$\pm$0.219} & 0.783\spstd{$\pm$0.186} & 0.831\spstd{$\pm$0.159} & \underline{0.954}\spstd{$\pm$0.061} & \underline{0.769}\spstd{$\pm$0.249} \\
\midrule
\benchrow{OCRBench} \\
SparseEval (Linear) & 0.556\spstd{$\pm$0.179} & 0.778\spstd{$\pm$0.099} & 0.381\spstd{$\pm$0.304} & 0.357\spstd{$\pm$0.290} & 0.071\spstd{$\pm$0.234} & 0.429\spstd{$\pm$0.312} \\
SparseEval (MLP) & 0.929\spstd{$\pm$0.041} & 0.905\spstd{$\pm$0.024} & 0.921\spstd{$\pm$0.036} & 0.929\spstd{$\pm$0.024} & 0.952\spstd{$\pm$0.063} & 0.927\spstd{$\pm$0.038} \\
Anchor Points & 0.984\spstd{$\pm$0.014} & 0.976\spstd{$\pm$0.000} & 0.841\spstd{$\pm$0.172} & 0.841\spstd{$\pm$0.234} & 0.881\spstd{$\pm$0.126} & 0.905\spstd{$\pm$0.136} \\
TinyBenchmarks & \textbf{0.992}\spstd{$\pm$0.014} & \textbf{1.000}\spstd{$\pm$0.000} & 0.992\spstd{$\pm$0.014} & \textbf{1.000}\spstd{$\pm$0.000} & \textbf{1.000}\spstd{$\pm$0.000} & \textbf{0.997}\spstd{$\pm$0.008} \\
ZipBench & \textbf{0.992}\spstd{$\pm$0.014} & 0.992\spstd{$\pm$0.014} & \textbf{1.000}\spstd{$\pm$0.000} & 0.992\spstd{$\pm$0.014} & \textbf{1.000}\spstd{$\pm$0.000} & 0.995\spstd{$\pm$0.010} \\
Random & 0.989\spstd{$\pm$0.016} & 0.995\spstd{$\pm$0.010} & 0.997\spstd{$\pm$0.008} & 0.998\spstd{$\pm$0.006} & 1.000\spstd{$\pm$0.001} & 0.996\spstd{$\pm$0.010} \\
\rowcolor{lightBlue}\textbf{BestRand (ours)} & \underline{0.990}\spstd{$\pm$0.015} & \underline{0.995}\spstd{$\pm$0.010} & \underline{0.997}\spstd{$\pm$0.008} & \underline{0.998}\spstd{$\pm$0.006} & \underline{1.000}\spstd{$\pm$0.001} & \underline{0.996}\spstd{$\pm$0.010} \\
\bottomrule
\end{tabular}
\end{table*}

\clearpage

\begin{table*}[t]
\centering
\caption{Agent benchmark Spearman correlations at 5--50\% retention. Values are mean $\pm$ sample standard deviation; higher is better. Best means are \textbf{bold} and second-best means are \underline{underlined}.}
\vspace{4pt}
\label{tab:representative-sp-agent}
\footnotesize
\setlength{\tabcolsep}{2.8pt}
\renewcommand{\arraystretch}{1.00}
\newcommand{\spstd}[1]{\,{\scriptsize\color{black!70}#1}}
\newcommand{\benchrow}[1]{\multicolumn{7}{l}{\textcolor{blue!45!black}{\textbf{#1}}}}
\begin{tabular}{lrrrrrr}
\toprule
& \multicolumn{6}{c}{\textbf{Average Spearman correlation $\uparrow$}\enspace {\scriptsize mean $\pm$ std.}} \\
\cmidrule(lr){2-7}
\rowcolor{lightBlue}\textbf{Method} & \textbf{5\%} & \textbf{10\%} & \textbf{15\%} & \textbf{20\%} & \textbf{50\%} & \textbf{Mean} \\
\midrule
\benchrow{Drafter-Bench} \\
SparseEval (Linear) & 0.287\spstd{$\pm$0.353} & -0.065\spstd{$\pm$0.383} & -0.055\spstd{$\pm$0.471} & 0.011\spstd{$\pm$0.322} & 0.240\spstd{$\pm$0.153} & 0.084\spstd{$\pm$0.336} \\
SparseEval (MLP) & 0.667\spstd{$\pm$0.170} & 0.685\spstd{$\pm$0.101} & 0.227\spstd{$\pm$0.627} & 0.556\spstd{$\pm$0.080} & 0.683\spstd{$\pm$0.097} & 0.564\spstd{$\pm$0.311} \\
Anchor Points & 0.797\spstd{$\pm$0.067} & 0.868\spstd{$\pm$0.056} & 0.853\spstd{$\pm$0.111} & 0.951\spstd{$\pm$0.038} & 0.964\spstd{$\pm$0.024} & 0.887\spstd{$\pm$0.086} \\
TinyBenchmarks & 0.842\spstd{$\pm$0.029} & 0.897\spstd{$\pm$0.009} & 0.919\spstd{$\pm$0.013} & 0.927\spstd{$\pm$0.029} & 0.938\spstd{$\pm$0.006} & 0.905\spstd{$\pm$0.039} \\
ZipBench & 0.896\spstd{$\pm$0.053} & 0.946\spstd{$\pm$0.035} & 0.959\spstd{$\pm$0.016} & \textbf{0.980}\spstd{$\pm$0.013} & 0.984\spstd{$\pm$0.009} & 0.953\spstd{$\pm$0.042} \\
Random & \underline{0.916}\spstd{$\pm$0.041} & \underline{0.955}\spstd{$\pm$0.021} & \underline{0.968}\spstd{$\pm$0.014} & 0.976\spstd{$\pm$0.011} & \underline{0.991}\spstd{$\pm$0.005} & \underline{0.961}\spstd{$\pm$0.034} \\
\rowcolor{lightBlue}\textbf{BestRand (ours)} & \textbf{0.920}\spstd{$\pm$0.036} & \textbf{0.957}\spstd{$\pm$0.019} & \textbf{0.969}\spstd{$\pm$0.014} & \underline{0.977}\spstd{$\pm$0.011} & \textbf{0.991}\spstd{$\pm$0.005} & \textbf{0.963}\spstd{$\pm$0.031} \\
\midrule
\benchrow{OSWorld} \\
SparseEval (Linear) & 0.556\spstd{$\pm$0.241} & 0.422\spstd{$\pm$0.197} & 0.675\spstd{$\pm$0.101} & 0.533\spstd{$\pm$0.024} & 0.435\spstd{$\pm$0.148} & 0.524\spstd{$\pm$0.166} \\
SparseEval (MLP) & 0.253\spstd{$\pm$0.235} & -0.223\spstd{$\pm$0.394} & 0.063\spstd{$\pm$0.376} & 0.409\spstd{$\pm$0.186} & 0.743\spstd{$\pm$0.168} & 0.249\spstd{$\pm$0.415} \\
Anchor Points & 0.811\spstd{$\pm$0.090} & 0.847\spstd{$\pm$0.041} & 0.861\spstd{$\pm$0.028} & 0.909\spstd{$\pm$0.017} & 0.953\spstd{$\pm$0.011} & 0.876\spstd{$\pm$0.065} \\
TinyBenchmarks & 0.779\spstd{$\pm$0.082} & 0.904\spstd{$\pm$0.031} & 0.925\spstd{$\pm$0.030} & 0.947\spstd{$\pm$0.014} & 0.985\spstd{$\pm$0.005} & 0.908\spstd{$\pm$0.080} \\
ZipBench & \textbf{0.919}\spstd{$\pm$0.043} & \textbf{0.949}\spstd{$\pm$0.012} & \textbf{0.962}\spstd{$\pm$0.012} & 0.952\spstd{$\pm$0.025} & \textbf{0.988}\spstd{$\pm$0.005} & \textbf{0.954}\spstd{$\pm$0.030} \\
Random & 0.896\spstd{$\pm$0.051} & 0.939\spstd{$\pm$0.030} & 0.955\spstd{$\pm$0.023} & \underline{0.965}\spstd{$\pm$0.018} & 0.987\spstd{$\pm$0.008} & 0.948\spstd{$\pm$0.043} \\
\rowcolor{lightBlue}\textbf{BestRand (ours)} & \underline{0.902}\spstd{$\pm$0.045} & \underline{0.943}\spstd{$\pm$0.027} & \underline{0.958}\spstd{$\pm$0.020} & \textbf{0.967}\spstd{$\pm$0.017} & \underline{0.987}\spstd{$\pm$0.008} & \underline{0.951}\spstd{$\pm$0.039} \\
\midrule
\benchrow{BFCL} \\
SparseEval (Linear) & 0.229\spstd{$\pm$0.369} & 0.288\spstd{$\pm$0.431} & 0.261\spstd{$\pm$0.313} & 0.340\spstd{$\pm$0.320} & 0.220\spstd{$\pm$0.528} & 0.268\spstd{$\pm$0.341} \\
SparseEval (MLP) & 0.450\spstd{$\pm$0.371} & 0.790\spstd{$\pm$0.144} & 0.804\spstd{$\pm$0.058} & 0.791\spstd{$\pm$0.041} & 0.695\spstd{$\pm$0.332} & 0.706\spstd{$\pm$0.241} \\
Anchor Points & 0.825\spstd{$\pm$0.131} & 0.706\spstd{$\pm$0.202} & 0.786\spstd{$\pm$0.041} & 0.483\spstd{$\pm$0.391} & 0.865\spstd{$\pm$0.111} & 0.733\spstd{$\pm$0.227} \\
TinyBenchmarks & \textbf{0.973}\spstd{$\pm$0.008} & \textbf{0.989}\spstd{$\pm$0.007} & 0.982\spstd{$\pm$0.006} & 0.988\spstd{$\pm$0.000} & 0.994\spstd{$\pm$0.003} & 0.985\spstd{$\pm$0.009} \\
ZipBench & 0.964\spstd{$\pm$0.004} & \textbf{0.989}\spstd{$\pm$0.004} & \textbf{0.991}\spstd{$\pm$0.003} & 0.990\spstd{$\pm$0.009} & \textbf{1.000}\spstd{$\pm$0.000} & \textbf{0.987}\spstd{$\pm$0.013} \\
Random & 0.969\spstd{$\pm$0.017} & 0.983\spstd{$\pm$0.009} & 0.988\spstd{$\pm$0.007} & \underline{0.990}\spstd{$\pm$0.006} & 0.996\spstd{$\pm$0.004} & 0.985\spstd{$\pm$0.013} \\
\rowcolor{lightBlue}\textbf{BestRand (ours)} & \underline{0.970}\spstd{$\pm$0.016} & \underline{0.984}\spstd{$\pm$0.009} & \underline{0.988}\spstd{$\pm$0.007} & \textbf{0.991}\spstd{$\pm$0.006} & \underline{0.996}\spstd{$\pm$0.004} & \underline{0.986}\spstd{$\pm$0.013} \\
\midrule
\benchrow{TauBench} \\
SparseEval (Linear) & 0.276\spstd{$\pm$0.468} & 0.046\spstd{$\pm$0.404} & 0.441\spstd{$\pm$0.269} & 0.479\spstd{$\pm$0.198} & 0.354\spstd{$\pm$0.232} & 0.319\spstd{$\pm$0.322} \\
SparseEval (MLP) & 0.419\spstd{$\pm$0.151} & 0.345\spstd{$\pm$0.164} & 0.589\spstd{$\pm$0.046} & 0.474\spstd{$\pm$0.206} & 0.739\spstd{$\pm$0.013} & 0.513\spstd{$\pm$0.184} \\
Anchor Points & 0.354\spstd{$\pm$0.317} & 0.658\spstd{$\pm$0.210} & 0.634\spstd{$\pm$0.167} & 0.668\spstd{$\pm$0.147} & 0.723\spstd{$\pm$0.111} & 0.607\spstd{$\pm$0.218} \\
TinyBenchmarks & \underline{0.633}\spstd{$\pm$0.057} & 0.632\spstd{$\pm$0.070} & 0.739\spstd{$\pm$0.092} & 0.723\spstd{$\pm$0.114} & 0.921\spstd{$\pm$0.023} & 0.730\spstd{$\pm$0.127} \\
ZipBench & 0.552\spstd{$\pm$0.238} & 0.686\spstd{$\pm$0.125} & 0.783\spstd{$\pm$0.105} & 0.730\spstd{$\pm$0.147} & 0.853\spstd{$\pm$0.055} & 0.721\spstd{$\pm$0.162} \\
Random & \textbf{0.638}\spstd{$\pm$0.154} & \textbf{0.747}\spstd{$\pm$0.111} & \textbf{0.800}\spstd{$\pm$0.090} & \textbf{0.834}\spstd{$\pm$0.076} & \textbf{0.939}\spstd{$\pm$0.029} & \textbf{0.792}\spstd{$\pm$0.141} \\
\rowcolor{lightBlue}\textbf{BestRand (ours)} & 0.629\spstd{$\pm$0.158} & \underline{0.744}\spstd{$\pm$0.113} & \underline{0.798}\spstd{$\pm$0.092} & \underline{0.832}\spstd{$\pm$0.077} & \underline{0.938}\spstd{$\pm$0.029} & \underline{0.788}\spstd{$\pm$0.145} \\
\midrule
\benchrow{NexusBench} \\
SparseEval (Linear) & 0.453\spstd{$\pm$0.225} & 0.140\spstd{$\pm$0.390} & 0.225\spstd{$\pm$0.523} & 0.242\spstd{$\pm$0.454} & 0.549\spstd{$\pm$0.306} & 0.322\spstd{$\pm$0.369} \\
SparseEval (MLP) & -0.061\spstd{$\pm$0.058} & -0.015\spstd{$\pm$0.153} & 0.027\spstd{$\pm$0.214} & 0.201\spstd{$\pm$0.104} & 0.209\spstd{$\pm$0.164} & 0.072\spstd{$\pm$0.171} \\
Anchor Points & 0.760\spstd{$\pm$0.036} & 0.671\spstd{$\pm$0.062} & 0.698\spstd{$\pm$0.035} & 0.690\spstd{$\pm$0.122} & 0.861\spstd{$\pm$0.053} & 0.736\spstd{$\pm$0.093} \\
TinyBenchmarks & 0.811\spstd{$\pm$0.031} & \textbf{0.916}\spstd{$\pm$0.065} & \textbf{0.953}\spstd{$\pm$0.036} & 0.936\spstd{$\pm$0.030} & \textbf{0.977}\spstd{$\pm$0.007} & 0.919\spstd{$\pm$0.068} \\
ZipBench & \underline{0.853}\spstd{$\pm$0.071} & 0.897\spstd{$\pm$0.044} & 0.933\spstd{$\pm$0.040} & \textbf{0.949}\spstd{$\pm$0.017} & 0.936\spstd{$\pm$0.044} & 0.914\spstd{$\pm$0.053} \\
Random & 0.851\spstd{$\pm$0.080} & \underline{0.909}\spstd{$\pm$0.047} & 0.930\spstd{$\pm$0.037} & 0.942\spstd{$\pm$0.031} & 0.971\spstd{$\pm$0.017} & \underline{0.921}\spstd{$\pm$0.062} \\
\rowcolor{lightBlue}\textbf{BestRand (ours)} & \textbf{0.859}\spstd{$\pm$0.076} & 0.908\spstd{$\pm$0.047} & \underline{0.933}\spstd{$\pm$0.035} & \underline{0.945}\spstd{$\pm$0.030} & \underline{0.972}\spstd{$\pm$0.017} & \textbf{0.923}\spstd{$\pm$0.059} \\
\midrule
\benchrow{ComplexFuncBench} \\
SparseEval (Linear) & -0.012\spstd{$\pm$0.255} & 0.640\spstd{$\pm$0.095} & 0.483\spstd{$\pm$0.254} & 0.580\spstd{$\pm$0.190} & 0.269\spstd{$\pm$0.354} & 0.392\spstd{$\pm$0.322} \\
SparseEval (MLP) & 0.274\spstd{$\pm$0.424} & 0.444\spstd{$\pm$0.472} & 0.466\spstd{$\pm$0.489} & 0.794\spstd{$\pm$0.107} & 0.585\spstd{$\pm$0.378} & 0.513\spstd{$\pm$0.381} \\
Anchor Points & 0.801\spstd{$\pm$0.087} & 0.785\spstd{$\pm$0.027} & 0.869\spstd{$\pm$0.054} & 0.828\spstd{$\pm$0.052} & 0.876\spstd{$\pm$0.039} & 0.832\spstd{$\pm$0.060} \\
TinyBenchmarks & \textbf{0.958}\spstd{$\pm$0.005} & 0.948\spstd{$\pm$0.007} & \textbf{0.972}\spstd{$\pm$0.015} & \textbf{0.975}\spstd{$\pm$0.004} & 0.983\spstd{$\pm$0.003} & \textbf{0.967}\spstd{$\pm$0.015} \\
ZipBench & 0.834\spstd{$\pm$0.084} & 0.903\spstd{$\pm$0.051} & 0.945\spstd{$\pm$0.026} & 0.949\spstd{$\pm$0.011} & 0.977\spstd{$\pm$0.003} & 0.922\spstd{$\pm$0.064} \\
Random & 0.922\spstd{$\pm$0.034} & \textbf{0.955}\spstd{$\pm$0.019} & 0.968\spstd{$\pm$0.014} & \underline{0.974}\spstd{$\pm$0.011} & \underline{0.989}\spstd{$\pm$0.005} & 0.962\spstd{$\pm$0.030} \\
\rowcolor{lightBlue}\textbf{BestRand (ours)} & \underline{0.922}\spstd{$\pm$0.034} & \underline{0.955}\spstd{$\pm$0.019} & \underline{0.968}\spstd{$\pm$0.014} & 0.974\spstd{$\pm$0.012} & \textbf{0.989}\spstd{$\pm$0.005} & \underline{0.962}\spstd{$\pm$0.030} \\
\bottomrule
\end{tabular}
\end{table*}

\clearpage

\begin{table*}[t]
\centering
\caption{Text benchmark MAE at 2--50\% retention. Values are mean $\pm$ sample standard deviation in percent; lower is better. Best means are \textbf{bold} and second-best means are \underline{underlined}.}
\vspace{4pt}
\label{tab:representative-mae-text}
\footnotesize
\setlength{\tabcolsep}{2.8pt}
\renewcommand{\arraystretch}{1.00}
\newcommand{\maestd}[1]{\,{\scriptsize\color{black!70}#1}}
\newcommand{\benchrow}[1]{\multicolumn{8}{l}{\textcolor{blue!45!black}{\textbf{#1}}}}
\resizebox{\textwidth}{!}{%
\begin{tabular}{lrrrrrrr}
\toprule
& \multicolumn{7}{c}{\textbf{Average MAE (\%) $\downarrow$}\enspace {\scriptsize mean $\pm$ std.}} \\
\cmidrule(lr){2-8}
\rowcolor{lightBlue}\textbf{Method} & \textbf{2\%} & \textbf{5\%} & \textbf{10\%} & \textbf{15\%} & \textbf{20\%} & \textbf{50\%} & \textbf{Mean} \\
\midrule
\benchrow{TruthfulQA} \\
SparseEval (Linear) & 15.04\maestd{$\pm$5.08} & 28.42\maestd{$\pm$3.90} & 18.03\maestd{$\pm$5.94} & 16.60\maestd{$\pm$3.22} & 15.08\maestd{$\pm$5.39} & 19.15\maestd{$\pm$5.62} & 18.72\maestd{$\pm$6.29} \\
SparseEval (MLP) & 9.21\maestd{$\pm$3.15} & 10.38\maestd{$\pm$1.97} & 9.66\maestd{$\pm$3.54} & 9.83\maestd{$\pm$3.86} & 9.72\maestd{$\pm$3.59} & 9.52\maestd{$\pm$2.94} & 9.72\maestd{$\pm$2.74} \\
Anchor Points & 7.15\maestd{$\pm$1.00} & 7.81\maestd{$\pm$0.87} & 4.81\maestd{$\pm$0.64} & 5.24\maestd{$\pm$0.08} & 2.70\maestd{$\pm$0.33} & 2.35\maestd{$\pm$0.17} & 5.01\maestd{$\pm$2.16} \\
TinyBenchmarks & \underline{6.21}\maestd{$\pm$1.15} & \underline{4.50}\maestd{$\pm$1.04} & 3.03\maestd{$\pm$0.24} & \underline{2.25}\maestd{$\pm$0.28} & \underline{2.27}\maestd{$\pm$0.37} & 1.26\maestd{$\pm$0.26} & \underline{3.25}\maestd{$\pm$1.79} \\
ZipBench & \textbf{5.36}\maestd{$\pm$1.05} & 4.53\maestd{$\pm$0.85} & \underline{2.62}\maestd{$\pm$0.75} & 3.65\maestd{$\pm$1.65} & 2.73\maestd{$\pm$0.87} & \underline{1.13}\maestd{$\pm$0.36} & 3.34\maestd{$\pm$1.65} \\
Random & 8.33\maestd{$\pm$3.52} & 5.12\maestd{$\pm$2.20} & 3.52\maestd{$\pm$1.52} & 2.79\maestd{$\pm$1.20} & 2.35\maestd{$\pm$1.01} & 1.17\maestd{$\pm$0.51} & 3.88\maestd{$\pm$3.02} \\
\rowcolor{lightBlue}\textbf{BestRand (ours)} & 6.34\maestd{$\pm$1.96} & \textbf{3.85}\maestd{$\pm$1.13} & \textbf{2.60}\maestd{$\pm$0.74} & \textbf{2.00}\maestd{$\pm$0.53} & \textbf{1.64}\maestd{$\pm$0.42} & \textbf{0.79}\maestd{$\pm$0.18} & \textbf{2.87}\maestd{$\pm$2.08} \\
\midrule
\benchrow{BoolQ} \\
SparseEval (Linear) & 15.53\maestd{$\pm$4.48} & 21.38\maestd{$\pm$4.02} & 15.34\maestd{$\pm$2.54} & 16.23\maestd{$\pm$3.09} & 16.56\maestd{$\pm$3.33} & 29.22\maestd{$\pm$19.94} & 19.04\maestd{$\pm$8.97} \\
SparseEval (MLP) & 20.51\maestd{$\pm$14.05} & 24.82\maestd{$\pm$8.90} & 26.04\maestd{$\pm$3.12} & 26.04\maestd{$\pm$2.83} & 22.15\maestd{$\pm$13.50} & 23.02\maestd{$\pm$8.99} & 23.76\maestd{$\pm$8.37} \\
Anchor Points & 8.86\maestd{$\pm$1.22} & 10.64\maestd{$\pm$1.24} & 15.29\maestd{$\pm$6.83} & 5.98\maestd{$\pm$0.57} & 8.65\maestd{$\pm$4.85} & 5.80\maestd{$\pm$2.80} & 9.20\maestd{$\pm$4.52} \\
TinyBenchmarks & 4.44\maestd{$\pm$0.55} & 2.63\maestd{$\pm$0.57} & \textbf{1.29}\maestd{$\pm$0.23} & \textbf{1.00}\maestd{$\pm$0.45} & \textbf{0.78}\maestd{$\pm$0.26} & \underline{0.47}\maestd{$\pm$0.02} & 1.77\maestd{$\pm$1.46} \\
ZipBench & 3.66\maestd{$\pm$0.42} & 2.61\maestd{$\pm$0.05} & 1.39\maestd{$\pm$0.14} & 1.56\maestd{$\pm$0.19} & 0.97\maestd{$\pm$0.22} & 0.60\maestd{$\pm$0.07} & 1.80\maestd{$\pm$1.09} \\
Random & \underline{3.51}\maestd{$\pm$1.11} & \underline{2.17}\maestd{$\pm$0.69} & 1.50\maestd{$\pm$0.48} & 1.19\maestd{$\pm$0.38} & 1.00\maestd{$\pm$0.31} & 0.50\maestd{$\pm$0.16} & \underline{1.64}\maestd{$\pm$1.15} \\
\rowcolor{lightBlue}\textbf{BestRand (ours)} & \textbf{3.21}\maestd{$\pm$0.86} & \textbf{1.99}\maestd{$\pm$0.53} & \underline{1.37}\maestd{$\pm$0.36} & \underline{1.07}\maestd{$\pm$0.28} & \underline{0.90}\maestd{$\pm$0.23} & \textbf{0.45}\maestd{$\pm$0.11} & \textbf{1.50}\maestd{$\pm$1.01} \\
\midrule
\benchrow{IFEval} \\
SparseEval (Linear) & 32.75\maestd{$\pm$32.08} & 21.84\maestd{$\pm$6.06} & 21.50\maestd{$\pm$2.41} & 21.18\maestd{$\pm$4.79} & 17.15\maestd{$\pm$3.31} & 28.37\maestd{$\pm$15.80} & 23.80\maestd{$\pm$13.71} \\
SparseEval (MLP) & 14.45\maestd{$\pm$2.31} & 18.21\maestd{$\pm$5.69} & 11.18\maestd{$\pm$4.12} & 12.93\maestd{$\pm$2.27} & 11.23\maestd{$\pm$5.81} & 11.82\maestd{$\pm$6.12} & 13.30\maestd{$\pm$4.67} \\
Anchor Points & \textbf{8.86}\maestd{$\pm$3.45} & 8.36\maestd{$\pm$1.57} & 7.27\maestd{$\pm$1.82} & 7.57\maestd{$\pm$1.37} & 7.99\maestd{$\pm$1.27} & 4.60\maestd{$\pm$0.61} & 7.44\maestd{$\pm$2.13} \\
TinyBenchmarks & 10.56\maestd{$\pm$0.92} & \underline{6.08}\maestd{$\pm$0.33} & 4.69\maestd{$\pm$1.54} & \underline{3.31}\maestd{$\pm$0.22} & \textbf{2.49}\maestd{$\pm$0.18} & 1.63\maestd{$\pm$0.45} & \underline{4.79}\maestd{$\pm$3.11} \\
ZipBench & 10.38\maestd{$\pm$1.44} & 8.12\maestd{$\pm$1.74} & 5.12\maestd{$\pm$0.20} & 3.61\maestd{$\pm$0.20} & 3.60\maestd{$\pm$0.39} & 1.85\maestd{$\pm$0.14} & 5.45\maestd{$\pm$3.11} \\
Random & 10.41\maestd{$\pm$2.53} & 6.49\maestd{$\pm$1.62} & \underline{4.47}\maestd{$\pm$1.11} & 3.54\maestd{$\pm$0.89} & 2.97\maestd{$\pm$0.75} & \underline{1.48}\maestd{$\pm$0.37} & 4.89\maestd{$\pm$3.22} \\
\rowcolor{lightBlue}\textbf{BestRand (ours)} & \underline{9.55}\maestd{$\pm$1.96} & \textbf{5.92}\maestd{$\pm$1.21} & \textbf{4.06}\maestd{$\pm$0.82} & \textbf{3.20}\maestd{$\pm$0.64} & \underline{2.67}\maestd{$\pm$0.53} & \textbf{1.32}\maestd{$\pm$0.26} & \textbf{4.45}\maestd{$\pm$2.88} \\
\midrule
\benchrow{Finance-Fundamentals} \\
SparseEval (Linear) & 25.66\maestd{$\pm$6.07} & 19.15\maestd{$\pm$5.22} & 22.12\maestd{$\pm$1.99} & 16.94\maestd{$\pm$2.13} & 19.21\maestd{$\pm$1.26} & 36.02\maestd{$\pm$26.96} & 23.18\maestd{$\pm$11.71} \\
SparseEval (MLP) & \textbf{8.90}\maestd{$\pm$1.29} & 8.72\maestd{$\pm$0.82} & 8.37\maestd{$\pm$1.09} & 7.09\maestd{$\pm$1.14} & 6.34\maestd{$\pm$0.35} & 8.74\maestd{$\pm$1.79} & 8.03\maestd{$\pm$1.40} \\
Anchor Points & 9.90\maestd{$\pm$5.05} & 9.57\maestd{$\pm$3.61} & 8.09\maestd{$\pm$2.43} & 8.49\maestd{$\pm$3.44} & 8.16\maestd{$\pm$4.27} & 6.25\maestd{$\pm$2.63} & 8.41\maestd{$\pm$3.33} \\
TinyBenchmarks & 11.84\maestd{$\pm$1.85} & \underline{6.58}\maestd{$\pm$0.65} & 5.53\maestd{$\pm$1.71} & 4.45\maestd{$\pm$0.68} & \underline{2.81}\maestd{$\pm$0.63} & 1.71\maestd{$\pm$0.33} & 5.49\maestd{$\pm$3.49} \\
ZipBench & \underline{9.52}\maestd{$\pm$0.39} & 7.11\maestd{$\pm$1.43} & 4.90\maestd{$\pm$0.67} & 4.13\maestd{$\pm$0.19} & 3.36\maestd{$\pm$0.50} & 1.93\maestd{$\pm$0.27} & \underline{5.16}\maestd{$\pm$2.65} \\
Random & 11.25\maestd{$\pm$3.35} & 7.00\maestd{$\pm$2.11} & \underline{4.81}\maestd{$\pm$1.45} & \underline{3.81}\maestd{$\pm$1.15} & 3.20\maestd{$\pm$0.97} & \underline{1.60}\maestd{$\pm$0.48} & 5.28\maestd{$\pm$3.63} \\
\rowcolor{lightBlue}\textbf{BestRand (ours)} & 9.99\maestd{$\pm$2.31} & \textbf{6.17}\maestd{$\pm$1.39} & \textbf{4.23}\maestd{$\pm$0.93} & \textbf{3.30}\maestd{$\pm$0.69} & \textbf{2.75}\maestd{$\pm$0.57} & \textbf{1.37}\maestd{$\pm$0.28} & \textbf{4.64}\maestd{$\pm$3.06} \\
\midrule
\benchrow{LIFBench} \\
SparseEval (Linear) & 22.59\maestd{$\pm$8.91} & 17.68\maestd{$\pm$6.25} & 39.51\maestd{$\pm$36.09} & 32.91\maestd{$\pm$23.26} & 27.01\maestd{$\pm$26.44} & 21.75\maestd{$\pm$17.14} & 26.91\maestd{$\pm$20.12} \\
SparseEval (MLP) & 13.20\maestd{$\pm$7.24} & 8.33\maestd{$\pm$1.14} & 8.89\maestd{$\pm$1.49} & 10.34\maestd{$\pm$1.55} & 11.39\maestd{$\pm$1.15} & 12.29\maestd{$\pm$1.14} & 10.74\maestd{$\pm$3.22} \\
Anchor Points & 2.84\maestd{$\pm$0.26} & 1.94\maestd{$\pm$0.56} & 1.65\maestd{$\pm$0.76} & 1.34\maestd{$\pm$0.16} & 1.70\maestd{$\pm$0.62} & 1.04\maestd{$\pm$0.35} & 1.75\maestd{$\pm$0.72} \\
TinyBenchmarks & 7.22\maestd{$\pm$1.70} & 6.24\maestd{$\pm$0.24} & 5.67\maestd{$\pm$0.43} & 6.46\maestd{$\pm$0.47} & 6.67\maestd{$\pm$0.17} & 6.55\maestd{$\pm$0.39} & 6.47\maestd{$\pm$0.81} \\
ZipBench & 3.06\maestd{$\pm$1.44} & \underline{1.58}\maestd{$\pm$0.46} & \underline{1.11}\maestd{$\pm$0.06} & \underline{0.88}\maestd{$\pm$0.19} & 0.94\maestd{$\pm$0.07} & 0.48\maestd{$\pm$0.21} & 1.34\maestd{$\pm$1.01} \\
Random & \underline{2.80}\maestd{$\pm$0.91} & 1.74\maestd{$\pm$0.56} & 1.20\maestd{$\pm$0.39} & 0.95\maestd{$\pm$0.31} & \underline{0.80}\maestd{$\pm$0.26} & \underline{0.40}\maestd{$\pm$0.13} & \underline{1.32}\maestd{$\pm$0.92} \\
\rowcolor{lightBlue}\textbf{BestRand (ours)} & \textbf{2.36}\maestd{$\pm$0.58} & \textbf{1.45}\maestd{$\pm$0.36} & \textbf{1.00}\maestd{$\pm$0.24} & \textbf{0.77}\maestd{$\pm$0.18} & \textbf{0.64}\maestd{$\pm$0.15} & \textbf{0.31}\maestd{$\pm$0.07} & \textbf{1.09}\maestd{$\pm$0.74} \\
\midrule
\benchrow{GPQA} \\
SparseEval (Linear) & 19.39\maestd{$\pm$2.81} & 23.64\maestd{$\pm$6.78} & 20.13\maestd{$\pm$1.11} & 20.74\maestd{$\pm$4.52} & 21.69\maestd{$\pm$1.96} & 34.14\maestd{$\pm$13.55} & 23.29\maestd{$\pm$7.60} \\
SparseEval (MLP) & \textbf{6.60}\maestd{$\pm$1.15} & \underline{6.85}\maestd{$\pm$0.66} & 7.57\maestd{$\pm$1.40} & 6.67\maestd{$\pm$0.61} & 5.91\maestd{$\pm$0.82} & 5.42\maestd{$\pm$0.52} & 6.50\maestd{$\pm$1.04} \\
Anchor Points & 14.17\maestd{$\pm$2.34} & 8.78\maestd{$\pm$1.89} & 8.47\maestd{$\pm$2.81} & 8.87\maestd{$\pm$1.13} & 10.12\maestd{$\pm$1.92} & 7.41\maestd{$\pm$1.31} & 9.64\maestd{$\pm$2.79} \\
TinyBenchmarks & 13.69\maestd{$\pm$3.37} & 9.34\maestd{$\pm$2.44} & \textbf{3.71}\maestd{$\pm$0.43} & \textbf{3.40}\maestd{$\pm$1.10} & 3.25\maestd{$\pm$1.33} & 2.32\maestd{$\pm$0.47} & 5.95\maestd{$\pm$4.55} \\
ZipBench & 12.59\maestd{$\pm$4.88} & 8.60\maestd{$\pm$0.96} & 5.41\maestd{$\pm$0.65} & 4.88\maestd{$\pm$1.12} & 3.80\maestd{$\pm$0.49} & 2.32\maestd{$\pm$0.55} & 6.27\maestd{$\pm$3.94} \\
Random & 10.85\maestd{$\pm$2.39} & 6.87\maestd{$\pm$1.49} & 4.68\maestd{$\pm$1.01} & 3.71\maestd{$\pm$0.81} & \underline{3.12}\maestd{$\pm$0.68} & \underline{1.56}\maestd{$\pm$0.34} & \underline{5.13}\maestd{$\pm$3.29} \\
\rowcolor{lightBlue}\textbf{BestRand (ours)} & \underline{10.56}\maestd{$\pm$2.25} & \textbf{6.64}\maestd{$\pm$1.36} & \underline{4.53}\maestd{$\pm$0.93} & \underline{3.58}\maestd{$\pm$0.74} & \textbf{3.00}\maestd{$\pm$0.60} & \textbf{1.49}\maestd{$\pm$0.30} & \textbf{4.97}\maestd{$\pm$3.19} \\
\bottomrule
\end{tabular}
}%
\end{table*}

\clearpage

\begin{table*}[t]
\centering
\caption{Multimodal benchmark MAE at 2--50\% retention. Values are mean $\pm$ sample standard deviation in percent; lower is better. Best means are \textbf{bold} and second-best means are \underline{underlined}.}
\vspace{4pt}
\label{tab:representative-mae-multimodal}
\footnotesize
\setlength{\tabcolsep}{2.8pt}
\renewcommand{\arraystretch}{1.00}
\newcommand{\maestd}[1]{\,{\scriptsize\color{black!70}#1}}
\newcommand{\benchrow}[1]{\multicolumn{8}{l}{\textcolor{blue!45!black}{\textbf{#1}}}}
\resizebox{\textwidth}{!}{%
\begin{tabular}{lrrrrrrr}
\toprule
& \multicolumn{7}{c}{\textbf{Average MAE (\%) $\downarrow$}\enspace {\scriptsize mean $\pm$ std.}} \\
\cmidrule(lr){2-8}
\rowcolor{lightBlue}\textbf{Method} & \textbf{2\%} & \textbf{5\%} & \textbf{10\%} & \textbf{15\%} & \textbf{20\%} & \textbf{50\%} & \textbf{Mean} \\
\midrule
\benchrow{MathVerse} \\
SparseEval (Linear) & 42.87\maestd{$\pm$30.75} & 33.96\maestd{$\pm$26.14} & 42.56\maestd{$\pm$27.64} & 44.15\maestd{$\pm$17.58} & 30.57\maestd{$\pm$23.97} & 16.27\maestd{$\pm$4.08} & 35.06\maestd{$\pm$22.11} \\
SparseEval (MLP) & 18.20\maestd{$\pm$4.98} & 15.64\maestd{$\pm$3.33} & 17.82\maestd{$\pm$1.94} & 13.49\maestd{$\pm$2.57} & 10.64\maestd{$\pm$2.59} & 13.95\maestd{$\pm$1.79} & 14.96\maestd{$\pm$3.72} \\
Anchor Points & 9.14\maestd{$\pm$1.59} & 7.81\maestd{$\pm$1.58} & 7.46\maestd{$\pm$1.23} & 7.81\maestd{$\pm$1.84} & 7.63\maestd{$\pm$1.61} & 4.41\maestd{$\pm$0.87} & 7.38\maestd{$\pm$1.93} \\
TinyBenchmarks & 9.54\maestd{$\pm$0.11} & 5.57\maestd{$\pm$0.42} & \underline{3.79}\maestd{$\pm$0.98} & 3.40\maestd{$\pm$1.05} & \underline{2.54}\maestd{$\pm$0.16} & \underline{1.26}\maestd{$\pm$0.47} & 4.35\maestd{$\pm$2.79} \\
ZipBench & \textbf{7.13}\maestd{$\pm$1.16} & \underline{5.30}\maestd{$\pm$1.12} & 4.66\maestd{$\pm$0.80} & 3.45\maestd{$\pm$0.90} & 3.05\maestd{$\pm$0.28} & 1.41\maestd{$\pm$0.37} & \underline{4.17}\maestd{$\pm$1.99} \\
Random & 8.90\maestd{$\pm$2.19} & 5.57\maestd{$\pm$1.40} & 3.81\maestd{$\pm$0.96} & \underline{3.02}\maestd{$\pm$0.76} & 2.54\maestd{$\pm$0.64} & 1.27\maestd{$\pm$0.32} & 4.19\maestd{$\pm$2.76} \\
\rowcolor{lightBlue}\textbf{BestRand (ours)} & \underline{8.27}\maestd{$\pm$1.70} & \textbf{5.12}\maestd{$\pm$1.04} & \textbf{3.48}\maestd{$\pm$0.71} & \textbf{2.74}\maestd{$\pm$0.55} & \textbf{2.28}\maestd{$\pm$0.45} & \textbf{1.14}\maestd{$\pm$0.22} & \textbf{3.84}\maestd{$\pm$2.49} \\
\midrule
\benchrow{OmniDocBench} \\
SparseEval (Linear) & 20.11\maestd{$\pm$6.80} & 20.37\maestd{$\pm$12.29} & 13.22\maestd{$\pm$2.91} & 12.83\maestd{$\pm$8.10} & 9.04\maestd{$\pm$5.33} & 9.32\maestd{$\pm$3.95} & 14.15\maestd{$\pm$7.71} \\
SparseEval (MLP) & 2.59\maestd{$\pm$1.30} & 2.03\maestd{$\pm$0.91} & 2.75\maestd{$\pm$1.92} & 2.14\maestd{$\pm$0.80} & 2.53\maestd{$\pm$1.54} & 2.24\maestd{$\pm$1.50} & 2.38\maestd{$\pm$1.19} \\
Anchor Points & \underline{1.66}\maestd{$\pm$0.20} & \textbf{0.88}\maestd{$\pm$0.38} & 0.90\maestd{$\pm$0.25} & 1.37\maestd{$\pm$0.79} & 1.03\maestd{$\pm$0.20} & 0.60\maestd{$\pm$0.05} & 1.07\maestd{$\pm$0.48} \\
TinyBenchmarks & 2.85\maestd{$\pm$0.09} & 2.34\maestd{$\pm$1.27} & 2.74\maestd{$\pm$0.50} & 2.82\maestd{$\pm$0.44} & 2.73\maestd{$\pm$0.25} & 2.52\maestd{$\pm$0.11} & 2.67\maestd{$\pm$0.54} \\
ZipBench & \textbf{1.37}\maestd{$\pm$0.69} & 1.31\maestd{$\pm$0.46} & \textbf{0.63}\maestd{$\pm$0.10} & \underline{0.61}\maestd{$\pm$0.09} & \underline{0.57}\maestd{$\pm$0.24} & 0.37\maestd{$\pm$0.17} & \underline{0.81}\maestd{$\pm$0.50} \\
Random & 2.03\maestd{$\pm$0.92} & 1.27\maestd{$\pm$0.57} & 0.87\maestd{$\pm$0.40} & 0.69\maestd{$\pm$0.31} & 0.58\maestd{$\pm$0.26} & \underline{0.29}\maestd{$\pm$0.13} & 0.96\maestd{$\pm$0.76} \\
\rowcolor{lightBlue}\textbf{BestRand (ours)} & 1.70\maestd{$\pm$0.65} & \underline{1.06}\maestd{$\pm$0.40} & \underline{0.72}\maestd{$\pm$0.27} & \textbf{0.57}\maestd{$\pm$0.21} & \textbf{0.47}\maestd{$\pm$0.17} & \textbf{0.23}\maestd{$\pm$0.08} & \textbf{0.79}\maestd{$\pm$0.59} \\
\midrule
\benchrow{SpatialEval} \\
SparseEval (Linear) & 18.82\maestd{$\pm$5.34} & 12.37\maestd{$\pm$4.15} & 19.93\maestd{$\pm$3.49} & 17.77\maestd{$\pm$4.96} & 17.27\maestd{$\pm$6.06} & 15.49\maestd{$\pm$9.10} & 16.94\maestd{$\pm$5.49} \\
SparseEval (MLP) & 25.25\maestd{$\pm$10.16} & 33.02\maestd{$\pm$6.53} & 23.33\maestd{$\pm$12.23} & 24.39\maestd{$\pm$12.19} & 30.81\maestd{$\pm$4.04} & 21.17\maestd{$\pm$10.90} & 26.33\maestd{$\pm$9.31} \\
Anchor Points & 4.97\maestd{$\pm$0.33} & 5.93\maestd{$\pm$2.25} & 6.09\maestd{$\pm$2.49} & 4.92\maestd{$\pm$2.53} & 5.80\maestd{$\pm$0.64} & 3.60\maestd{$\pm$1.64} & 5.22\maestd{$\pm$1.80} \\
TinyBenchmarks & 3.99\maestd{$\pm$0.53} & \underline{1.54}\maestd{$\pm$0.50} & 1.11\maestd{$\pm$0.20} & \underline{0.84}\maestd{$\pm$0.13} & \textbf{0.61}\maestd{$\pm$0.28} & 0.57\maestd{$\pm$0.08} & 1.44\maestd{$\pm$1.25} \\
ZipBench & \underline{2.33}\maestd{$\pm$1.00} & 1.75\maestd{$\pm$0.50} & \textbf{0.96}\maestd{$\pm$0.65} & 0.85\maestd{$\pm$0.24} & 0.81\maestd{$\pm$0.26} & 0.64\maestd{$\pm$0.17} & \underline{1.22}\maestd{$\pm$0.78} \\
Random & 2.62\maestd{$\pm$1.06} & 1.63\maestd{$\pm$0.66} & 1.12\maestd{$\pm$0.45} & 0.89\maestd{$\pm$0.36} & 0.75\maestd{$\pm$0.30} & \underline{0.37}\maestd{$\pm$0.15} & 1.23\maestd{$\pm$0.93} \\
\rowcolor{lightBlue}\textbf{BestRand (ours)} & \textbf{2.30}\maestd{$\pm$0.83} & \textbf{1.42}\maestd{$\pm$0.52} & \underline{0.98}\maestd{$\pm$0.35} & \textbf{0.77}\maestd{$\pm$0.27} & \underline{0.64}\maestd{$\pm$0.23} & \textbf{0.32}\maestd{$\pm$0.11} & \textbf{1.07}\maestd{$\pm$0.79} \\
\midrule
\benchrow{LogicVista} \\
SparseEval (Linear) & 28.92\maestd{$\pm$3.49} & 21.78\maestd{$\pm$3.79} & 25.59\maestd{$\pm$5.46} & 35.28\maestd{$\pm$5.20} & 29.37\maestd{$\pm$7.28} & 26.44\maestd{$\pm$5.22} & 27.90\maestd{$\pm$6.11} \\
SparseEval (MLP) & \underline{11.20}\maestd{$\pm$1.89} & 10.27\maestd{$\pm$1.57} & 9.12\maestd{$\pm$0.96} & 9.24\maestd{$\pm$2.57} & 9.72\maestd{$\pm$1.42} & 8.34\maestd{$\pm$0.56} & 9.65\maestd{$\pm$1.66} \\
Anchor Points & 11.41\maestd{$\pm$1.66} & \underline{7.81}\maestd{$\pm$1.91} & 8.10\maestd{$\pm$0.64} & 6.20\maestd{$\pm$0.79} & 7.52\maestd{$\pm$2.76} & 4.49\maestd{$\pm$1.59} & 7.59\maestd{$\pm$2.60} \\
TinyBenchmarks & 11.62\maestd{$\pm$1.77} & 8.34\maestd{$\pm$0.79} & \textbf{5.00}\maestd{$\pm$0.66} & \textbf{3.82}\maestd{$\pm$0.32} & \underline{3.45}\maestd{$\pm$0.13} & 2.54\maestd{$\pm$0.07} & \underline{5.79}\maestd{$\pm$3.36} \\
ZipBench & \textbf{9.72}\maestd{$\pm$0.39} & 8.02\maestd{$\pm$0.68} & 6.37\maestd{$\pm$0.35} & 5.26\maestd{$\pm$0.46} & 4.08\maestd{$\pm$1.03} & \underline{1.68}\maestd{$\pm$0.61} & 5.85\maestd{$\pm$2.74} \\
Random & 12.81\maestd{$\pm$2.98} & 7.96\maestd{$\pm$1.90} & 5.44\maestd{$\pm$1.29} & 4.33\maestd{$\pm$1.02} & 3.65\maestd{$\pm$0.86} & 1.81\maestd{$\pm$0.43} & 6.00\maestd{$\pm$3.93} \\
\rowcolor{lightBlue}\textbf{BestRand (ours)} & 11.86\maestd{$\pm$2.34} & \textbf{7.40}\maestd{$\pm$1.47} & \underline{5.06}\maestd{$\pm$0.98} & \underline{4.01}\maestd{$\pm$0.77} & \textbf{3.34}\maestd{$\pm$0.63} & \textbf{1.64}\maestd{$\pm$0.30} & \textbf{5.55}\maestd{$\pm$3.55} \\
\midrule
\benchrow{RealWorldQA} \\
SparseEval (Linear) & 17.28\maestd{$\pm$2.79} & 24.08\maestd{$\pm$14.00} & 19.67\maestd{$\pm$4.21} & 17.69\maestd{$\pm$4.59} & 20.10\maestd{$\pm$7.48} & 14.89\maestd{$\pm$9.57} & 18.95\maestd{$\pm$7.39} \\
SparseEval (MLP) & 19.57\maestd{$\pm$14.34} & 26.67\maestd{$\pm$8.86} & 27.42\maestd{$\pm$11.33} & 26.42\maestd{$\pm$11.46} & 17.26\maestd{$\pm$9.83} & 21.77\maestd{$\pm$9.79} & 23.18\maestd{$\pm$10.13} \\
Anchor Points & 8.40\maestd{$\pm$2.87} & 10.46\maestd{$\pm$4.50} & 6.68\maestd{$\pm$2.63} & 5.14\maestd{$\pm$1.46} & 9.16\maestd{$\pm$3.79} & 8.60\maestd{$\pm$3.43} & 8.07\maestd{$\pm$3.26} \\
TinyBenchmarks & 9.62\maestd{$\pm$6.00} & 5.09\maestd{$\pm$1.68} & 3.37\maestd{$\pm$1.12} & \textbf{2.03}\maestd{$\pm$0.25} & 3.30\maestd{$\pm$2.57} & 1.91\maestd{$\pm$1.00} & 4.22\maestd{$\pm$3.60} \\
ZipBench & 8.82\maestd{$\pm$3.05} & 6.37\maestd{$\pm$0.83} & 3.66\maestd{$\pm$1.25} & 2.99\maestd{$\pm$1.79} & \textbf{1.78}\maestd{$\pm$0.91} & 2.14\maestd{$\pm$1.26} & 4.30\maestd{$\pm$2.95} \\
Random & \underline{7.90}\maestd{$\pm$3.31} & \underline{4.91}\maestd{$\pm$2.00} & \underline{3.35}\maestd{$\pm$1.38} & 2.65\maestd{$\pm$1.11} & 2.24\maestd{$\pm$0.92} & \underline{1.12}\maestd{$\pm$0.46} & \underline{3.70}\maestd{$\pm$2.83} \\
\rowcolor{lightBlue}\textbf{BestRand (ours)} & \textbf{7.24}\maestd{$\pm$2.79} & \textbf{4.54}\maestd{$\pm$1.66} & \textbf{3.06}\maestd{$\pm$1.12} & \underline{2.42}\maestd{$\pm$0.93} & \underline{2.02}\maestd{$\pm$0.75} & \textbf{1.00}\maestd{$\pm$0.37} & \textbf{3.38}\maestd{$\pm$2.52} \\
\midrule
\benchrow{OCRBench} \\
SparseEval (Linear) & 16.95\maestd{$\pm$5.37} & 15.14\maestd{$\pm$1.55} & 12.74\maestd{$\pm$3.40} & 33.01\maestd{$\pm$22.07} & 22.23\maestd{$\pm$10.23} & 20.70\maestd{$\pm$1.75} & 20.13\maestd{$\pm$11.00} \\
SparseEval (MLP) & 6.84\maestd{$\pm$5.75} & 9.07\maestd{$\pm$8.89} & 7.67\maestd{$\pm$8.39} & 6.47\maestd{$\pm$6.18} & 5.38\maestd{$\pm$5.46} & 3.88\maestd{$\pm$1.00} & 6.55\maestd{$\pm$5.69} \\
Anchor Points & 3.13\maestd{$\pm$1.35} & 2.07\maestd{$\pm$0.65} & 2.67\maestd{$\pm$0.59} & 2.55\maestd{$\pm$1.74} & 3.87\maestd{$\pm$2.26} & 2.82\maestd{$\pm$1.06} & 2.85\maestd{$\pm$1.31} \\
TinyBenchmarks & \underline{1.72}\maestd{$\pm$0.09} & 2.32\maestd{$\pm$0.30} & 1.81\maestd{$\pm$0.19} & 2.20\maestd{$\pm$0.12} & 2.21\maestd{$\pm$0.22} & 1.81\maestd{$\pm$0.13} & 2.01\maestd{$\pm$0.29} \\
ZipBench & 1.83\maestd{$\pm$0.44} & \underline{1.07}\maestd{$\pm$0.09} & \underline{0.59}\maestd{$\pm$0.16} & \textbf{0.39}\maestd{$\pm$0.14} & \underline{0.39}\maestd{$\pm$0.21} & 0.29\maestd{$\pm$0.09} & \underline{0.76}\maestd{$\pm$0.59} \\
Random & 1.86\maestd{$\pm$0.92} & 1.15\maestd{$\pm$0.57} & 0.80\maestd{$\pm$0.39} & 0.63\maestd{$\pm$0.31} & 0.53\maestd{$\pm$0.26} & \underline{0.27}\maestd{$\pm$0.13} & 0.87\maestd{$\pm$0.72} \\
\rowcolor{lightBlue}\textbf{BestRand (ours)} & \textbf{1.38}\maestd{$\pm$0.50} & \textbf{0.85}\maestd{$\pm$0.29} & \textbf{0.57}\maestd{$\pm$0.19} & \underline{0.44}\maestd{$\pm$0.14} & \textbf{0.36}\maestd{$\pm$0.11} & \textbf{0.17}\maestd{$\pm$0.05} & \textbf{0.63}\maestd{$\pm$0.47} \\
\bottomrule
\end{tabular}
}%
\end{table*}

\clearpage

\begin{table*}[t]
\centering
\caption{Agent benchmark MAE at 2--50\% retention. Values are mean $\pm$ sample standard deviation in percent; lower is better. Best means are \textbf{bold} and second-best means are \underline{underlined}.}
\vspace{4pt}
\label{tab:representative-mae-agent}
\footnotesize
\setlength{\tabcolsep}{2.8pt}
\renewcommand{\arraystretch}{1.00}
\newcommand{\maestd}[1]{\,{\scriptsize\color{black!70}#1}}
\newcommand{\benchrow}[1]{\multicolumn{8}{l}{\textcolor{blue!45!black}{\textbf{#1}}}}
\resizebox{\textwidth}{!}{%
\begin{tabular}{lrrrrrrr}
\toprule
& \multicolumn{7}{c}{\textbf{Average MAE (\%) $\downarrow$}\enspace {\scriptsize mean $\pm$ std.}} \\
\cmidrule(lr){2-8}
\rowcolor{lightBlue}\textbf{Method} & \textbf{2\%} & \textbf{5\%} & \textbf{10\%} & \textbf{15\%} & \textbf{20\%} & \textbf{50\%} & \textbf{Mean} \\
\midrule
\benchrow{Drafter-Bench} \\
SparseEval (Linear) & 23.58\maestd{$\pm$11.47} & 27.37\maestd{$\pm$13.35} & 41.79\maestd{$\pm$42.16} & 39.31\maestd{$\pm$40.39} & 17.49\maestd{$\pm$5.47} & 14.58\maestd{$\pm$4.61} & 27.35\maestd{$\pm$23.54} \\
SparseEval (MLP) & 11.92\maestd{$\pm$13.27} & 11.08\maestd{$\pm$12.11} & 12.94\maestd{$\pm$15.06} & 13.69\maestd{$\pm$14.96} & 13.94\maestd{$\pm$16.42} & 13.01\maestd{$\pm$15.91} & 12.77\maestd{$\pm$12.39} \\
Anchor Points & 3.06\maestd{$\pm$1.67} & 2.44\maestd{$\pm$0.40} & 2.12\maestd{$\pm$0.66} & 1.88\maestd{$\pm$0.30} & 1.48\maestd{$\pm$0.12} & 1.13\maestd{$\pm$0.07} & 2.02\maestd{$\pm$0.91} \\
TinyBenchmarks & 6.84\maestd{$\pm$2.08} & 3.66\maestd{$\pm$0.62} & 3.69\maestd{$\pm$0.55} & 3.79\maestd{$\pm$0.70} & 3.88\maestd{$\pm$0.67} & 3.26\maestd{$\pm$0.32} & 4.19\maestd{$\pm$1.50} \\
ZipBench & 3.96\maestd{$\pm$0.75} & \underline{1.61}\maestd{$\pm$0.16} & \textbf{1.09}\maestd{$\pm$0.16} & \textbf{0.84}\maestd{$\pm$0.21} & \textbf{0.53}\maestd{$\pm$0.12} & 0.45\maestd{$\pm$0.07} & 1.41\maestd{$\pm$1.27} \\
Random & \underline{2.91}\maestd{$\pm$0.85} & 1.80\maestd{$\pm$0.53} & 1.24\maestd{$\pm$0.36} & 0.98\maestd{$\pm$0.29} & 0.83\maestd{$\pm$0.24} & \underline{0.41}\maestd{$\pm$0.12} & \underline{1.36}\maestd{$\pm$0.93} \\
\rowcolor{lightBlue}\textbf{BestRand (ours)} & \textbf{2.60}\maestd{$\pm$0.60} & \textbf{1.60}\maestd{$\pm$0.36} & \underline{1.10}\maestd{$\pm$0.25} & \underline{0.87}\maestd{$\pm$0.19} & \underline{0.72}\maestd{$\pm$0.15} & \textbf{0.36}\maestd{$\pm$0.07} & \textbf{1.21}\maestd{$\pm$0.80} \\
\midrule
\benchrow{OSWorld} \\
SparseEval (Linear) & 14.41\maestd{$\pm$3.28} & 17.37\maestd{$\pm$2.78} & 18.95\maestd{$\pm$5.56} & 19.52\maestd{$\pm$5.86} & 16.75\maestd{$\pm$7.35} & 16.70\maestd{$\pm$1.96} & 17.28\maestd{$\pm$4.43} \\
SparseEval (MLP) & 15.82\maestd{$\pm$3.77} & 14.88\maestd{$\pm$2.14} & 16.93\maestd{$\pm$3.68} & 14.58\maestd{$\pm$3.25} & 12.86\maestd{$\pm$0.73} & 9.89\maestd{$\pm$1.44} & 14.16\maestd{$\pm$3.29} \\
Anchor Points & 13.59\maestd{$\pm$3.67} & 7.78\maestd{$\pm$1.33} & 6.16\maestd{$\pm$2.24} & 6.26\maestd{$\pm$1.78} & 6.38\maestd{$\pm$1.76} & 3.21\maestd{$\pm$0.72} & 7.23\maestd{$\pm$3.71} \\
TinyBenchmarks & 12.84\maestd{$\pm$1.28} & 8.77\maestd{$\pm$1.02} & 5.56\maestd{$\pm$1.23} & 4.02\maestd{$\pm$1.47} & 4.48\maestd{$\pm$0.95} & 1.83\maestd{$\pm$0.39} & 6.25\maestd{$\pm$3.82} \\
ZipBench & 13.74\maestd{$\pm$2.73} & \textbf{5.50}\maestd{$\pm$2.27} & \underline{5.18}\maestd{$\pm$1.22} & \underline{3.59}\maestd{$\pm$1.19} & 3.84\maestd{$\pm$0.93} & 1.84\maestd{$\pm$0.68} & \underline{5.62}\maestd{$\pm$4.18} \\
Random & \underline{12.78}\maestd{$\pm$3.97} & 7.85\maestd{$\pm$2.44} & 5.38\maestd{$\pm$1.68} & 4.26\maestd{$\pm$1.34} & \underline{3.58}\maestd{$\pm$1.12} & \underline{1.78}\maestd{$\pm$0.56} & 5.94\maestd{$\pm$4.17} \\
\rowcolor{lightBlue}\textbf{BestRand (ours)} & \textbf{10.71}\maestd{$\pm$2.56} & \underline{6.57}\maestd{$\pm$1.51} & \textbf{4.44}\maestd{$\pm$0.98} & \textbf{3.49}\maestd{$\pm$0.76} & \textbf{2.87}\maestd{$\pm$0.61} & \textbf{1.40}\maestd{$\pm$0.29} & \textbf{4.91}\maestd{$\pm$3.31} \\
\midrule
\benchrow{BFCL} \\
SparseEval (Linear) & 24.27\maestd{$\pm$4.48} & 18.69\maestd{$\pm$1.99} & 25.90\maestd{$\pm$9.78} & 17.41\maestd{$\pm$1.88} & 19.55\maestd{$\pm$1.36} & 20.48\maestd{$\pm$4.04} & 21.05\maestd{$\pm$5.14} \\
SparseEval (MLP) & 14.63\maestd{$\pm$7.26} & 13.73\maestd{$\pm$8.13} & 13.29\maestd{$\pm$7.43} & 11.69\maestd{$\pm$6.03} & 12.56\maestd{$\pm$8.35} & 12.68\maestd{$\pm$5.74} & 13.10\maestd{$\pm$6.14} \\
Anchor Points & 7.39\maestd{$\pm$1.82} & 4.66\maestd{$\pm$1.04} & 5.60\maestd{$\pm$1.08} & 5.17\maestd{$\pm$0.26} & 5.98\maestd{$\pm$1.15} & 4.09\maestd{$\pm$0.70} & 5.48\maestd{$\pm$1.43} \\
TinyBenchmarks & \underline{3.25}\maestd{$\pm$1.09} & \textbf{1.77}\maestd{$\pm$0.14} & 1.73\maestd{$\pm$0.73} & 1.30\maestd{$\pm$0.56} & 1.05\maestd{$\pm$0.26} & 0.66\maestd{$\pm$0.04} & \underline{1.62}\maestd{$\pm$0.98} \\
ZipBench & 3.58\maestd{$\pm$0.52} & 2.27\maestd{$\pm$0.55} & 1.57\maestd{$\pm$0.46} & \underline{1.20}\maestd{$\pm$0.34} & \textbf{0.80}\maestd{$\pm$0.25} & 0.61\maestd{$\pm$0.05} & 1.67\maestd{$\pm$1.09} \\
Random & 3.58\maestd{$\pm$1.34} & 2.22\maestd{$\pm$0.83} & \underline{1.53}\maestd{$\pm$0.58} & 1.21\maestd{$\pm$0.46} & 1.02\maestd{$\pm$0.39} & \underline{0.51}\maestd{$\pm$0.19} & 1.68\maestd{$\pm$1.24} \\
\rowcolor{lightBlue}\textbf{BestRand (ours)} & \textbf{2.98}\maestd{$\pm$0.81} & \underline{1.84}\maestd{$\pm$0.49} & \textbf{1.25}\maestd{$\pm$0.32} & \textbf{0.97}\maestd{$\pm$0.24} & \underline{0.80}\maestd{$\pm$0.19} & \textbf{0.39}\maestd{$\pm$0.09} & \textbf{1.37}\maestd{$\pm$0.95} \\
\midrule
\benchrow{TauBench} \\
SparseEval (Linear) & \underline{17.81}\maestd{$\pm$3.91} & 36.39\maestd{$\pm$33.85} & 24.90\maestd{$\pm$5.52} & 19.37\maestd{$\pm$4.48} & 20.51\maestd{$\pm$1.95} & 20.72\maestd{$\pm$3.45} & 23.28\maestd{$\pm$13.63} \\
SparseEval (MLP) & \textbf{14.85}\maestd{$\pm$1.05} & 15.01\maestd{$\pm$2.02} & 12.98\maestd{$\pm$1.97} & 13.04\maestd{$\pm$3.90} & 10.59\maestd{$\pm$2.17} & 10.16\maestd{$\pm$1.69} & 12.77\maestd{$\pm$2.73} \\
Anchor Points & 22.44\maestd{$\pm$2.50} & 14.86\maestd{$\pm$2.42} & 10.95\maestd{$\pm$4.18} & 7.95\maestd{$\pm$1.23} & 7.32\maestd{$\pm$1.55} & 6.93\maestd{$\pm$0.95} & 11.74\maestd{$\pm$6.01} \\
TinyBenchmarks & 22.46\maestd{$\pm$5.51} & 13.57\maestd{$\pm$3.74} & 9.68\maestd{$\pm$3.10} & 7.51\maestd{$\pm$2.17} & 6.63\maestd{$\pm$1.51} & 2.98\maestd{$\pm$0.53} & 10.47\maestd{$\pm$6.96} \\
ZipBench & 18.39\maestd{$\pm$5.60} & 15.18\maestd{$\pm$4.23} & 11.08\maestd{$\pm$3.24} & 6.96\maestd{$\pm$2.07} & 7.77\maestd{$\pm$2.09} & 3.56\maestd{$\pm$0.96} & 10.49\maestd{$\pm$5.93} \\
Random & 21.55\maestd{$\pm$6.52} & \underline{13.02}\maestd{$\pm$3.80} & \underline{8.62}\maestd{$\pm$2.57} & \underline{6.91}\maestd{$\pm$2.05} & \underline{5.84}\maestd{$\pm$1.74} & \underline{2.90}\maestd{$\pm$0.86} & \underline{9.81}\maestd{$\pm$6.99} \\
\rowcolor{lightBlue}\textbf{BestRand (ours)} & 19.16\maestd{$\pm$5.39} & \textbf{11.72}\maestd{$\pm$2.74} & \textbf{7.64}\maestd{$\pm$1.71} & \textbf{6.11}\maestd{$\pm$1.33} & \textbf{5.10}\maestd{$\pm$1.07} & \textbf{2.52}\maestd{$\pm$0.51} & \textbf{8.71}\maestd{$\pm$6.06} \\
\midrule
\benchrow{NexusBench} \\
SparseEval (Linear) & 28.04\maestd{$\pm$23.24} & 22.53\maestd{$\pm$4.64} & 23.56\maestd{$\pm$7.45} & 22.86\maestd{$\pm$5.36} & 20.68\maestd{$\pm$2.42} & 21.42\maestd{$\pm$3.79} & 23.18\maestd{$\pm$9.18} \\
SparseEval (MLP) & 11.39\maestd{$\pm$1.05} & 9.70\maestd{$\pm$0.65} & 8.51\maestd{$\pm$0.97} & 8.16\maestd{$\pm$0.18} & 7.57\maestd{$\pm$0.48} & 7.43\maestd{$\pm$0.79} & 8.79\maestd{$\pm$1.55} \\
Anchor Points & 10.94\maestd{$\pm$2.70} & 8.46\maestd{$\pm$2.62} & 8.96\maestd{$\pm$4.66} & 10.36\maestd{$\pm$2.03} & 9.41\maestd{$\pm$2.13} & 6.35\maestd{$\pm$0.86} & 9.08\maestd{$\pm$2.76} \\
TinyBenchmarks & \textbf{5.51}\maestd{$\pm$0.78} & \textbf{3.82}\maestd{$\pm$0.36} & \textbf{2.90}\maestd{$\pm$0.83} & \underline{2.56}\maestd{$\pm$0.45} & 2.32\maestd{$\pm$0.67} & 1.45\maestd{$\pm$0.16} & \textbf{3.09}\maestd{$\pm$1.42} \\
ZipBench & 8.96\maestd{$\pm$2.41} & 4.64\maestd{$\pm$1.11} & 3.75\maestd{$\pm$0.51} & 2.64\maestd{$\pm$0.29} & \underline{2.03}\maestd{$\pm$0.14} & 1.39\maestd{$\pm$0.57} & 3.90\maestd{$\pm$2.75} \\
Random & 8.20\maestd{$\pm$2.38} & 5.02\maestd{$\pm$1.48} & 3.45\maestd{$\pm$1.02} & 2.74\maestd{$\pm$0.81} & 2.30\maestd{$\pm$0.68} & \underline{1.15}\maestd{$\pm$0.34} & 3.81\maestd{$\pm$2.63} \\
\rowcolor{lightBlue}\textbf{BestRand (ours)} & \underline{7.21}\maestd{$\pm$1.73} & \underline{4.40}\maestd{$\pm$1.07} & \underline{2.99}\maestd{$\pm$0.72} & \textbf{2.34}\maestd{$\pm$0.56} & \textbf{1.95}\maestd{$\pm$0.46} & \textbf{0.97}\maestd{$\pm$0.22} & \underline{3.31}\maestd{$\pm$2.24} \\
\midrule
\benchrow{ComplexFuncBench} \\
SparseEval (Linear) & 31.30\maestd{$\pm$14.50} & 37.30\maestd{$\pm$12.37} & 16.94\maestd{$\pm$2.07} & 19.40\maestd{$\pm$3.77} & 18.43\maestd{$\pm$3.17} & 35.76\maestd{$\pm$28.61} & 26.52\maestd{$\pm$14.79} \\
SparseEval (MLP) & 13.50\maestd{$\pm$0.98} & 12.03\maestd{$\pm$3.50} & 10.72\maestd{$\pm$3.50} & 10.82\maestd{$\pm$2.75} & 12.35\maestd{$\pm$3.56} & 11.30\maestd{$\pm$1.94} & 11.79\maestd{$\pm$2.61} \\
Anchor Points & 10.36\maestd{$\pm$3.87} & 10.21\maestd{$\pm$3.02} & 10.88\maestd{$\pm$3.14} & 10.20\maestd{$\pm$1.22} & 9.08\maestd{$\pm$1.73} & 6.88\maestd{$\pm$1.02} & 9.60\maestd{$\pm$2.55} \\
TinyBenchmarks & \underline{7.59}\maestd{$\pm$1.17} & \textbf{3.45}\maestd{$\pm$0.41} & 3.56\maestd{$\pm$0.56} & \textbf{2.42}\maestd{$\pm$0.64} & \underline{2.27}\maestd{$\pm$0.66} & 1.39\maestd{$\pm$0.30} & \textbf{3.45}\maestd{$\pm$2.13} \\
ZipBench & 8.33\maestd{$\pm$1.87} & 7.10\maestd{$\pm$0.80} & 3.69\maestd{$\pm$0.65} & 3.30\maestd{$\pm$0.86} & 2.57\maestd{$\pm$0.28} & 1.54\maestd{$\pm$0.32} & 4.42\maestd{$\pm$2.64} \\
Random & 8.15\maestd{$\pm$2.07} & 5.05\maestd{$\pm$1.30} & \underline{3.48}\maestd{$\pm$0.90} & 2.76\maestd{$\pm$0.71} & 2.31\maestd{$\pm$0.60} & \underline{1.16}\maestd{$\pm$0.30} & 3.82\maestd{$\pm$2.54} \\
\rowcolor{lightBlue}\textbf{BestRand (ours)} & \textbf{7.50}\maestd{$\pm$1.57} & \underline{4.63}\maestd{$\pm$0.98} & \textbf{3.17}\maestd{$\pm$0.65} & \underline{2.50}\maestd{$\pm$0.50} & \textbf{2.09}\maestd{$\pm$0.42} & \textbf{1.04}\maestd{$\pm$0.20} & \underline{3.49}\maestd{$\pm$2.26} \\
\bottomrule
\end{tabular}
}%
\end{table*}

\clearpage

\clearpage

\subsection{Comparison with a Broader Set of Baselines}
\label{app:broader-baseline-comparison}

Table~\ref{tab:category-main-results} in Section~\ref{sec:experiments} compares BestRand with Random, two SparseEval variants, Anchor Points, TinyBenchmarks, and ZipBench across 70 benchmarks. Appendix~\ref{app:data-only-methods-effectiveness} further evaluates Facility Location, MIS, Agreement, and Coverage. Here, we combine these methods with five additional baselines. They are SubLIME~\citep{saranathan2025sublime}, MINCE~\citep{das2026mince}, EssenceBench~\citep{wang2026rethinking}, FlashEval~\citep{zhao2024flasheval}, and DatBench~\citep{joshi2026datbench}. We evaluate all 15 baselines and BestRand on 12 benchmarks. All methods use the same data splits and retention ratios.

The five additional baselines cover a diverse set of compression strategies. SubLIME predicts how well a subset preserves model rankings using anchor model responses and properties of benchmark items. MINCE uses Monte Carlo calibration to estimate the score drift of random subsets. EssenceBench combines semantic and response filtering with score prediction and subset search. FlashEval searches for subsets that preserve model rankings. DatBench selects items according to the point-biserial correlation between item responses and full benchmark scores.

\begin{figure*}[p]
\centering
\includegraphics[width=\textwidth,height=0.98\textheight,keepaspectratio]{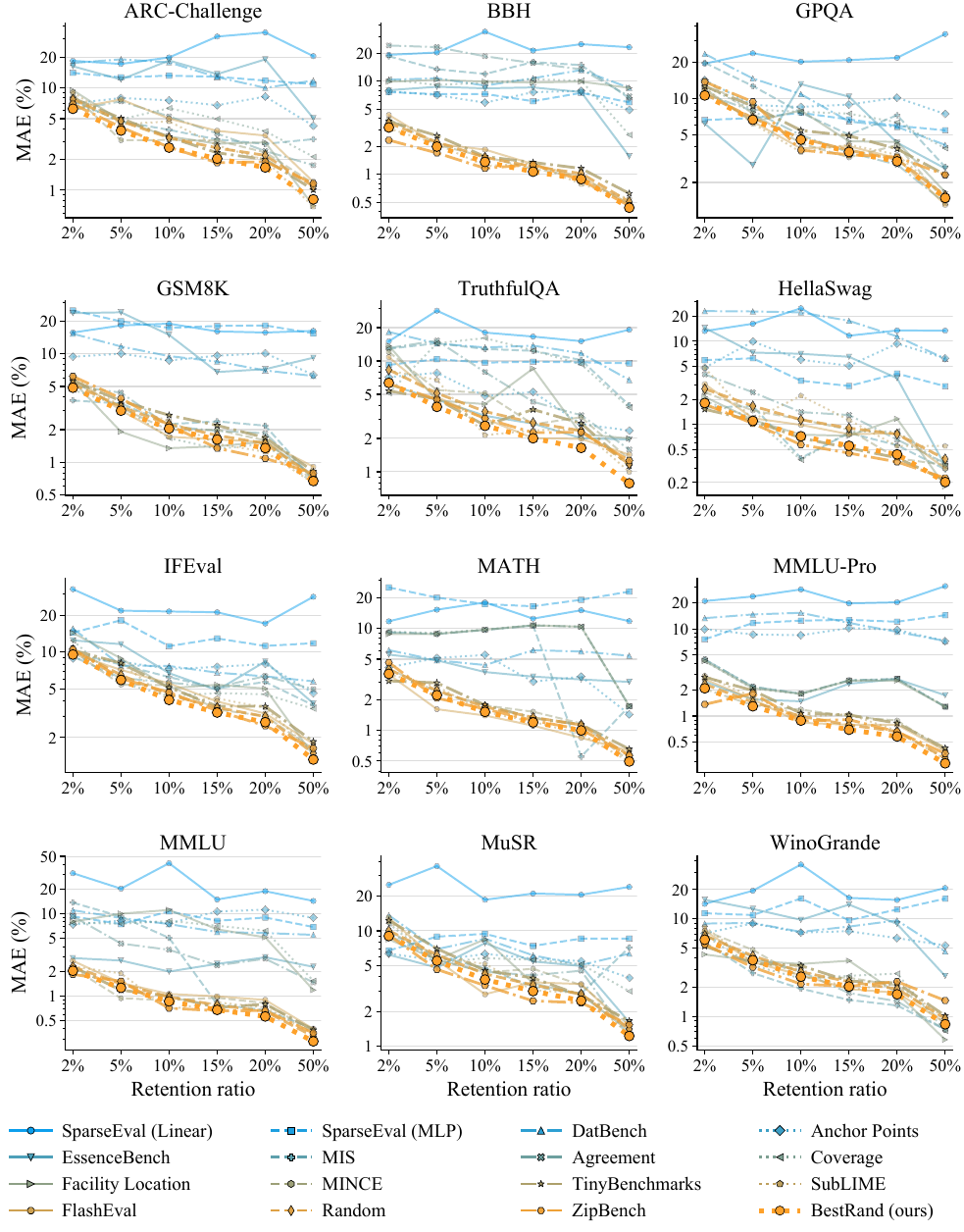}
\caption{MAE of BestRand and 15 baselines across six retention ratios on 12 benchmarks. Each panel shows one benchmark, with MAE plotted on a logarithmic scale. BestRand consistently achieves the lowest or near-lowest error across benchmarks and retention ratios.}
\label{fig:additional-baselines-curves}
\end{figure*}

Table~\ref{tab:additional-baselines} summarizes the results across 12 benchmarks and six retention ratios. Against 15 baselines, BestRand achieves the lowest overall MAE and ranks first or second at every retention ratio. Its advantage therefore holds from aggressive to mild compression. Figure~\ref{fig:additional-baselines-curves} further breaks down the comparison by benchmark. BestRand remains among the strongest methods across individual benchmarks, with its error decreasing steadily as more items are retained. These consistent trends show that the aggregate advantage holds broadly rather than arising from a few favorable benchmarks. This comparison complements our main study on 70 text, multimodal, and agent benchmarks, as well as our extension to VLA evaluation. Across the full study, we consider a wide range of retention ratios, evaluation metrics, model splits, construction budgets, and distribution shifts. We also compare methods from many baseline families and examine average accuracy, ranking fidelity, variation across seeds, high-error risk, construction cost, and OOD generalization. Additional ablations, data-only experiments, and multi-metric evaluations test the method under further settings and design choices. Together, these experiments provide evidence at the aggregate, category, and individual-benchmark levels. BestRand consistently delivers accurate and stable estimates at low construction cost across this extensive evaluation. The scale and diversity of the results show that its gains persist across benchmarks, compression settings, evaluation criteria, and competing methods.

\begin{table*}[t]
\centering
\caption{Compared with 15 baselines, BestRand delivers the strongest overall performance, achieving the lowest average MAE. Each entry reports the average MAE (\%) and sample standard deviation across benchmarks; the final column aggregates all benchmark--budget combinations. Lower is better; the best and second-best results are shown in \textbf{bold} and \underline{underlined}, respectively.}
\vspace{4pt}
\label{tab:additional-baselines}
\footnotesize
\setlength{\tabcolsep}{2.8pt}
\renewcommand{\arraystretch}{1.08}
\newcommand{\std}[1]{\,{\scriptsize\color{black!70}#1}}
\resizebox{\textwidth}{!}{%
\begin{tabular}{lrrrrrrr}
\toprule
& \multicolumn{7}{c}{\textbf{Average MAE (\%) $\downarrow$}\enspace{\scriptsize mean $\pm$ std.}} \\
\cmidrule(lr){2-8}
\rowcolor{lightBlue}
\textbf{Method} & \textbf{2\%} & \textbf{5\%} & \textbf{10\%} & \textbf{15\%} & \textbf{20\%} & \textbf{50\%} & \textbf{Mean} \\
\midrule
SparseEval (Linear) & 19.73\std{$\pm$6.78} & 21.72\std{$\pm$5.86} & 24.98\std{$\pm$8.09} & 18.65\std{$\pm$5.32} & 19.43\std{$\pm$5.88} & 21.39\std{$\pm$7.08} & 20.98\std{$\pm$6.67} \\
SparseEval (MLP) & 11.94\std{$\pm$6.77} & 11.72\std{$\pm$5.06} & 11.30\std{$\pm$4.28} & 10.32\std{$\pm$4.44} & 10.81\std{$\pm$4.47} & 10.90\std{$\pm$5.64} & 11.17\std{$\pm$5.03} \\
DatBench & 14.70\std{$\pm$5.29} & 12.12\std{$\pm$5.22} & 11.11\std{$\pm$5.12} & 9.57\std{$\pm$3.75} & 8.44\std{$\pm$2.78} & 6.54\std{$\pm$2.02} & 10.41\std{$\pm$4.86} \\
Anchor Points & 7.85\std{$\pm$2.60} & 8.04\std{$\pm$1.51} & 6.99\std{$\pm$1.27} & 7.32\std{$\pm$2.30} & 7.71\std{$\pm$2.73} & 5.25\std{$\pm$2.15} & 7.19\std{$\pm$2.29} \\
EssenceBench & 10.00\std{$\pm$6.47} & 8.14\std{$\pm$6.33} & 8.06\std{$\pm$5.31} & 6.75\std{$\pm$4.12} & 6.22\std{$\pm$4.67} & 2.97\std{$\pm$2.30} & 7.02\std{$\pm$5.35} \\
MIS & 9.71\std{$\pm$5.68} & 7.27\std{$\pm$4.53} & 5.70\std{$\pm$4.09} & 5.30\std{$\pm$4.99} & 4.30\std{$\pm$4.47} & 2.70\std{$\pm$2.33} & 5.83\std{$\pm$4.85} \\
Agreement & 9.28\std{$\pm$5.43} & 7.64\std{$\pm$6.03} & 5.69\std{$\pm$4.73} & 4.75\std{$\pm$4.24} & 4.38\std{$\pm$3.86} & 2.00\std{$\pm$1.93} & 5.62\std{$\pm$4.99} \\
Coverage & 7.82\std{$\pm$3.49} & 6.16\std{$\pm$3.70} & 6.18\std{$\pm$4.47} & 5.90\std{$\pm$3.85} & 5.30\std{$\pm$3.29} & 2.09\std{$\pm$1.25} & 5.57\std{$\pm$3.79} \\
Facility Location & 8.41\std{$\pm$3.83} & 5.68\std{$\pm$3.24} & 5.58\std{$\pm$3.67} & 5.00\std{$\pm$3.24} & 4.15\std{$\pm$3.09} & 1.78\std{$\pm$2.18} & 5.10\std{$\pm$3.71} \\
MINCE & 6.91\std{$\pm$4.14} & \underline{3.60}\std{$\pm$2.10} & 2.76\std{$\pm$1.62} & 2.25\std{$\pm$1.21} & 1.98\std{$\pm$1.20} & 0.88\std{$\pm$0.49} & 3.06\std{$\pm$2.81} \\
TinyBenchmarks & 6.07\std{$\pm$3.88} & 4.19\std{$\pm$2.56} & 2.76\std{$\pm$1.63} & 2.32\std{$\pm$1.41} & 1.93\std{$\pm$1.12} & 1.01\std{$\pm$0.64} & 3.04\std{$\pm$2.67} \\
SubLIME & 6.86\std{$\pm$3.55} & 3.97\std{$\pm$2.28} & 2.64\std{$\pm$1.47} & 2.11\std{$\pm$1.18} & 1.77\std{$\pm$1.10} & \underline{0.82}\std{$\pm$0.43} & 3.03\std{$\pm$2.72} \\
FlashEval & 6.76\std{$\pm$3.71} & 3.84\std{$\pm$2.39} & 2.56\std{$\pm$1.57} & 2.15\std{$\pm$1.30} & 1.89\std{$\pm$1.11} & 0.92\std{$\pm$0.49} & 3.02\std{$\pm$2.74} \\
Random & 6.18\std{$\pm$3.26} & 3.83\std{$\pm$2.03} & 2.63\std{$\pm$1.39} & 2.09\std{$\pm$1.10} & 1.76\std{$\pm$0.93} & 0.88\std{$\pm$0.46} & 2.89\std{$\pm$2.44} \\
ZipBench & \underline{5.84}\std{$\pm$3.85} & 3.66\std{$\pm$2.37} & \textbf{2.23}\std{$\pm$1.30} & \textbf{1.77}\std{$\pm$0.97} & \underline{1.60}\std{$\pm$0.93} & 1.01\std{$\pm$0.65} & \underline{2.68}\std{$\pm$2.54} \\
\rowcolor{lightBlue}
\textbf{BestRand (ours)} & \textbf{5.45}\std{$\pm$3.06} & \textbf{3.36}\std{$\pm$1.90} & \underline{2.29}\std{$\pm$1.30} & \underline{1.81}\std{$\pm$1.03} & \textbf{1.50}\std{$\pm$0.86} & \textbf{0.74}\std{$\pm$0.43} & \textbf{2.52}\std{$\pm$2.23} \\
\bottomrule
\end{tabular}%
}
\end{table*}

\clearpage

\begin{table*}[t]
\centering
\caption{Stability of Random and BestRand on the 12 benchmarks. Entries report mean MAE and sample standard deviation across seeds. Reduction rows show BestRand's relative reductions in mean MAE and standard deviation compared with Random, formatted as mean/std. Random and BestRand use 10{,}000 seeds. Across all settings, BestRand achieves both a lower mean MAE and a lower standard deviation than Random, demonstrating consistently better accuracy and stability.}
\vspace{4pt}
\label{tab:per-benchmark-seed-stability}
\small
\setlength{\tabcolsep}{4.5pt}
\renewcommand{\arraystretch}{1.28}
\newcommand{\seedvalue}[2]{#1\,{\scriptsize\color{black!70}{$\pm$#2}}}
\newcommand{\bestseedvalue}[2]{\cellcolor{lightBlue}\textbf{#1}\,{\scriptsize\color{black!70}{$\pm$#2}}}
\newcommand{\improvecell}[2]{\textcolor{blue!45!black}{\textbf{#1\%/#2\%}}}
\resizebox{\textwidth}{!}{%
\begin{tabular}{@{}llrrrrrr@{}}
\toprule
& & \multicolumn{6}{c}{\textbf{Mean Absolute Error (\%) $\downarrow$}\enspace{\scriptsize mean $\pm$ std.}} \\
\cmidrule(lr){3-8}
\rowcolor{lightBlue}
\textbf{Benchmark} & \textbf{Method} & \textbf{2\%} & \textbf{5\%} & \textbf{10\%} & \textbf{15\%} & \textbf{20\%} & \textbf{50\%} \\
\midrule
\multirow{3}{*}{ARC-Challenge} & Random & \seedvalue{7.70}{2.89} & \seedvalue{4.73}{1.78} & \seedvalue{3.27}{1.23} & \seedvalue{2.59}{0.98} & \seedvalue{2.18}{0.82} & \seedvalue{1.09}{0.41} \\
& \cellcolor{lightBlue}\textbf{BestRand} & \bestseedvalue{6.25}{1.66} & \bestseedvalue{3.82}{1.00} & \bestseedvalue{2.61}{0.64} & \bestseedvalue{2.03}{0.49} & \bestseedvalue{1.67}{0.38} & \bestseedvalue{0.81}{0.18} \\
& \textbf{Reduction} & \improvecell{18.9}{42.5} & \improvecell{19.1}{43.6} & \improvecell{20.2}{47.8} & \improvecell{21.5}{49.9} & \improvecell{23.6}{53.5} & \improvecell{25.2}{56.9} \\
\midrule
\multirow{3}{*}{BBH} & Random & \seedvalue{3.56}{0.92} & \seedvalue{2.22}{0.57} & \seedvalue{1.53}{0.39} & \seedvalue{1.21}{0.31} & \seedvalue{1.02}{0.26} & \seedvalue{0.51}{0.13} \\
& \cellcolor{lightBlue}\textbf{BestRand} & \bestseedvalue{3.19}{0.68} & \bestseedvalue{1.99}{0.42} & \bestseedvalue{1.36}{0.28} & \bestseedvalue{1.07}{0.22} & \bestseedvalue{0.89}{0.18} & \bestseedvalue{0.44}{0.09} \\
& \textbf{Reduction} & \improvecell{10.5}{26.4} & \improvecell{10.4}{26.6} & \improvecell{11.0}{29.1} & \improvecell{11.8}{30.0} & \improvecell{12.5}{32.9} & \improvecell{13.8}{34.2} \\
\midrule
\multirow{3}{*}{GPQA} & Random & \seedvalue{10.85}{2.39} & \seedvalue{6.87}{1.49} & \seedvalue{4.68}{1.01} & \seedvalue{3.71}{0.81} & \seedvalue{3.12}{0.68} & \seedvalue{1.56}{0.34} \\
& \cellcolor{lightBlue}\textbf{BestRand} & \bestseedvalue{10.56}{2.25} & \bestseedvalue{6.64}{1.36} & \bestseedvalue{4.53}{0.93} & \bestseedvalue{3.58}{0.74} & \bestseedvalue{3.00}{0.60} & \bestseedvalue{1.49}{0.30} \\
& \textbf{Reduction} & \improvecell{2.7}{5.8} & \improvecell{3.4}{8.8} & \improvecell{3.1}{8.6} & \improvecell{3.4}{8.3} & \improvecell{4.1}{11.5} & \improvecell{4.4}{11.1} \\
\midrule
\multirow{3}{*}{GSM8K} & Random & \seedvalue{5.27}{1.56} & \seedvalue{3.25}{0.97} & \seedvalue{2.24}{0.67} & \seedvalue{1.78}{0.53} & \seedvalue{1.49}{0.45} & \seedvalue{0.74}{0.22} \\
& \cellcolor{lightBlue}\textbf{BestRand} & \bestseedvalue{4.87}{1.23} & \bestseedvalue{2.99}{0.75} & \bestseedvalue{2.04}{0.50} & \bestseedvalue{1.62}{0.39} & \bestseedvalue{1.35}{0.32} & \bestseedvalue{0.67}{0.16} \\
& \textbf{Reduction} & \improvecell{7.6}{21.6} & \improvecell{8.0}{22.7} & \improvecell{8.7}{25.3} & \improvecell{8.7}{26.4} & \improvecell{9.3}{27.7} & \improvecell{9.8}{28.9} \\
\midrule
\multirow{3}{*}{HellaSwag} & Random & \seedvalue{2.67}{1.50} & \seedvalue{1.66}{0.93} & \seedvalue{1.14}{0.64} & \seedvalue{0.91}{0.51} & \seedvalue{0.76}{0.43} & \seedvalue{0.38}{0.21} \\
& \cellcolor{lightBlue}\textbf{BestRand} & \bestseedvalue{1.81}{0.74} & \bestseedvalue{1.10}{0.45} & \bestseedvalue{0.72}{0.27} & \bestseedvalue{0.55}{0.19} & \bestseedvalue{0.44}{0.14} & \bestseedvalue{0.20}{0.06} \\
& \textbf{Reduction} & \improvecell{32.2}{50.6} & \improvecell{33.4}{51.8} & \improvecell{37.0}{57.7} & \improvecell{39.1}{61.9} & \improvecell{42.8}{67.7} & \improvecell{46.8}{74.0} \\
\midrule
\multirow{3}{*}{IFEval} & Random & \seedvalue{10.41}{2.53} & \seedvalue{6.49}{1.62} & \seedvalue{4.47}{1.11} & \seedvalue{3.54}{0.89} & \seedvalue{2.97}{0.75} & \seedvalue{1.48}{0.37} \\
& \cellcolor{lightBlue}\textbf{BestRand} & \bestseedvalue{9.55}{1.96} & \bestseedvalue{5.92}{1.21} & \bestseedvalue{4.06}{0.82} & \bestseedvalue{3.20}{0.64} & \bestseedvalue{2.67}{0.53} & \bestseedvalue{1.32}{0.26} \\
& \textbf{Reduction} & \improvecell{8.2}{22.5} & \improvecell{8.8}{25.0} & \improvecell{9.1}{26.7} & \improvecell{9.6}{28.3} & \improvecell{10.2}{29.5} & \improvecell{10.8}{30.3} \\
\midrule
\multirow{3}{*}{MATH} & Random & \seedvalue{3.97}{1.18} & \seedvalue{2.45}{0.73} & \seedvalue{1.70}{0.50} & \seedvalue{1.34}{0.39} & \seedvalue{1.13}{0.33} & \seedvalue{0.56}{0.17} \\
& \cellcolor{lightBlue}\textbf{BestRand} & \bestseedvalue{3.58}{0.85} & \bestseedvalue{2.21}{0.53} & \bestseedvalue{1.52}{0.35} & \bestseedvalue{1.20}{0.27} & \bestseedvalue{1.00}{0.22} & \bestseedvalue{0.49}{0.11} \\
& \textbf{Reduction} & \improvecell{9.8}{28.2} & \improvecell{9.8}{26.8} & \improvecell{10.6}{30.2} & \improvecell{10.6}{30.9} & \improvecell{11.6}{33.1} & \improvecell{12.4}{33.5} \\
\midrule
\multirow{3}{*}{MMLU} & Random & \seedvalue{2.27}{0.75} & \seedvalue{1.41}{0.47} & \seedvalue{0.97}{0.32} & \seedvalue{0.77}{0.26} & \seedvalue{0.65}{0.21} & \seedvalue{0.32}{0.11} \\
& \cellcolor{lightBlue}\textbf{BestRand} & \bestseedvalue{2.02}{0.53} & \bestseedvalue{1.26}{0.33} & \bestseedvalue{0.86}{0.22} & \bestseedvalue{0.68}{0.17} & \bestseedvalue{0.56}{0.14} & \bestseedvalue{0.28}{0.07} \\
& \textbf{Reduction} & \improvecell{10.8}{28.8} & \improvecell{11.1}{29.6} & \improvecell{11.7}{32.1} & \improvecell{12.3}{33.5} & \improvecell{13.2}{34.8} & \improvecell{13.4}{36.2} \\
\midrule
\multirow{3}{*}{MMLU-Pro} & Random & \seedvalue{2.31}{0.75} & \seedvalue{1.44}{0.47} & \seedvalue{0.99}{0.32} & \seedvalue{0.78}{0.25} & \seedvalue{0.66}{0.21} & \seedvalue{0.33}{0.11} \\
& \cellcolor{lightBlue}\textbf{BestRand} & \bestseedvalue{2.08}{0.55} & \bestseedvalue{1.29}{0.34} & \bestseedvalue{0.89}{0.23} & \bestseedvalue{0.70}{0.18} & \bestseedvalue{0.58}{0.15} & \bestseedvalue{0.29}{0.07} \\
& \textbf{Reduction} & \improvecell{10.2}{27.2} & \improvecell{10.1}{26.9} & \improvecell{10.5}{28.5} & \improvecell{11.3}{29.8} & \improvecell{11.9}{31.7} & \improvecell{13.3}{35.3} \\
\midrule
\multirow{3}{*}{MuSR} & Random & \seedvalue{10.03}{2.88} & \seedvalue{6.16}{1.79} & \seedvalue{4.24}{1.24} & \seedvalue{3.38}{0.99} & \seedvalue{2.84}{0.83} & \seedvalue{1.42}{0.42} \\
& \cellcolor{lightBlue}\textbf{BestRand} & \bestseedvalue{9.01}{2.05} & \bestseedvalue{5.51}{1.25} & \bestseedvalue{3.79}{0.84} & \bestseedvalue{3.01}{0.66} & \bestseedvalue{2.49}{0.53} & \bestseedvalue{1.23}{0.26} \\
& \textbf{Reduction} & \improvecell{10.1}{28.9} & \improvecell{10.6}{30.2} & \improvecell{10.7}{32.5} & \improvecell{10.9}{33.4} & \improvecell{12.3}{36.2} & \improvecell{13.1}{37.9} \\
\midrule
\multirow{3}{*}{TruthfulQA} & Random & \seedvalue{8.33}{3.52} & \seedvalue{5.12}{2.20} & \seedvalue{3.52}{1.52} & \seedvalue{2.79}{1.20} & \seedvalue{2.35}{1.01} & \seedvalue{1.17}{0.51} \\
& \cellcolor{lightBlue}\textbf{BestRand} & \bestseedvalue{6.34}{1.96} & \bestseedvalue{3.85}{1.13} & \bestseedvalue{2.60}{0.74} & \bestseedvalue{2.00}{0.53} & \bestseedvalue{1.64}{0.42} & \bestseedvalue{0.79}{0.18} \\
& \textbf{Reduction} & \improvecell{23.9}{44.3} & \improvecell{24.8}{48.5} & \improvecell{26.3}{51.3} & \improvecell{28.2}{55.8} & \improvecell{30.1}{58.6} & \improvecell{32.9}{63.6} \\
\midrule
\multirow{3}{*}{WinoGrande} & Random & \seedvalue{6.77}{1.98} & \seedvalue{4.20}{1.23} & \seedvalue{2.87}{0.84} & \seedvalue{2.28}{0.67} & \seedvalue{1.91}{0.56} & \seedvalue{0.96}{0.28} \\
& \cellcolor{lightBlue}\textbf{BestRand} & \bestseedvalue{6.12}{1.44} & \bestseedvalue{3.77}{0.89} & \bestseedvalue{2.55}{0.58} & \bestseedvalue{2.03}{0.47} & \bestseedvalue{1.69}{0.38} & \bestseedvalue{0.83}{0.18} \\
& \textbf{Reduction} & \improvecell{9.6}{27.1} & \improvecell{10.1}{28.2} & \improvecell{10.9}{31.2} & \improvecell{11.0}{30.5} & \improvecell{11.9}{32.8} & \improvecell{13.0}{36.0} \\
\bottomrule
\end{tabular}%
}
\end{table*}

Table~\ref{tab:per-benchmark-seed-stability} presents the accuracy and stability of Random and BestRand. Each entry reports the mean and standard deviation of MAE across independent seeds. BestRand achieves both a lower mean error and a lower standard deviation than Random in every setting, demonstrating consistently higher accuracy and greater stability across benchmarks and compression levels. In particular, the lower standard deviation demonstrates that BestRand is consistently more stable across random seeds. These results are consistent with Theorem~\ref{thm:3}, which shows that probe-guided matching reduces the probe-explained component of the common-shift variance. The consistent improvements across all settings provide strong empirical support for the proposed variance-reduction mechanism.

\clearpage

\section{Full Proofs and Derivations}\label{app:proofs}

\subsection{Notation and basic conventions}\label{app:notation}

Fix the score matrix $X\in[0,1]^{N\times M}$, where $X_{s,m}$ is the score of model $m$ on item $s$ over $N$ items and $M$ models. In the finite-population analysis, $X$ is non-random and randomness comes only from subset sampling. A single subset $\mathcal S$ is drawn uniformly without replacement from $\{1,\dots,N\}$, with $|\mathcal S|=k$ and $1\le k\le N$, where $N>1$. The non-degenerate distributional analysis below concerns $k<N$; at $k=N$, all errors vanish. The selection analysis additionally considers multiple candidate draws. Following the main text,
\begin{equation}
\mu_m=\frac1N\sum_{s=1}^N X_{s,m},\quad
\hat\mu_m=\frac1k\sum_{s\in\mathcal S}X_{s,m},\quad
E_m=\hat\mu_m-\mu_m,\quad
\mathrm{MAE}=\frac1M\sum_{m=1}^M|E_m|,
\end{equation}
and the errors are collected into $E=(E_1,\dots,E_M)^\top$. Item difficulty and its empirical mean and variance are
\begin{equation}
g_s=\frac1M\sum_{m=1}^M X_{s,m},\qquad
\bar g_N=\frac1N\sum_{s=1}^N g_s,\qquad
\mathrm{Var}_N(g)=\frac1N\sum_{s=1}^N\big(g_s-\bar g_N\big)^2 .
\end{equation}
The common shift is $\bar E=\frac1M\sum_m E_m$ and the finite-population factor is $c(k)=\sqrt{(N-k)/[k(N-1)]}>0$ with $c(N)=0$.

Two variance symbols must be kept apart. A subscript $N$ denotes an empirical average over the fixed population of $N$ items and carries no randomness, whereas $\mathrm{Var}(\cdot)$ and $\mathrm{Cov}(\cdot)$ without a subscript denote sampling variance and covariance taken as the subset $\mathcal S$ is redrawn. We write $\Phi$ for the standard normal cumulative distribution function and $\varphi(t)=\frac1{\sqrt{2\pi}}e^{-t^2/2}$ for its density.

Let $z_s=X_{s,\cdot}-\mu\in\mathbb R^M$ be the $s$-th row of $X$ with the column means $\mu=(\mu_1,\dots,\mu_M)^\top$ removed. Because each column is centered,
\begin{equation}
\sum_{s=1}^N z_s=\sum_{s=1}^N X_{s,\cdot}-N\mu=0. \tag{A.1}\label{eq:proof-A.1}
\end{equation}
In line with Assumption~\ref{assump:1}, the population score covariance matrix is
\begin{equation}
C_{ml}=\frac1N\sum_{s=1}^N\big(X_{s,m}-\mu_m\big)\big(X_{s,l}-\mu_l\big),\qquad
C=\frac1N\sum_{s=1}^N z_s z_s^\top , \tag{A.2}\label{eq:proof-A.2}
\end{equation}
which is positive semidefinite since $a^\top C a=\frac1N\sum_s(a^\top z_s)^2\ge0$ for every $a\in\mathbb R^M$. Finally, let $I_s=\mathbf 1\{s\in\mathcal S\}$ be the sampling indicators, so that $\sum_{s=1}^N I_s=k$.

The matrix $C$ is the population score covariance, whereas the sampling error covariance is $\mathrm{Cov}(E)=c(k)^2C$. 
Theorem~\ref{thm:2} uses $C_0$ for the scalar constant $C_0=\sqrt{\pi/2-1}\,u^{*}\approx0.578$.
The subscript distinguishes it from the covariance matrix $C$.
To keep the limiting-density prefactor separate from both, Appendix~\ref{app:thm2} denotes it by $A_0$. Exact finite-population identities are distinguished below from distributional conclusions obtained within the Gaussian approximation of Theorem~\ref{thm:1}.

\subsection{Numerical validation of the assumptions}\label{app:assumption-validation}

The empirical checks concern how well Assumptions~\ref{assump:1} and~\ref{assump:2} approximate the benchmark data. Assumption~\ref{assump:1} posits equal positive diagonal entries $\sigma^2$ and equal off-diagonal entries $\rho\sigma^2$ in $C$, with $\rho\in[0,1)$. The empirical summaries used in the main text are $\sigma^2=M^{-1}\sum_m C_{mm}$ and, for $M>1$, $\rho=[M(M-1)]^{-1}\sum_{m\ne l}C_{ml}/\sigma^2$. Under equal variances, the latter is the average pairwise correlation; with unequal variances it is a normalized average covariance. Assumption~\ref{assump:2} requires $\mathrm{Var}_N(g)>0$ and positive explained variance $R^2>0$ in a least-squares regression with an intercept. The normal equations ensure that the residual has zero empirical mean and is orthogonal to the probes; they do not require a perfect linear fit or independence of residuals and probe scores.



\paragraph{Validation of Assumption~\ref{assump:1}.}
We validate Assumption~\ref{assump:1} by testing the equal-variance and equal-covariance conditions directly from the observed model scores. For each benchmark, we remove only models whose scores are constant across all items, as required by Assumption~\ref{assump:1}. We then center each remaining model over the $N$ items and form the empirical covariance matrix
\begin{equation*}
C=\frac{1}{N}Z^\top Z,
\qquad Z_{sm}=X_{s,m}-\mu_m .
\end{equation*}

Let $d=(C_{11},\ldots,C_{MM})$ contain the score variances of the $M$ models. Let $o$ contain the $K=M(M-1)/2$ covariances between unordered model pairs. We compute
\begin{equation*}
A_{\rm diag}=100\frac{\langle d,\mathbf 1\rangle}{\lVert d\rVert_2\sqrt M},
\qquad
P_{+}=\frac{100}{K}\sum_{m<l}\mathbf 1\{C_{ml}>0\},
\qquad
A_{\rm off}=100\frac{\langle o,\mathbf 1\rangle}{\lVert o\rVert_2\sqrt K}.
\end{equation*}
$A_{\rm diag}$ measures variance similarity. $P_{+}$ is the proportion of model pairs with positive covariance. $A_{\rm off}$ measures covariance similarity. The two alignment scores are cosine similarities to a positive constant vector. They reach 100\% only when the corresponding entries are equal and positive. Higher values therefore indicate closer agreement with Assumption~\ref{assump:1}.

Table~\ref{tab:app-assumption1-alignment} shows that score variances are highly similar across models and that almost all model pairs have positive covariance. The pairwise covariance values also remain close to a common level for most benchmarks. This agreement in both variance and covariance strongly supports Assumption~\ref{assump:1} as a compact approximation of the observed covariance structure.

\begin{table}[t]
\centering
\caption{Empirical results strongly support Assumption~\ref{assump:1}. Across diverse benchmarks, model-score variances are highly similar, while pairwise model-score covariances are overwhelmingly positive and close to a common value. $A_{\rm diag}$ measures variance similarity, $P_{+}$ is the proportion of positive pairwise covariances, and $A_{\rm off}$ measures covariance similarity. All are percentages; higher values indicate closer agreement with the assumption, and 100\% indicates perfect agreement.}
\vspace{3pt}
\label{tab:app-assumption1-alignment}
\begingroup
\normalsize
\setlength{\tabcolsep}{5pt}
\renewcommand{\arraystretch}{1.00}
\begin{tabular}{@{}lccc@{\hspace{5pt}\vrule width 0.35pt\hspace{5pt}}lccc@{}}
\toprule
\textbf{Benchmark} & $\mathbf{A}_{\mathrm{diag}}\!\uparrow$ & $\mathbf{P}_{+}\!\uparrow$ & $\mathbf{A}_{\mathrm{off}}\!\uparrow$ & \textbf{Benchmark} & $\mathbf{A}_{\mathrm{diag}}\!\uparrow$ & $\mathbf{P}_{+}\!\uparrow$ & $\mathbf{A}_{\mathrm{off}}\!\uparrow$ \\
\midrule
ACEBench & 98.8 & 100.0 & 99.2 & MathVista & 94.9 & 100.0 & 95.3 \\
AGC-Bench & 98.7 & 100.0 & 94.9 & MedQA & 99.3 & 100.0 & 92.8 \\
ARC-Challenge & 99.5 & 100.0 & 94.1 & MMMU & 97.4 & 100.0 & 95.3 \\
BBEH & 92.5 & 100.0 & 87.7 & MMMU-Pro & 98.9 & 100.0 & 92.0 \\
BBH & 99.8 & 96.4 & 85.1 & MMStar & 98.2 & 100.0 & 95.9 \\
Bench2Drive & 99.2 & 100.0 & 95.6 & MMVP & 96.9 & 100.0 & 95.1 \\
BFCL & 99.1 & 97.1 & 93.2 & MultiChallenge & 99.4 & 100.0 & 94.5 \\
BigCodeBench & 95.2 & 100.0 & 87.9 & MuSR & 99.9 & 99.8 & 93.2 \\
BoolQ & 95.6 & 97.9 & 91.8 & NexusBench & 99.5 & 100.0 & 90.6 \\
CommonsenseQA & 98.8 & 97.9 & 86.8 & OCRBench v2 & 99.7 & 100.0 & 99.5 \\
ComplexFuncBench & 96.5 & 99.5 & 88.6 & OpenBookQA & 96.4 & 97.2 & 72.3 \\
CyberSecEval 2 & 100.0 & 100.0 & 98.5 & OSWorld & 95.6 & 99.4 & 86.9 \\
DrafterBench & 94.1 & 99.5 & 94.2 & PAWS & 97.3 & 97.9 & 92.4 \\
DynaMath & 98.8 & 100.0 & 92.7 & Pre-Flight & 99.6 & 100.0 & 97.7 \\
Finance Fund. & 99.7 & 100.0 & 95.9 & RACE-H & 95.0 & 100.0 & 97.1 \\
GSM-Symbolic & 94.1 & 100.0 & 90.4 & RealWorldQA & 97.9 & 100.0 & 92.0 \\
HellaSwag & 99.7 & 100.0 & 97.9 & SciBench & 98.8 & 100.0 & 92.8 \\
HumanEval & 97.1 & 100.0 & 92.0 & SciCode & 96.7 & 100.0 & 95.6 \\
IFEval & 96.8 & 97.5 & 86.0 & tau-bench & 99.0 & 99.7 & 91.1 \\
InFoBench & 99.3 & 100.0 & 98.8 & Terminal-Bench & 98.0 & 99.3 & 90.9 \\
LiveCodeBench & 99.0 & 100.0 & 96.3 & TheoremQA & 98.8 & 100.0 & 94.2 \\
LogicVista & 99.7 & 99.6 & 89.6 & Toolathlon & 94.2 & 98.1 & 87.4 \\
LogiQA & 98.5 & 100.0 & 97.7 & TruthfulQA & 99.6 & 100.0 & 95.7 \\
LongBench v2 & 99.3 & 95.6 & 86.9 & VisualPuzzles & 99.8 & 100.0 & 94.9 \\
MathVerse & 96.8 & 98.0 & 84.3 & WinoGrande & 98.0 & 97.7 & 90.6 \\
\midrule
\multicolumn{8}{c}{
  \textit{Mean:}\qquad
  $A_{\rm diag}=97.9\%$\qquad
  $P_{+}=99.4\%$\qquad
  $A_{\rm off}=92.4\%$
} \\
\bottomrule
\end{tabular}
\endgroup
\end{table}

\begin{figure*}[h]
\centering
\includegraphics[width=\textwidth,height=0.98\textheight,keepaspectratio]{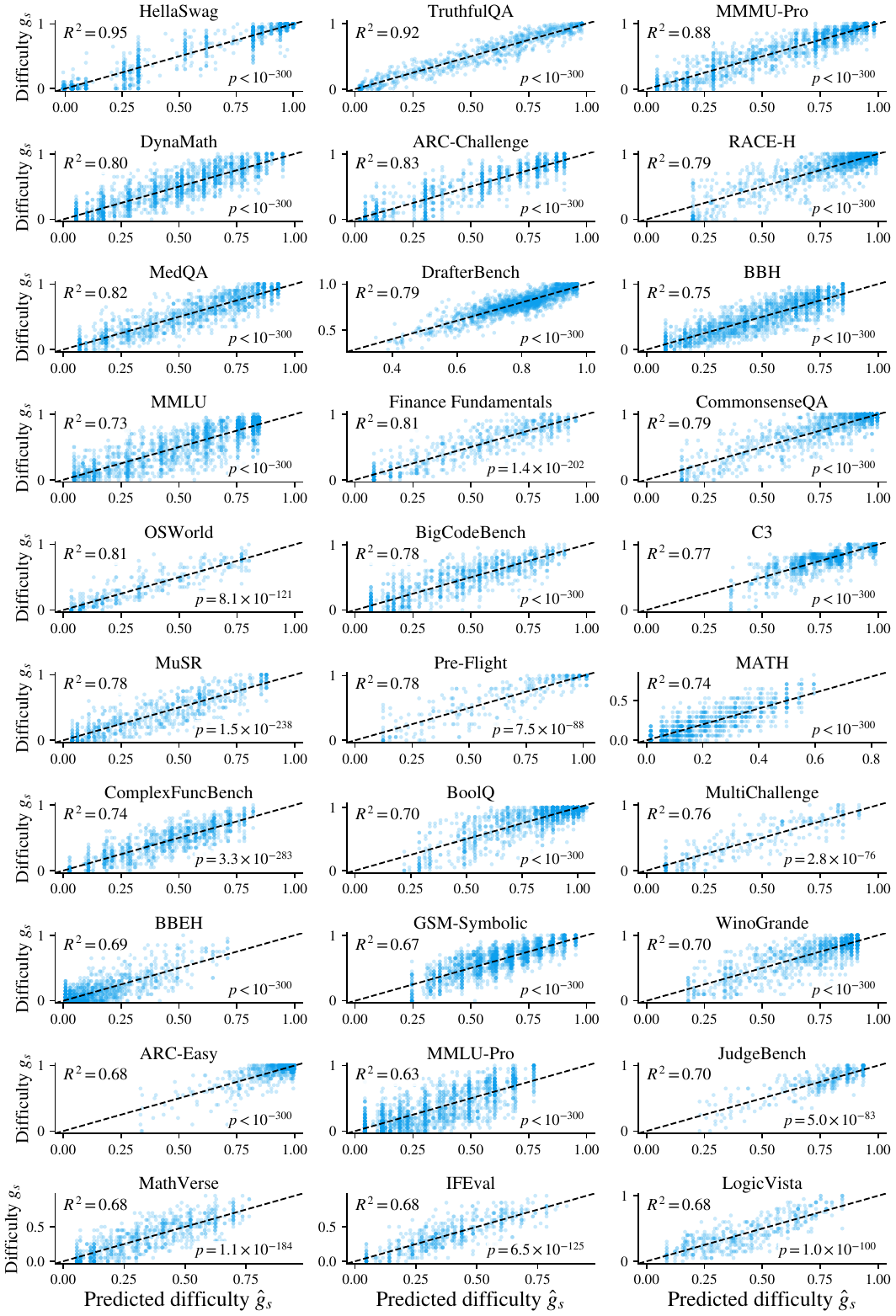}
\vspace{-18pt}
\caption{Item-level numerical validation on 30 benchmarks using ten probe models. The panels show a subset of the evaluated benchmarks, while aggregate results across all benchmarks are reported in Figure~\ref{fig:app-assumption2}. Across the displayed benchmarks, $R^2$ is consistently well above zero, all $p$-values are small, and every regression is statistically significant. The dashed line denotes equality.}
\label{fig:app-assumption2-item-regressions}
\end{figure*}

\clearpage

\paragraph{Validation of Assumption~\ref{assump:2}.}
We test whether probe scores explain item difficulty measured across the test models, which are disjoint from the probe models. Let $\mathcal T$ be the set of test models and $\mathcal P_p$ be a set of $p$ probe models, with $\mathcal T\cap\mathcal P_p=\varnothing$. We define item difficulty as $g_s=|\mathcal T|^{-1}\sum_{m\in\mathcal T}X_{s,m}$ and vary the number of probes from one to ten. Since no probe score enters $g_s$, the measured association cannot result from overlap between the predictors and the response. A positive $R^2$ therefore shows that the probes capture difficulty shared with unseen models.

\begin{figure}[t]\centering
\includegraphics[width=0.999\columnwidth]{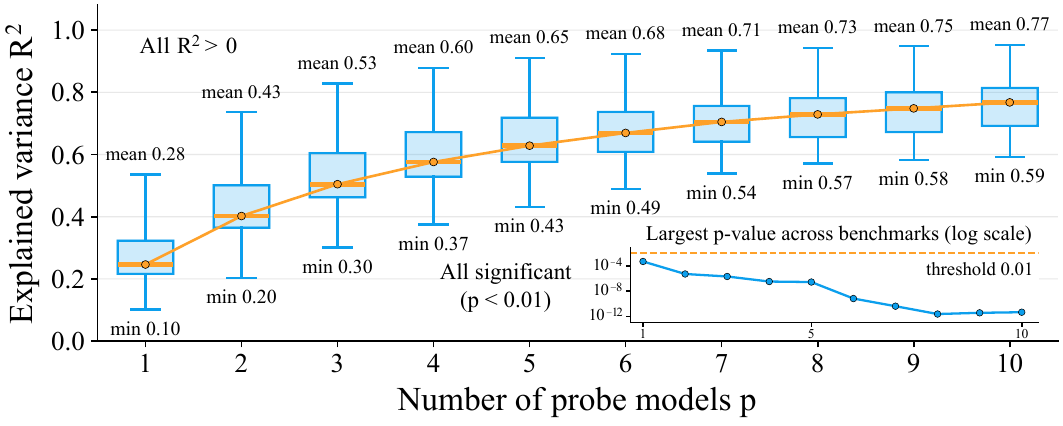}
\vspace{-20pt}
\caption{Numerical experiments consistently validate Assumption~\ref{assump:2}.
Each box summarizes the distribution of explained variance $R^2$ across benchmarks for a given number of probe models, with whiskers spanning the full observed range. The small plot in the lower-right corner reports the largest benchmark-level $p$-value at each probe count. All $R^2$ values are positive and all benchmark-level results are statistically significant, providing strong empirical support for the assumption.}
\label{fig:app-assumption2}\end{figure}

For each probe set, we regress $g_s$ on the probe scores by ordinary least squares with an intercept. Repeated probe choices are summarized within each benchmark, and all benchmarks receive equal weight. We report $R^2$ and use the largest $p$-value across benchmark summaries as a conservative significance check. 
Figure~\ref{fig:app-assumption2-item-regressions} shows that predictions from ten probes closely track the difficulty defined by the disjoint test models at the item level. The consistent alignment with the equality line shows that the explained variance reflects a broad item-level relation rather than a small number of items.
Figure~\ref{fig:app-assumption2} further shows that $R^2$ increases steadily with the number of probes, with smaller gains at larger probe counts. All $R^2$ values are positive, all benchmark summaries are significant, and most $R^2$ values are well above zero. 
Together, these results strongly validate Assumption~\ref{assump:2} across the evaluated benchmarks. By Theorem~\ref{thm:3}, the positive explained variance allows probe matching to reduce common-shift variance, predicting greater stability for BestRand than for Random.

\subsection{Basic lemmas: moments of sampling without replacement}\label{app:moments}

The two lemmas below are the only source of every variance and covariance computed later.

\medskip
\noindent\theorylabel{lem:B.1}{B.1}\textbf{Lemma B.1.}
Under uniform sampling without replacement, for any $s\ne t$,
\begin{equation}
\mathbb E[I_s]=\frac kN,\qquad
\mathrm{Var}(I_s)=\frac{k(N-k)}{N^2},\qquad
\mathrm{Cov}(I_s,I_t)=-\frac{k(N-k)}{N^2(N-1)} . \tag{B.1}\label{eq:proof-B.1}
\end{equation}

\noindent\textbf{Proof.}
Every item is equally likely to be chosen and every pair is equally likely to be chosen together,
\begin{equation*}
\Pr(s\in\mathcal S)=\frac kN,\qquad
\Pr(s,t\in\mathcal S)=\frac{k(k-1)}{N(N-1)}\quad(s\ne t).
\end{equation*}
The first probability holds because the $k$ chosen positions are spread uniformly over $N$ items, and the second because fixing two given items to be chosen amounts to choosing the remaining $k-2$ from the other $N-2$, with probability $\binom{N-2}{k-2}/\binom Nk=\frac{k(k-1)}{N(N-1)}$. Hence $\mathbb E[I_s]=k/N$, and since $I_s\in\{0,1\}$ gives $I_s^2=I_s$,
\begin{equation*}
\mathrm{Var}(I_s)=\mathbb E[I_s]-\mathbb E[I_s]^2=\frac kN-\Big(\frac kN\Big)^2=\frac{k(N-k)}{N^2}.
\end{equation*}
For $s\ne t$,
\begin{equation*}
\mathrm{Cov}(I_s,I_t)=\Pr(s,t\in\mathcal S)-\Pr(s\in\mathcal S)\Pr(t\in\mathcal S)=\frac{k(k-1)}{N(N-1)}-\frac{k^2}{N^2}.
\end{equation*}
Over a common denominator the numerator is $k[(k-1)N-k(N-1)]=k[kN-N-kN+k]=k(k-N)=-k(N-k)$, which gives $\mathrm{Cov}(I_s,I_t)=-k(N-k)/[N^2(N-1)]$. \hfill$\blacksquare$

\medskip
A compact form that we use repeatedly aligns the variance and the covariance,
\begin{equation}
\mathrm{Cov}(I_s,I_t)=\frac{k(N-k)}{N^2}\Big(\delta_{st}-\frac{1-\delta_{st}}{N-1}\Big),\qquad
\delta_{st}=\mathbf 1\{s=t\}. \tag{B.2}\label{eq:proof-B.2}
\end{equation}

\medskip
\noindent\theorylabel{lem:B.2}{B.2}\textbf{Lemma B.2.}
Let $h,h':\{1,\dots,N\}\to\mathbb R$ be any two per-item scalars, write the subset-minus-population mean deviation as $\Delta\bar h:=\bar h_{\mathcal S}-\bar h_N$ with $\bar h_{\mathcal S}=\frac1k\sum_{s\in\mathcal S}h_s$ and $\bar h_N=\frac1N\sum_s h_s$, and let $\mathrm{Cov}_N(h,h')=\frac1N\sum_s(h_s-\bar h_N)(h'_s-\bar h'_N)$. Then
\begin{equation}
\mathbb E[\Delta\bar h]=0,\qquad
\mathrm{Cov}(\Delta\bar h,\Delta\bar h')=c(k)^2\,\mathrm{Cov}_N(h,h') . \tag{B.3}\label{eq:proof-B.3}
\end{equation}
In particular $\mathrm{Var}(\Delta\bar h)=c(k)^2\,\mathrm{Var}_N(h)$.

\noindent\textbf{Proof.}
Without loss of generality take $h,h'$ centered, so that $\bar h_N=\bar h'_N=0$, since replacing $h_s$ by $h_s-\bar h_N$ leaves $\Delta\bar h$ unchanged. Then $\Delta\bar h=\bar h_{\mathcal S}=\frac1k\sum_s I_s h_s$. From $\mathbb E[I_s]=k/N$,
\begin{equation*}
\mathbb E[\Delta\bar h]=\frac1k\sum_{s=1}^N\mathbb E[I_s]\,h_s=\frac1k\cdot\frac kN\sum_{s=1}^N h_s=\bar h_N=0 .
\end{equation*}
Because the mean is zero,
\begin{equation*}
\mathrm{Cov}(\Delta\bar h,\Delta\bar h')
=\mathbb E\Big[\Big(\tfrac1k\sum_s I_s h_s\Big)\Big(\tfrac1k\sum_t I_t h'_t\Big)\Big]
=\frac1{k^2}\sum_{s,t}\mathbb E[I_s I_t]\,h_s h'_t .
\end{equation*}
Writing $\mathbb E[I_sI_t]=\mathrm{Cov}(I_s,I_t)+\mathbb E[I_s]\mathbb E[I_t]$ and using $\sum_s h_s=\sum_t h'_t=0$, the product-of-means term $\frac{k^2}{N^2}(\sum_s h_s)(\sum_t h'_t)$ vanishes and only the covariance term survives. Substituting \theoryeqref{eq:proof-B.2},
\begin{equation*}
\mathrm{Cov}(\Delta\bar h,\Delta\bar h')
=\frac1{k^2}\sum_{s,t}\frac{k(N-k)}{N^2}\Big(\delta_{st}-\frac{1-\delta_{st}}{N-1}\Big)h_s h'_t .
\end{equation*}
The diagonal contributes $\sum_s h_sh'_s$, and the off-diagonal contributes
\begin{equation*}
\sum_{s\ne t}h_s h'_t=\Big(\sum_s h_s\Big)\Big(\sum_t h'_t\Big)-\sum_s h_s h'_s=-\sum_s h_s h'_s .
\end{equation*}
The bracket therefore equals $\big(1+\tfrac1{N-1}\big)\sum_s h_sh'_s=\tfrac{N}{N-1}\sum_s h_sh'_s$, and
\begin{equation*}
\mathrm{Cov}(\Delta\bar h,\Delta\bar h')
=\frac1{k^2}\cdot\frac{k(N-k)}{N^2}\cdot\frac{N}{N-1}\sum_s h_s h'_s
=\frac{N-k}{k(N-1)}\cdot\frac1N\sum_s h_s h'_s .
\end{equation*}
Here $\frac1N\sum_s h_sh'_s=\mathrm{Cov}_N(h,h')$ for centered $h,h'$ and $\frac{N-k}{k(N-1)}=c(k)^2$, which gives \theoryeqref{eq:proof-B.3}. \hfill$\blacksquare$

\medskip
The factor $c(k)^2=\frac{N-k}{k(N-1)}$ has two parts, the usual $\frac1k$ dilution from averaging over $k$ draws and the finite-population correction $\frac{N-k}{N-1}$ from sampling without replacement, which vanishes at $k=N$ where no fluctuation remains \citep{cochran1977sampling}.

\subsection{Proof of Theorem \ref{thm:1}}\label{app:thm1}

\noindent\textbf{Theorem \ref{thm:1}.}
Under Assumption~\ref{assump:1}, when the effective sample size $\min\{k,N-k\}$ is large,
\begin{equation*}
E\ \overset{\cdot}{\sim}\ \mathcal N\big(0,c(k)^2C\big),\qquad
\mathrm{MAE}=\frac1M\sum_{m=1}^M|E_m|.
\end{equation*}
Under this Gaussian approximation, the MAE distribution is that of the average absolute value of correlated zero-mean Gaussian variables. The pointwise inequality $\mathrm{MAE}\ge|\bar E|$ and the approximation $\bar E\ \overset{\cdot}{\sim}\ \mathcal N(0,s^2)$ give, for every threshold $t>0$,
\begin{equation*}
\Pr(\mathrm{MAE}>t)\ge\Pr(|\bar E|>t)\approx2\Phi(-t/s),
\end{equation*}
where $s=c(k)\sigma\sqrt{\rho+(1-\rho)/M}$ is the standard deviation of the common shift $\bar E=M^{-1}\sum_m E_m$.

The proof has four parts, the exact first and second moments of $E$, its multivariate normal approximation and the induced MAE distribution, the identity for the common shift together with a two-sided sandwich for the $\mathrm{MAE}$, and the right-tail lower bound. The tail bound follows because $\mathrm{MAE}\ge|\bar E|$ holds pointwise and $\bar E$ is approximately $\mathcal N(0,s^2)$ under Assumption~\ref{assump:1}, so $\Pr(\mathrm{MAE}>t)\ge\Pr(|\bar E|>t)\approx2\Phi(-t/s)$.

\subsubsection{Exact first and second moments of the error vector}

\noindent\theorylabel{prop:C.1}{C.1}\textbf{Proposition C.1.}
Under uniform sampling without replacement, for any finite $N$ and $1\le k<N$,
\begin{equation}
\mathbb E[E]=0,\qquad \mathrm{Cov}(E)=c(k)^2\,C . \tag{C.1}\label{eq:proof-C.1}
\end{equation}
This is exact and requires no distributional assumption.

\noindent\textbf{Proof.}
Write $E$ as a linear combination of the indicators. From $\hat\mu_m=\frac1k\sum_{s\in\mathcal S}X_{s,m}=\frac1k\sum_{s=1}^N I_s X_{s,m}$ and $X_{s,\cdot}=z_s+\mu$,
\begin{equation*}
E=\frac1k\sum_{s=1}^N I_s X_{s,\cdot}-\mu
=\frac1k\sum_{s=1}^N I_s z_s+\frac1k\Big(\sum_{s=1}^N I_s\Big)\mu-\mu .
\end{equation*}
Since $\sum_s I_s=k$, the middle term equals $\mu$ and cancels $-\mu$, which leaves the clean linear form
\begin{equation}
E=\frac1k\sum_{s=1}^N I_s z_s . \tag{C.2}\label{eq:proof-C.2}
\end{equation}
For the first moment, $\mathbb E[I_s]=k/N$ and $\sum_s z_s=0$ from \theoryeqref{eq:proof-A.1} give
\begin{equation*}
\mathbb E[E]=\frac1k\sum_{s=1}^N\mathbb E[I_s]z_s=\frac1k\cdot\frac kN\sum_{s=1}^N z_s=0 .
\end{equation*}
For the second moment, since the mean is zero,
\begin{equation*}
\mathrm{Cov}(E)=\mathbb E[EE^\top]
=\frac1{k^2}\sum_{s,t}\mathbb E[I_sI_t]\,z_sz_t^\top
=\frac1{k^2}\sum_{s,t}\mathrm{Cov}(I_s,I_t)\,z_sz_t^\top,
\end{equation*}
where the product-of-means term again vanishes because $\sum_s z_s=0$. Splitting into $s=t$ and $s\ne t$ and using Lemma \ref{lem:B.1},
\begin{equation*}
\mathrm{Cov}(E)=\frac1{k^2}\Bigg[\frac{k(N-k)}{N^2}\sum_{s}z_sz_s^\top
-\frac{k(N-k)}{N^2(N-1)}\sum_{s\ne t}z_sz_t^\top\Bigg].
\end{equation*}
With $\sum_{s\ne t}z_sz_t^\top=(\sum_s z_s)(\sum_t z_t)^\top-\sum_s z_sz_s^\top=-\sum_s z_sz_s^\top$, the bracket becomes $\frac{k(N-k)}{N^2}\big(1+\frac1{N-1}\big)\sum_s z_sz_s^\top=\frac{k(N-k)}{N(N-1)}\sum_s z_sz_s^\top$, hence
\begin{equation*}
\mathrm{Cov}(E)=\frac{N-k}{k(N-1)}\cdot\frac1N\sum_s z_sz_s^\top=c(k)^2\,C .
\end{equation*}
\hfill$\blacksquare$

\begin{figure*}[t]
\centering
\includegraphics[width=0.99\textwidth,height=0.80\textheight,keepaspectratio]{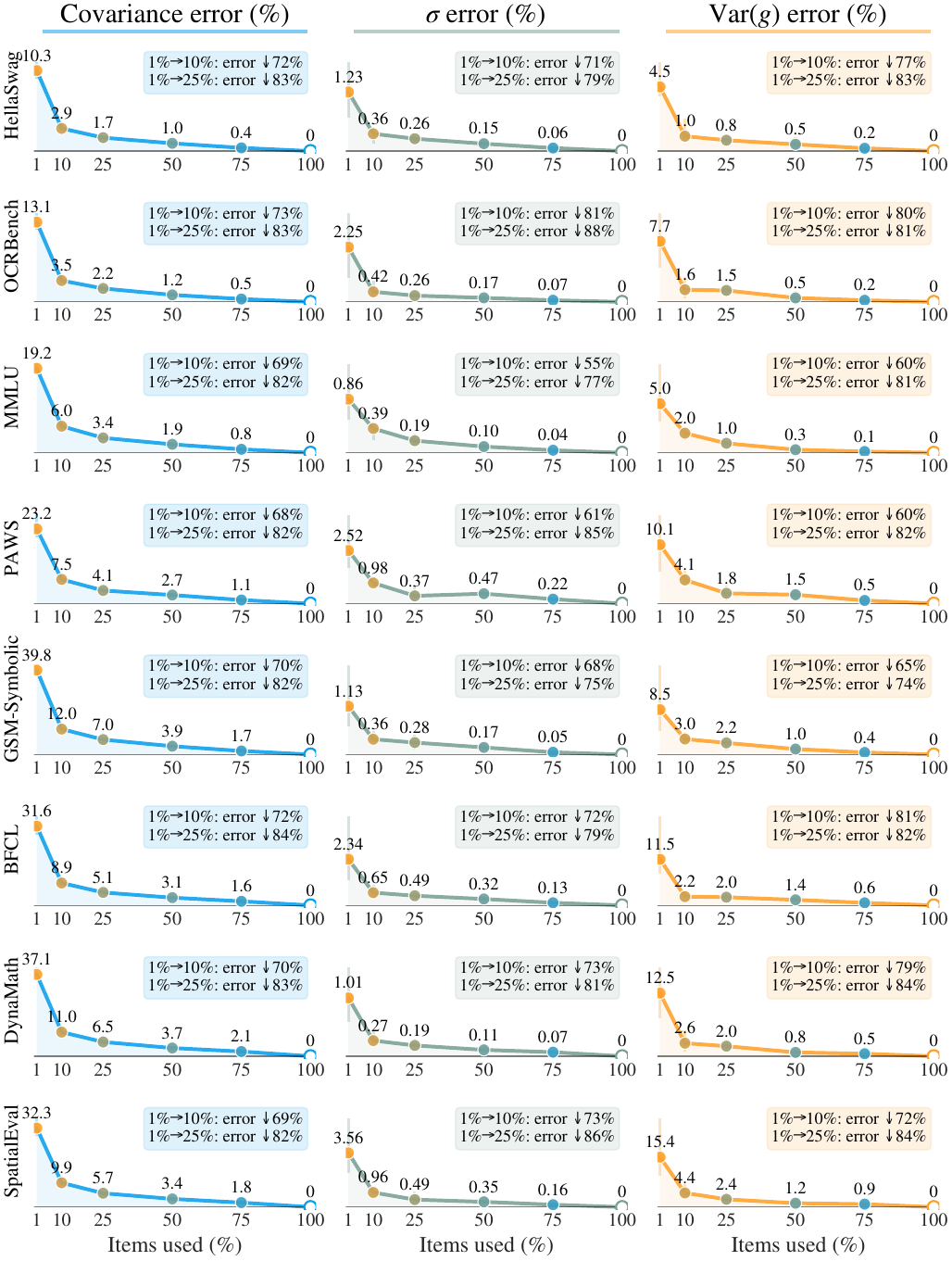}
\caption{Finite sample stability of the empirical covariance and its derived summaries. Each benchmark uses the same model set across all item subsets. The quantities $C_n$, $\sigma_n$, and $\mathrm{Var}_n(g)$ are recomputed from nested random item subsets. Each point is the median over 30 random item orderings, and the vertical bars cover the middle half of the results. The text boxes report the error reduction from 1\% to 10\% and 25\%. Across all eight benchmarks, the errors decrease sharply after only 10\% of the items and become small by 25\%, showing that partial item sets provide accurate estimates of the full benchmark covariance structure and its derived summaries.}
\label{fig:app-sigmaCn}
\end{figure*}

\clearpage

\subsubsection{Multivariate normal approximation}

Proposition~\ref{prop:C.1} gives exact moments, but a statement about near normality needs convergence, which requires embedding the fixed population into a sequence indexed by a growing $N$.

For each $N$, fix a population $X_{N,1},\dots,X_{N,N}\in[0,1]^M$ and a subset size $k_N$, with the associated quantities carrying the subscript $N$, namely $\mu_N$, $z_{N,s}=X_{N,s}-\mu_N$, $C_N=\frac1N\sum_s z_{N,s}z_{N,s}^\top$, $c_N=\sqrt{(N-k_N)/[k_N(N-1)]}$, and $E_N=\frac1{k_N}\sum_s I_s z_{N,s}$. Define the effective sample size $h_N=\min\{k_N,N-k_N\}$. We impose two conditions. First, $h_N\to\infty$, so both the retained and omitted item counts diverge. Second, $C_N\to\Sigma$ for a positive semidefinite $\Sigma$, possibly singular, which fixes the limiting covariance after scaling. These conditions specify the asymptotic regime underlying the normal approximation; a large effective sample size alone does not guarantee its accuracy for an arbitrary finite population \citep{bowyer2025position}.

The second condition above requires the score covariance to stabilize as $N$ grows. We examine its finite sample behavior by fixing all models with nonconstant scores and recomputing the covariance from nested random subsets of $n$ items. Figure~\ref{fig:app-sigmaCn} reports the relative Frobenius error of $C_n$ and the relative errors of $\sigma_n$ and $\mathrm{Var}_n(g)$, each measured against its full benchmark estimate. Here $\sigma_n^2=M^{-1}\sum_m(C_n)_{mm}$. Across all eight benchmarks, the errors fall sharply when item coverage increases from 1\% to 10\%. By 25\% of the items, all three estimates are close to their full benchmark values and improve further as more items are used. This consistent behavior provides empirical support for the covariance stability required by the asymptotic regime.

\medskip
\noindent\theorylabel{prop:C.2}{C.2}\textbf{Proposition C.2.}
Under the framework above, if $M$ is fixed and the two conditions hold, then
\begin{equation}
c_N^{-1}E_N\ \xrightarrow{\ d\ }\ W\sim\mathcal N_M(0,\Sigma). \tag{C.3}\label{eq:proof-C.3}
\end{equation}
Moreover, writing $\mathrm{MAE}_N=M^{-1}\sum_{m=1}^M|E_{N,m}|$, we have
\begin{equation*}
c_N^{-1}\mathrm{MAE}_N\ \xrightarrow{\ d\ }\ \frac1M\sum_{m=1}^M|W_m|.
\end{equation*}
Hence the working Gaussian approximation for $E$ also determines the approximate MAE distribution through the average of its absolute coordinates. Here $W$ denotes only the limiting vector of the scaled errors in \theoryeqref{eq:proof-C.3}.

\noindent\textbf{Proof.}
We use the Cram\'er--Wold device, so that convergence of every scalar projection $a^\top(c_N^{-1}E_N)$ to the matching one-dimensional normal implies convergence of the vector. Fix a direction $a\in\mathbb R^M$ and study the projection.

First, express the projection as a sample-mean deviation. Define the scalar population $y_{N,s}=a^\top z_{N,s}$, whose population mean is $\bar y_N=\frac1N\sum_s y_{N,s}=0$ by $\sum_s z_{N,s}=0$, and whose population variance normalized by $N-1$ is
\begin{equation*}
v_{N,a}=\frac1{N-1}\sum_{s=1}^N(a^\top z_{N,s})^2=\frac{N}{N-1}\,a^\top C_N a .
\end{equation*}
With $\bar y_{\mathcal S}=\frac1{k_N}\sum_{s\in\mathcal S}y_{N,s}$ and $E_N=\frac1{k_N}\sum_s I_sz_{N,s}$ from \theoryeqref{eq:proof-C.2}, the sample mean minus the population mean is the projection itself,
\begin{equation}
\bar y_{\mathcal S}-\bar y_N=a^\top E_N . \tag{C.4}\label{eq:proof-C.4}
\end{equation}

Second, treat a nondegenerate direction with $a^\top\Sigma a>0$. Because $X_{N,s}\in[0,1]^M$ gives $z_{N,s}\in[-1,1]^M$, each term is bounded by $|y_{N,s}|=|a^\top z_{N,s}|\le\|a\|_1<\infty$. The second condition gives $v_{N,a}\to a^\top\Sigma a>0$ and the first gives $h_N\to\infty$, so
\begin{equation*}
\frac{\max_s y_{N,s}^2}{h_N\,v_{N,a}}\le\frac{\|a\|_1^2}{h_N\,v_{N,a}}\longrightarrow0,
\end{equation*}
which is the Lindeberg-type condition of the H\'ajek finite-population central limit theorem \citep{hajek1960limiting, li2017general}. That theorem gives, for simple random sampling without replacement,
\begin{equation*}
\frac{\bar y_{\mathcal S}-\bar y_N}{\sqrt{(\tfrac1{k_N}-\tfrac1N)v_{N,a}}}\ \xrightarrow{\ d\ }\ \mathcal N(0,1).
\end{equation*}
The variance in the denominator simplifies through $v_{N,a}=\frac N{N-1}a^\top C_Na$,
\begin{equation*}
\Big(\frac1{k_N}-\frac1N\Big)v_{N,a}
=\frac{N-k_N}{k_NN}\cdot\frac N{N-1}\,a^\top C_Na
=c_N^2\,a^\top C_Na ,
\end{equation*}
so with \theoryeqref{eq:proof-C.4} it reads $a^\top E_N/(c_N\sqrt{a^\top C_Na})\xrightarrow{d}\mathcal N(0,1)$. Since $a^\top C_Na\to a^\top\Sigma a$, Slutsky's theorem yields
\begin{equation}
c_N^{-1}a^\top E_N\ \xrightarrow{\ d\ }\ \mathcal N(0,\,a^\top\Sigma a). \tag{C.5}\label{eq:proof-C.5}
\end{equation}

Third, treat a degenerate direction with $a^\top\Sigma a=0$, so that $a^\top C_Na\to0$. By Proposition~\ref{prop:C.1}, $\mathbb E[c_N^{-1}a^\top E_N]=0$ and
\begin{equation*}
\mathrm{Var}(c_N^{-1}a^\top E_N)=c_N^{-2}\,a^\top\mathrm{Cov}(E_N)a=a^\top C_Na\to0 .
\end{equation*}
Chebyshev's inequality then gives $c_N^{-1}a^\top E_N\xrightarrow{p}0$, which is the normal limit $\mathcal N(0,0)$ and agrees with \theoryeqref{eq:proof-C.5} at $a^\top\Sigma a=0$.

Fourth, reassemble the vector. For every $a$ we have $c_N^{-1}a^\top E_N\xrightarrow{d}\mathcal N(0,a^\top\Sigma a)$. Taking $W=\Sigma^{1/2}G$ with $G\sim\mathcal N_M(0,I_M)$ and $\Sigma^{1/2}$ the positive semidefinite square root, every projection satisfies $a^\top W\sim\mathcal N(0,a^\top\Sigma a)$, so all one-dimensional limits come from the single $W\sim\mathcal N_M(0,\Sigma)$. The Cram\'er--Wold device gives $c_N^{-1}E_N\xrightarrow{d}W\sim\mathcal N_M(0,\Sigma)$.

Finally, the map $v\mapsto M^{-1}\sum_{m=1}^M|v_m|$ is continuous on $\mathbb R^M$. Applying the continuous mapping theorem to \theoryeqref{eq:proof-C.3} yields
\begin{equation*}
\frac{\mathrm{MAE}_N}{c_N}
=\frac1M\sum_{m=1}^M\left|\frac{E_{N,m}}{c_N}\right|
\ \xrightarrow{\ d\ }\ \frac1M\sum_{m=1}^M|W_m|.
\end{equation*}
This establishes the induced MAE limit without assuming independence of the error coordinates. \hfill$\blacksquare$

\medskip
Using the finite-$N$ covariance $C$ and scale $c(k)$ gives the working formulation of Theorem~\ref{thm:1}, in the asymptotic regime specified above:
\begin{equation*}
E\ \overset{\cdot}{\sim}\ \mathcal N_M\big(0,c(k)^2C\big),\qquad
\mathrm{MAE}=\frac1M\sum_{m=1}^M|E_m|.
\end{equation*}
The MAE identity is exact; its distribution is approximated by evaluating this function under the joint Gaussian law for $E$. Each absolute coordinate then has a half-normal marginal, but the coordinates are generally dependent, so MAE is not itself asserted to be Gaussian.

\subsubsection{The common-shift identity and the two-sided sandwich}

\noindent\theorylabel{prop:C.3}{C.3}\textbf{Proposition C.3.}
The common shift satisfies the identity
\begin{equation}
\bar E=\bar g_{\mathcal S}-\bar g_N,\qquad
\bar g_{\mathcal S}=\frac1k\sum_{s\in\mathcal S}g_s,\quad
\bar g_N=\frac1N\sum_{s=1}^N g_s, \tag{C.6}\label{eq:proof-C.6}
\end{equation}
and the $\mathrm{MAE}$ is sandwiched by
\begin{equation}
|\bar E|\ \le\ \mathrm{MAE}\ \le\ |\bar E|+\mathrm{floor},\qquad
\mathrm{floor}:=\frac1M\sum_{m=1}^M|R_m|,\quad R_m:=E_m-\bar E . \tag{C.7}\label{eq:proof-C.7}
\end{equation}
Both are identities or exact inequalities and need no assumption.

\noindent\textbf{Proof.}
For \theoryeqref{eq:proof-C.6}, substitute the definitions and exchange the order of summation,
\begin{equation*}
\bar E=\frac1M\sum_{m=1}^M E_m
=\frac1k\sum_{s\in\mathcal S}\frac1M\sum_{m=1}^M X_{s,m}-\frac1N\sum_{s=1}^N\frac1M\sum_{m=1}^M X_{s,m}
=\bar g_{\mathcal S}-\bar g_N ,
\end{equation*}
since the inner average over $m$ is exactly $g_s$. The common shift is therefore the mean difficulty of the drawn subset minus that of the full set, which depends only on which items are drawn and acts identically on every model. For the lower bound, the triangle inequality gives $\mathrm{MAE}=\frac1M\sum_m|E_m|\ge|\frac1M\sum_m E_m|=|\bar E|$. For the upper bound, write $E_m=\bar E+R_m$ and use $|\bar E+R_m|\le|\bar E|+|R_m|$, then average over $m$ to obtain $\mathrm{MAE}\le|\bar E|+\mathrm{floor}$. \hfill$\blacksquare$

\medskip
The sandwich \theoryeqref{eq:proof-C.7} shows that a large common shift necessarily produces a large $\mathrm{MAE}$. The quantity $\mathrm{floor}$ is a random residual term, not a deterministic lower endpoint or a quantity known to remain fixed under selection. Thus the sandwich identifies the common shift as one source of high-error risk, but does not imply that it is the only source or that the entire $\mathrm{MAE}$ distribution is determined by $s$.

\subsubsection{The scale of the common shift and the right-tail lower bound}

\noindent\theorylabel{prop:C.4}{C.4}\textbf{Proposition C.4.}
The sampling variance of the common shift is
\begin{equation}
s^2:=\mathrm{Var}(\bar E)=c(k)^2\,\mathrm{Var}_N(g)=\frac{c(k)^2}{M^2}\,\mathbf 1^\top C\,\mathbf 1, \tag{C.8}\label{eq:proof-C.8}
\end{equation}
with $\mathbf 1=(1,\dots,1)^\top\in\mathbb R^M$. When the effective sample size is large, $\bar E\ \overset{\cdot}{\sim}\ \mathcal N(0,s^2)$ and therefore
\begin{equation}
\Pr(\mathrm{MAE}>t)\ \ge\ \Pr(|\bar E|>t)\ \approx\ 2\,\Phi\!\Big(-\frac ts\Big). \tag{C.9}\label{eq:proof-C.9}
\end{equation}

\noindent\textbf{Proof.}
By \theoryeqref{eq:proof-C.6}, $\bar E=\bar g_{\mathcal S}-\bar g_N=\Delta\bar g$ is the sample-mean deviation of the per-item difficulty $g_s$, so Lemma \ref{lem:B.2} with $h=h'=g$ gives $\mathrm{Var}(\bar E)=c(k)^2\mathrm{Var}_N(g)$. To express $\mathrm{Var}_N(g)$ through $C$, note $g_s-\bar g_N=\frac1M\sum_m(X_{s,m}-\mu_m)=\frac1M\mathbf 1^\top z_s$, hence
\begin{equation}
\mathrm{Var}_N(g)=\frac1N\sum_{s=1}^N\Big(\frac1M\mathbf 1^\top z_s\Big)^2
=\frac1{M^2}\,\mathbf 1^\top\Big(\frac1N\sum_{s=1}^N z_sz_s^\top\Big)\mathbf 1
=\frac1{M^2}\,\mathbf 1^\top C\,\mathbf 1, \tag{C.10}\label{eq:proof-C.10}
\end{equation}
which is the mean of every entry of $C$, and \theoryeqref{eq:proof-C.8} follows. For the tail, the lower bound $\mathrm{MAE}\ge|\bar E|$ of Proposition~\ref{prop:C.3} gives $\{|\bar E|>t\}\subseteq\{\mathrm{MAE}>t\}$, so $\Pr(\mathrm{MAE}>t)\ge\Pr(|\bar E|>t)$. The one-dimensional case of Proposition~\ref{prop:C.2} with $a=\frac1M\mathbf 1$ makes $\bar E$ approximately $\mathcal N(0,s^2)$, so for $Z\sim\mathcal N(0,s^2)$,
\begin{equation*}
\Pr(|\bar E|>t)\approx\Pr(|Z|>t)=2\big(1-\Phi(t/s)\big)=2\,\Phi(-t/s) .
\end{equation*}
\hfill$\blacksquare$

\medskip
Since $2\Phi(-t/s)$ increases with $s$, a larger scale raises the Gaussian approximation to the common-shift tail and hence the lower bound on high-error risk at any fixed threshold. This proves Theorem~\ref{thm:1}; it does not identify the lower bound with the actual $\mathrm{MAE}$ tail probability. Within the Gaussian model, $2\Phi(-t/s)\sim\sqrt{\tfrac2\pi}\,\tfrac st\,e^{-t^2/(2s^2)}$ as $t/s\to\infty$. This asymptotic describes the Gaussian common-shift tail, not a relative-error guarantee for extreme tails of the finite-population distribution. In particular, ``long-tailed'' here describes the observed right-skewed error distributions, not a heavy-tailed asymptotic law. The empirical comparison discussed in the main text is Figure~\ref{fig:zzzz}.

\subsection{Closed forms under Assumption \ref{assump:1}}\label{app:assump1}

This part inserts Assumption~\ref{assump:1}, the homogeneous variance and homogeneous pairwise correlation, and derives closed forms for the common-shift scale and the mean $\mathrm{MAE}$, together with the single-common-factor representation used by Theorems~\ref{thm:2} and~\ref{thm:3}.

\subsubsection{The equicorrelation structure}

Assumption~\ref{assump:1} sets $C_{mm}=\sigma^2>0$ and $C_{ml}=\rho\sigma^2$ for $m\ne l$ with $\rho\in[0,1)$, which is exactly
\begin{equation}
C=\sigma^2\big[(1-\rho)\,I+\rho\,\mathbf 1\mathbf 1^\top\big]. \tag{D.1}\label{eq:proof-D.1}
\end{equation}
The diagonal is $\sigma^2[(1-\rho)+\rho]=\sigma^2$ and each off-diagonal is $\sigma^2[\,0+\rho\,]=\rho\sigma^2$, as required.

\subsubsection{Closed form of the common-shift scale}

\noindent\theorylabel{prop:D.2}{D.2}\textbf{Proposition D.2.}
Under Assumption~\ref{assump:1},
\begin{equation}
s=c(k)\,\sigma\sqrt{\rho+\frac{1-\rho}{M}} . \tag{D.2}\label{eq:proof-D.2}
\end{equation}

\noindent\textbf{Proof.}
From \theoryeqref{eq:proof-C.8}, $s^2=\frac{c(k)^2}{M^2}\mathbf 1^\top C\mathbf 1$. With \theoryeqref{eq:proof-D.1} and $\mathbf 1^\top I\mathbf 1=M$, $\mathbf 1^\top\mathbf 1\mathbf 1^\top\mathbf 1=M^2$,
\begin{equation*}
\mathbf 1^\top C\mathbf 1=\sigma^2\big[(1-\rho)M+\rho M^2\big],
\end{equation*}
so $s^2=\frac{c(k)^2}{M^2}\sigma^2[(1-\rho)M+\rho M^2]=c(k)^2\sigma^2\big(\rho+\frac{1-\rho}{M}\big)$, and taking the square root gives \theoryeqref{eq:proof-D.2}. \hfill$\blacksquare$

For $M>1$, the squared scale increases with $\rho$ because $\partial[\rho+(1-\rho)/M]/\partial\rho=1-1/M>0$. For fixed $k<N$, $\sigma>0$, and $\rho>0$, the expression tends to $c(k)\sigma\sqrt{\rho}>0$ as $M$ grows. Thus increasing the number of similarly behaving models does not remove the correlation-dependent component. At $\rho=0$, by contrast, $s=c(k)\sigma/\sqrt M$ vanishes in this limit.

\subsubsection{The mean of the MAE}

\noindent\theorylabel{prop:D.3}{D.3}\textbf{Proposition D.3.}
Under Assumption~\ref{assump:1} and the normal approximation of Theorem~\ref{thm:1},
\begin{equation}
\mathbb E[\mathrm{MAE}]\approx\sqrt{\tfrac2\pi}\,c(k)\,\sigma . \tag{D.3}\label{eq:proof-D.3}
\end{equation}

\noindent\textbf{Proof.}
By Theorem~\ref{thm:1} the marginal of each error is approximately $E_m\sim\mathcal N(0,c(k)^2\sigma^2)$, since the diagonal of $C$ is $\sigma^2$. A zero-mean normal $Z\sim\mathcal N(0,\tau^2)$ has $\mathbb E|Z|=\sqrt{2/\pi}\,\tau$, which is the folded-normal mean of Lemma~\ref{lem:E.2} at zero mean, so with $\tau=c(k)\sigma$ each $\mathbb E|E_m|=\sqrt{2/\pi}\,c(k)\sigma$. Averaging the equal marginals over $m$ leaves the value unchanged, giving \theoryeqref{eq:proof-D.3}. \hfill$\blacksquare$

\medskip
Within the Gaussian approximation, the mean does not depend on $\rho$, since it is fixed by the single-model marginal scale $c(k)\sigma$. Dividing \theoryeqref{eq:proof-D.2} by the Gaussian-model mean in \theoryeqref{eq:proof-D.3} gives the common-shift scale relative to the mean error,
\begin{equation}
\frac{s}{\mathbb E[\mathrm{MAE}]}=\sqrt{\tfrac\pi2}\,\sqrt{\rho+\frac{1-\rho}{M}}\ \xrightarrow{\ M\to\infty\ }\ \sqrt{\tfrac\pi2}\,\sqrt\rho, \tag{D.4}\label{eq:proof-D.4}
\end{equation}
where $\sigma$ and $c(k)$ cancel and only $\rho$ and $M$ remain.

\subsubsection{The single-common-factor representation}

\noindent\theorylabel{prop:D.4}{D.4}\textbf{Proposition D.4.}
Within the Gaussian approximation of Theorem~\ref{thm:1}, Assumption~\ref{assump:1} gives the representation
\begin{equation}
E_m\ \overset{d}{=}\ c(k)\,\sigma\big(\sqrt\rho\,G+\sqrt{1-\rho}\,Z_m\big),\qquad
G,Z_1,\dots,Z_M\ \overset{\text{iid}}{\sim}\ \mathcal N(0,1). \tag{D.5}\label{eq:proof-D.5}
\end{equation}

\noindent\textbf{Proof.}
The right side is a linear combination of independent normals, so $(E_1,\dots,E_M)$ is jointly normal and it suffices to match the mean and covariance with $\mathcal N(0,c(k)^2C)$. The mean is zero, and the variance is
\begin{equation*}
\mathrm{Var}(E_m)=c(k)^2\sigma^2\big(\rho+(1-\rho)\big)=c(k)^2\sigma^2=c(k)^2C_{mm} .
\end{equation*}
For $m\ne l$, only the shared $G$ contributes, so
\begin{equation*}
\mathrm{Cov}(E_m,E_l)=c(k)^2\sigma^2\,\rho=c(k)^2C_{ml} .
\end{equation*}
Two zero-mean Gaussian vectors with equal covariance are identical in distribution. The equality in distribution in \theoryeqref{eq:proof-D.5} is exact for this Gaussian model, not for the original finite-population sampling distribution. \hfill$\blacksquare$

\medskip
The representation \theoryeqref{eq:proof-D.5} separates the randomness into a common factor $G$, which reflects whether the drawn items are collectively harder or easier and is shared by all models, and an individual noise $Z_m$, which is the sampling luck of each model and is independent across models. This is the starting point of the asymptotic analysis of Theorem~\ref{thm:2}.

\subsection{Proof of Theorem \ref{thm:2}}\label{app:thm2}

\noindent\textbf{Theorem \ref{thm:2}.}
Under Assumption~\ref{assump:1} with $0<\rho<1$ and the Gaussian approximation of Theorem~\ref{thm:1}, as $M\to\infty$ the $\mathrm{MAE}$ converges to a limiting distribution with lower endpoint $y_0=\sqrt{2/\pi}\,c(k)\sigma\sqrt{1-\rho}$. The mode $\hat y(M)$ of the sampling distribution of the $\mathrm{MAE}$ approaches $y_0$ at the universal rate $\hat y(M)/y_0-1=C_0/\sqrt M+O(1/M)$, with $C_0=\sqrt{\pi/2-1}\,\arg\max_{u>0}\int_0^\infty v^{-1/2}\varphi(u-v)\,\mathrm dv\approx0.578$, and the constant $C_0$ does not depend on $c(k),\sigma,\rho$. For brevity the proof writes $u^{*}=\arg\max_{u>0}\int_0^\infty v^{-1/2}\varphi(u-v)\,\mathrm dv\approx0.765$ for the maximizer.

Throughout this part write $c:=c(k)>0$ and work within the Gaussian representation \theoryeqref{eq:proof-D.5}, holding $c$, $\sigma>0$, and $0<\rho<1$ fixed as $M\to\infty$. This is a large-$M$ analysis of the Gaussian model, separate from the fixed-$M$, large-$N$ justification in Proposition~\ref{prop:C.2}; it is not a joint-limit theorem for arbitrary growing score matrices. The remainder constants may depend on the fixed parameters, so the expansion is not uniform as $\rho\to0$ or $\rho\to1$. For $Z\sim\mathcal N(0,1)$ and a real argument $g$, define
\begin{equation}
r_\rho(g):=\mathbb E\big|\sqrt\rho\,g+\sqrt{1-\rho}\,Z\big|,\qquad
v_\rho(g):=\mathrm{Var}\big(\big|\sqrt\rho\,g+\sqrt{1-\rho}\,Z\big|\big). \tag{E.0}\label{eq:proof-E.0}
\end{equation}
By \theoryeqref{eq:proof-D.5}, given the common factor $G=g$ the single-model absolute error is $|E_m|=c\sigma|\sqrt\rho\,g+\sqrt{1-\rho}\,Z_m|$, with conditional mean $c\sigma\,r_\rho(g)$ and conditional variance $c^2\sigma^2 v_\rho(g)$. Write $\mathrm{MAE}_M:=\frac1M\sum_{m=1}^M|E_m|$.

\subsubsection{Conditioning on the common factor}

\noindent\theorylabel{lem:E.1}{E.1}\textbf{Lemma E.1.}
Given $G=g$,
\begin{equation}
\mathbb E[\mathrm{MAE}_M\mid G=g]=c\sigma\,r_\rho(g),\qquad
\mathrm{Var}[\mathrm{MAE}_M\mid G=g]=\frac{c^2\sigma^2}{M}\,v_\rho(g). \tag{E.1}\label{eq:proof-E.1}
\end{equation}

\noindent\textbf{Proof.}
Fix $G=g$. By \theoryeqref{eq:proof-D.5} the $Z_1,\dots,Z_M$ remain independent and identically distributed given $G$, so the $|E_1|,\dots,|E_M|$ are independent and identically distributed with mean $c\sigma r_\rho(g)$ and variance $c^2\sigma^2 v_\rho(g)$. Averaging $M$ such variables keeps the mean and divides the variance by $M$. \hfill$\blacksquare$

\medskip
The randomness has two layers, an outer $G\sim\mathcal N(0,1)$ that decides how hard the draw is, and an inner average of $M$ individual noises whose variance carries the factor $1/M$ and concentrates as $M$ grows.

\subsubsection{Closed form and local expansion of $r_\rho$}

\noindent\theorylabel{lem:E.2}{E.2}\textbf{Lemma E.2.}
The function $r_\rho$ is even and
\begin{equation}
r_\rho(g)=\sqrt{1-\rho}\,\sqrt{\tfrac2\pi}\,e^{-\frac{\rho g^2}{2(1-\rho)}}
+\sqrt\rho\,g\,\mathrm{erf}\!\Big(\frac{\sqrt\rho\,g}{\sqrt{2(1-\rho)}}\Big). \tag{E.2}\label{eq:proof-E.2}
\end{equation}
In particular
\begin{equation}
r_\rho(0)=\sqrt{\tfrac2\pi}\sqrt{1-\rho},\qquad
r_\rho'(0)=0,\qquad
r_\rho''(0)=\sqrt{\tfrac2\pi}\,\frac{\rho}{\sqrt{1-\rho}}>0, \tag{E.3}\label{eq:proof-E.3}
\end{equation}
so near $g=0$,
\begin{equation}
r_\rho(g)=r_\rho(0)+\tfrac12 r_\rho''(0)\,g^2+O(g^4), \tag{E.4}\label{eq:proof-E.4}
\end{equation}
and $g=0$ is the unique strict minimum of $r_\rho$ with a parabolic bottom.

\noindent\textbf{Proof.}
Let $W=\sqrt\rho\,g+\sqrt{1-\rho}\,Z\sim\mathcal N(m,\varsigma^2)$ with $m=\sqrt\rho\,g$ and $\varsigma=\sqrt{1-\rho}$.

For the closed form, use the Gaussian identity $\mathbb E[\max(W,0)]=m\,\Phi(m/\varsigma)+\varsigma\,\varphi(m/\varsigma)$, obtained by substituting $u=(w-m)/\varsigma$ in $\int_0^\infty w\,\frac1\varsigma\varphi(\frac{w-m}\varsigma)dw$ and integrating the resulting $u\varphi$ and $\varphi$ parts. Since $|W|=2\max(W,0)-W$ and $\mathbb E W=m$,
\begin{equation*}
\mathbb E|W|=2\big[m\Phi(m/\varsigma)+\varsigma\varphi(m/\varsigma)\big]-m=m\big[2\Phi(m/\varsigma)-1\big]+2\varsigma\varphi(m/\varsigma).
\end{equation*}
Using $2\Phi(z)-1=\mathrm{erf}(z/\sqrt2)$ and $2\varsigma\varphi(m/\varsigma)=\varsigma\sqrt{2/\pi}\,e^{-m^2/2\varsigma^2}$,
\begin{equation*}
\mathbb E|W|=\varsigma\sqrt{\tfrac2\pi}\,e^{-m^2/2\varsigma^2}+m\,\mathrm{erf}\!\Big(\frac m{\varsigma\sqrt2}\Big),
\end{equation*}
and substituting $m=\sqrt\rho\,g$ and $\varsigma=\sqrt{1-\rho}$ gives \theoryeqref{eq:proof-E.2}. Evenness follows from $r_\rho(-g)=\mathbb E|\sqrt\rho\,g-\sqrt{1-\rho}\,Z|=r_\rho(g)$ by the symmetry of $Z$.

For the first derivative, $\partial_g|\sqrt\rho\,g+\sqrt{1-\rho}\,Z|=\sqrt\rho\,\mathrm{sign}(W)$ away from a null set, so
\begin{equation}
r_\rho'(g)=\sqrt\rho\,\mathbb E\,\mathrm{sign}(W)=\sqrt\rho\big(1-2\Pr(W<0)\big)
=\sqrt\rho\,\mathrm{erf}\!\Big(\frac{\sqrt\rho\,g}{\sqrt{2(1-\rho)}}\Big), \tag{E.5}\label{eq:proof-E.5}
\end{equation}
and $r_\rho'(0)=\sqrt\rho\,\mathrm{erf}(0)=0$. For the second derivative, differentiate \theoryeqref{eq:proof-E.5} using $\frac{d}{du}\mathrm{erf}(u)=\frac2{\sqrt\pi}e^{-u^2}$ and the chain rule with inner derivative $\frac{\sqrt\rho}{\sqrt{2(1-\rho)}}$,
\begin{equation*}
r_\rho''(g)=\sqrt\rho\cdot\frac2{\sqrt\pi}\,e^{-\frac{\rho g^2}{2(1-\rho)}}\cdot\frac{\sqrt\rho}{\sqrt{2(1-\rho)}}
=\frac{2\rho}{\sqrt{2\pi}\,\sqrt{1-\rho}}\,e^{-\frac{\rho g^2}{2(1-\rho)}}.
\end{equation*}
At $g=0$ this is $r_\rho''(0)=\sqrt{2/\pi}\,\rho/\sqrt{1-\rho}>0$. Substituting $r_\rho'(0)=0$ and $r_\rho''(0)>0$ into the expansion gives \theoryeqref{eq:proof-E.4}, and since $r_\rho''(g)>0$ everywhere, $g=0$ is the unique global minimum. \hfill$\blacksquare$

\subsubsection{The inner variance at the boundary}

\noindent\theorylabel{lem:E.3}{E.3}\textbf{Lemma E.3.}
\begin{equation}
v_\rho(0)=(1-\rho)\Big(1-\tfrac2\pi\Big). \tag{E.6}\label{eq:proof-E.6}
\end{equation}

\noindent\textbf{Proof.}
At $g=0$, $W=\sqrt{1-\rho}\,Z$, so $|W|=\sqrt{1-\rho}\,|Z|$ and
\begin{equation*}
v_\rho(0)=(1-\rho)\,\mathrm{Var}(|Z|)=(1-\rho)\big(\mathbb E Z^2-(\mathbb E|Z|)^2\big)=(1-\rho)\Big(1-\tfrac2\pi\Big),
\end{equation*}
using $\mathbb E Z^2=1$ and $\mathbb E|Z|=\sqrt{2/\pi}$. \hfill$\blacksquare$

\subsubsection{Conditional central limit theorem}

\noindent\theorylabel{lem:E.4}{E.4}\textbf{Lemma E.4.}
For fixed $g$, as $M\to\infty$,
\begin{equation}
\mathrm{MAE}_M\mid G=g\ \approx\ \mathcal N\!\Big(c\sigma\,r_\rho(g),\ \frac{c^2\sigma^2}{M}\,v_\rho(g)\Big), \tag{E.7}\label{eq:proof-E.7}
\end{equation}
with an $O(1/\sqrt M)$ skewness correction.

\noindent\textbf{Proof.}
By Lemma \ref{lem:E.1}, $\mathrm{MAE}_M\mid g$ is the sample average of $M$ independent and identically distributed variables $|E_m|$ with finite moments, since $|E_m|=c\sigma|\sqrt\rho g+\sqrt{1-\rho}Z_m|$ has finite moments of every order. The Lindeberg--L\'evy central limit theorem gives
\begin{equation*}
\sqrt M\Big(\frac1M\sum_{m=1}^M|E_m|-c\sigma r_\rho(g)\Big)\ \xrightarrow{\ d\ }\ \mathcal N\big(0,\,c^2\sigma^2 v_\rho(g)\big),
\end{equation*}
which is \theoryeqref{eq:proof-E.7}. The next Edgeworth term comes from the third moment of $|E_m|$ and is of order $O(1/\sqrt M)$, affecting only the $O(1/M)$ remainder of the main result. \hfill$\blacksquare$

\subsubsection{Limiting law and its lower endpoint}

\noindent\theorylabel{prop:E.5}{E.5}\textbf{Proposition E.5.}
As $M\to\infty$, the inner variance in Lemma \ref{lem:E.1} is proportional to $1/M$ and vanishes, so $\mathrm{MAE}_M$ converges to the limit random variable
\begin{equation}
Y:=c\sigma\,r_\rho(G),\qquad G\sim\mathcal N(0,1), \tag{E.8}\label{eq:proof-E.8}
\end{equation}
whose support has infimum
\begin{equation}
\inf\mathrm{supp}(Y)=\sqrt{\tfrac2\pi}\,c\,\sigma\,\sqrt{1-\rho}=:y_0, \tag{E.9}\label{eq:proof-E.9}
\end{equation}
and $Y\ge y_0$ everywhere with equality only at $G=0$.

\noindent\textbf{Proof.}
First, the conditional mean in Lemma~\ref{lem:E.1} equals $Y$, so
\begin{equation*}
\mathbb E[(\mathrm{MAE}_M-Y)^2]=\frac{c^2\sigma^2}{M}\,\mathbb E[v_\rho(G)]\le\frac{c^2\sigma^2}{M}\longrightarrow0,
\end{equation*}
where $v_\rho(g)\le\mathbb E[(\sqrt\rho\,g+\sqrt{1-\rho}\,Z)^2]=\rho g^2+1-\rho$. Thus $\mathrm{MAE}_M\to Y$ in mean square and hence in distribution. By Lemma \ref{lem:E.2}, $r_\rho''>0$ everywhere and $r_\rho'(0)=0$, so $g=0$ is the unique global minimum and $r_\rho(g)\ge r_\rho(0)=\sqrt{2/\pi}\sqrt{1-\rho}$ for all $g$, with equality only at $g=0$. Multiplying by $c\sigma>0$ gives $Y\ge y_0$. Since $G\sim\mathcal N(0,1)$ places positive probability on every neighborhood of $0$ and $r_\rho$ is continuous, $Y$ comes arbitrarily close to $y_0$, so $y_0$ is the infimum of its support. \hfill$\blacksquare$

\medskip
The value $y_0$ is the lower endpoint of the limiting distribution in Theorem~\ref{thm:2}, and it relates to the mean \theoryeqref{eq:proof-D.3} by $y_0=\sqrt{2/\pi}\,c\sigma\sqrt{1-\rho}=\mathbb E[\mathrm{MAE}]\cdot\sqrt{1-\rho}$, smaller than the mean by a factor $\sqrt{1-\rho}$.

\subsubsection{The square-root singularity of the limiting density}

\noindent\theorylabel{lem:E.6}{E.6}\textbf{Lemma E.6.}
The density of $Y=c\sigma\,r_\rho(G)$ diverges to the right of $y_0$,
\begin{equation}
f_Y(y)\ \sim\ A_0\,(y-y_0)^{-1/2}\quad(y\downarrow y_0),\qquad
A_0=\frac1{\sqrt{2\pi\kappa}},\quad \kappa=\tfrac12 c\sigma\,r_\rho''(0)>0 . \tag{E.10}\label{eq:proof-E.10}
\end{equation}

\noindent\textbf{Proof.}
The infimum $y_0$ is given by Proposition \ref{prop:E.5}, so it remains to find the shape just above it. Let $\Psi(g)=c\sigma\,r_\rho(g)$, so $Y=\Psi(G)$. By the local expansion \theoryeqref{eq:proof-E.4}, near $g=0$,
\begin{equation*}
y=\Psi(g)=y_0+\kappa g^2+O(g^4),\qquad \kappa=\tfrac12 c\sigma\,r_\rho''(0)>0,
\end{equation*}
an upward parabola with bottom at $g=0$. Every $y$ slightly above $y_0$ has two symmetric preimages $g_\pm(y)=\pm\sqrt{(y-y_0)/\kappa}\,(1+o(1))$. The change-of-variables formula sums over preimages,
\begin{equation*}
f_Y(y)=\sum_{i:\,\Psi(g_i)=y}\frac{\varphi(g_i)}{|\Psi'(g_i)|}.
\end{equation*}
Here $\Psi'(g)=c\sigma r_\rho'(g)$, and by \theoryeqref{eq:proof-E.5} near zero $r_\rho'(g)=r_\rho''(0)g+o(g)$, so $\Psi'(g)=2\kappa g+o(g)$ and $|\Psi'(g_\pm)|\sim2\sqrt{\kappa(y-y_0)}$ at both roots. The numerators tend to $\varphi(0)=1/\sqrt{2\pi}$, so summing the two equal contributions,
\begin{equation*}
f_Y(y)\sim2\cdot\frac{1/\sqrt{2\pi}}{2\sqrt{\kappa(y-y_0)}}=\frac1{\sqrt{2\pi\kappa}}\,(y-y_0)^{-1/2},
\end{equation*}
which is \theoryeqref{eq:proof-E.10}. \hfill$\blacksquare$

\medskip
The limiting density is zero below $y_0$ and diverges as $(y-y_0)^{-1/2}$ when approaching $y_0$ from above. The finite-$M$ analysis below shows how conditional fluctuations of width $O(1/\sqrt M)$ smooth this singularity and displace the mode. The endpoint $y_0$ belongs to the limiting law, not to the finite-$M$ distribution, which can assign positive probability below $y_0$.

\subsubsection{Finite $M$ as a convolution and the universal peak}

\noindent\theorylabel{lem:E.7}{E.7}\textbf{Lemma E.7.}
Let the smoothing width be
\begin{equation}
\delta_M=c\sigma\sqrt{\frac{v_\rho(0)}M}=c\sigma\sqrt{\frac{(1-\rho)(1-2/\pi)}M}. \tag{E.11}\label{eq:proof-E.11}
\end{equation}
In an $O(\delta_M)$ neighborhood of $y_0$,
\begin{equation}
f_{\mathrm{MAE}_M}(y)\ \approx\ \big(f_Y*\mathcal N(0,\delta_M^2)\big)(y)
\approx A_0\,\delta_M^{-1/2}\,h\!\Big(\frac{y-y_0}{\delta_M}\Big), \tag{E.12}\label{eq:proof-E.12}
\end{equation}
where
\begin{equation}
h(u)=\frac1{\sqrt{2\pi}}\int_0^\infty v^{-1/2}\,e^{-(u-v)^2/2}\,dv=\int_0^\infty v^{-1/2}\varphi(u-v)\,dv, \tag{E.13}\label{eq:proof-E.13}
\end{equation}
so the leading-order location of the mode of $f_{\mathrm{MAE}_M}$ is
\begin{equation}
\hat y(M)=y_0+u^{*}\,\delta_M+O(M^{-1}),\qquad u^{*}=\arg\max_{u>0} h(u). \tag{E.14}\label{eq:proof-E.14}
\end{equation}

\noindent\textbf{Proof.}
For the convolution approximation, integrate the approximate conditional density of Lemma \ref{lem:E.4} over the outer $G$,
\begin{equation*}
f_{\mathrm{MAE}_M}(y)\approx\int_{\mathbb R}\frac1{\sqrt{2\pi}\,\varsigma(g)}\exp\!\Big(-\frac{(y-c\sigma r_\rho(g))^2}{2\varsigma(g)^2}\Big)\varphi(g)\,dg,\qquad
\varsigma(g)=c\sigma\sqrt{v_\rho(g)/M}.
\end{equation*}
Near $y_0$ the dominant contribution is from $g\approx0$, since only $c\sigma r_\rho(g)\approx y$ keeps the inner Gaussian from being suppressed and $y\approx y_0$ corresponds to $g\approx0$. In this neighborhood the slowly varying $\varsigma(g)$ is frozen at $\varsigma(0)=c\sigma\sqrt{v_\rho(0)/M}=\delta_M$, and the map $g\mapsto c\sigma r_\rho(g)$ becomes the image of the limiting law. This is exactly the statement that $G$ is first mapped to $Y=c\sigma r_\rho(G)$ and then blurred by an independent $\mathcal N(0,\delta_M^2)$, so $f_{\mathrm{MAE}_M}\approx f_Y*\mathcal N(0,\delta_M^2)$.

For the scaling, insert the near-boundary form $f_Y(t)\approx A_0(t-y_0)^{-1/2}\mathbf 1_{t>y_0}$ of Lemma \ref{lem:E.6},
\begin{equation*}
f_{\mathrm{MAE}_M}(y)\approx\int_{y_0}^\infty A_0(t-y_0)^{-1/2}\,\frac1{\sqrt{2\pi}\,\delta_M}\,e^{-\frac{(y-t)^2}{2\delta_M^2}}\,dt .
\end{equation*}
Substituting $v=(t-y_0)/\delta_M$ and $u=(y-y_0)/\delta_M$, so that $t-y_0=\delta_M v$, $y-t=\delta_M(u-v)$ and $dt=\delta_M dv$,
\begin{equation*}
f_{\mathrm{MAE}_M}(y)\approx A_0\,\delta_M^{-1/2}\cdot\frac1{\sqrt{2\pi}}\int_0^\infty v^{-1/2}e^{-(u-v)^2/2}\,dv=A_0\,\delta_M^{-1/2}\,h(u),
\end{equation*}
which is \theoryeqref{eq:proof-E.12}, and the weight $v^{-1/2}$ in \theoryeqref{eq:proof-E.13} is precisely the square-root singularity of Lemma \ref{lem:E.6}.

For the leading-order peak, the dependence on $y$ enters only through $u=(y-y_0)/\delta_M$, while $A_0\delta_M^{-1/2}$ is free of $y$. Maximizing this profile therefore amounts to maximizing $h$. To establish uniqueness, differentiation under the integral and integration by parts give
\begin{equation*}
h'(u)=\int_0^\infty v^{-1/2}(v-u)\varphi(u-v)\,dv,\qquad
h''(u)+u h'(u)+\tfrac12 h(u)=0.
\end{equation*}
The first identity gives $h'(u)>0$ for $u\le0$. Since $h(u)\to0$ at both infinities, it has an interior maximum at some $u^{*}>0$. At every stationary point, the second identity gives $h''(u)=-h(u)/2<0$, so every stationary point is a strict maximum and there can be only one. Thus the leading profile peaks at $y_0+u^{*}\delta_M$. The remainder estimate in \theoryeqref{eq:proof-E.14} follows from the local expansion detailed in Theorem~\ref{thm:E.8}. \hfill$\blacksquare$

\subsubsection{Assembling the main result}

\noindent\theorylabel{thm:E.8}{E.8}\textbf{Theorem E.8.}
Suppose \theoryeqref{eq:proof-D.5} holds with fixed $c\sigma>0$ and $0<\rho<1$. Then for large $M$ the dominant mode $\hat y(M)$ of the density of $\mathrm{MAE}_M$ satisfies
\begin{equation}
\hat y(M)=y_0+u^{*}\,\delta_M+O\!\Big(\frac1M\Big),\qquad u^{*}=\arg\max_u h(u), \tag{T}\label{eq:proof-T}
\end{equation}
and equivalently the relative gap is the universal quantity
\begin{equation}
\frac{\hat y(M)}{y_0}-1=\frac{u^{*}\sqrt{\pi/2-1}}{\sqrt M}+O\!\Big(\tfrac1M\Big)=\frac{C_0}{\sqrt M}+O\!\Big(\tfrac1M\Big),\qquad
C_0=\sqrt{\tfrac\pi2-1}\,u^{*}. \tag{T$'$}\label{eq:proof-T-prime}
\end{equation}

\noindent\textbf{Proof.}
Expanding $\delta_M$ in the mode formula \theoryeqref{eq:proof-E.14} with $v_\rho(0)=(1-\rho)(1-2/\pi)$ from Lemma \ref{lem:E.3},
\begin{equation*}
\hat y(M)=y_0+u^{*}\,c\sigma\sqrt{\frac{(1-\rho)(1-2/\pi)}M}+O\!\Big(\frac1M\Big),
\end{equation*}
which gives the leading term in \theoryeqref{eq:proof-T}. For the remainder, set $y=y_0+\delta_M u$ and $g=\sqrt{\delta_M/\kappa}\,w$. For bounded $u,w$, the $O(g^4)$ term in $c\sigma r_\rho(g)$ contributes $O(M^{-1/2})$ to the standardized conditional mean. Evenness gives $v_\rho(g)=v_\rho(0)+O(g^2)$ and $\varphi(g)=\varphi(0)+O(g^2)$, also producing relative $O(M^{-1/2})$ corrections. The conditional Edgeworth correction has the same order. The folded-normal summands have Gaussian tails and smooth parameter dependence near $g=0$; their local density expansion and its first two derivatives can therefore be integrated over the boundary region. This gives, uniformly for $u$ in a fixed neighborhood of $u^{*}$,
\begin{equation*}
\frac{\delta_M^{1/2}}{A_0}f_{\mathrm{MAE}_M}(y_0+\delta_M u)
=h(u)+O(M^{-1/2}),
\end{equation*}
with the same error order after one or two derivatives with respect to $u$. Since $h''(u^{*})=-h(u^{*})/2<0$, the stationary point is displaced by $O(M^{-1/2})$ in $u$, hence by $O(\delta_M M^{-1/2})=O(M^{-1})$ in $y$. The boundary peak has height of order $\delta_M^{-1/2}$, whereas the limiting density is bounded away from $y_0$; the decaying tails of $h$ localize the dominant peak to this boundary region. These observations justify the mode expansion without asserting that a central limit theorem alone controls density maxima.

For the relative gap, divide by $y_0=\sqrt{2/\pi}\,c\sigma\sqrt{1-\rho}$,
\begin{equation*}
\frac{\hat y(M)-y_0}{y_0}
=\frac{u^{*}\,c\sigma\sqrt{(1-\rho)(1-2/\pi)/M}}{\sqrt{2/\pi}\,c\sigma\sqrt{1-\rho}}
=\frac{u^{*}}{\sqrt M}\sqrt{\frac{1-2/\pi}{2/\pi}} ,
\end{equation*}
and since $\frac{1-2/\pi}{2/\pi}=\frac{\pi(1-2/\pi)}2=\frac{\pi-2}2=\frac\pi2-1$, this is \theoryeqref{eq:proof-T-prime}. The factors $c,\sigma,\rho$ all cancel, so the relative gap is universal. \hfill$\blacksquare$

\medskip
Collecting the definitions, the universal constant of Theorem~\ref{thm:2} is
\begin{equation}
C_0=\sqrt{\tfrac\pi2-1}\,u^{*},\qquad u^{*}=\arg\max_{u>0}\int_0^\infty v^{-1/2}\varphi(u-v)\,dv . \tag{E.15}\label{eq:proof-E.15}
\end{equation}

\begin{figure}[ht]\centering
\includegraphics[width=0.98\columnwidth]{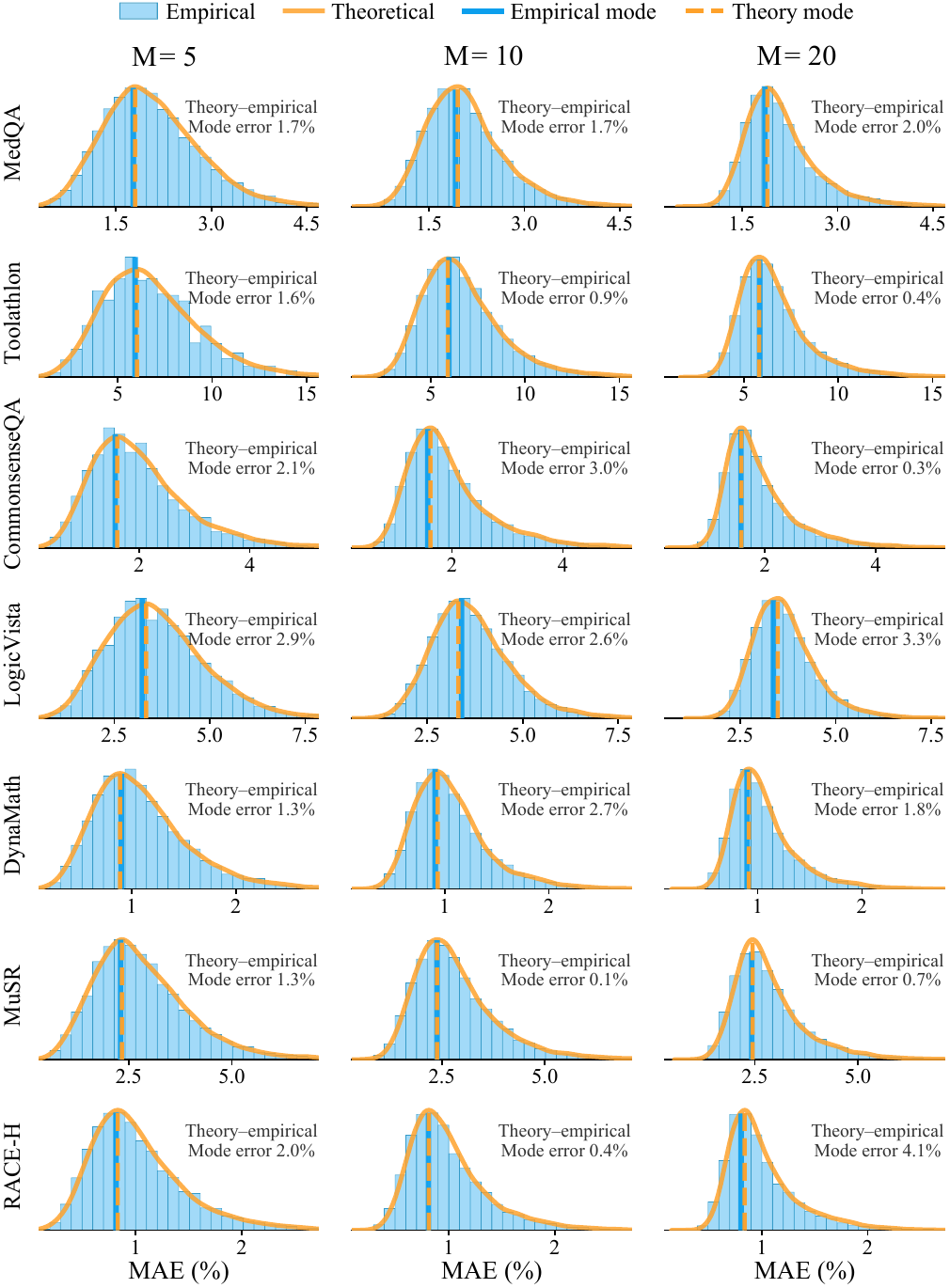}
\vspace{-5pt}
\caption{Empirical MAE distributions closely match the theoretical predictions of Theorem~\ref{thm:2} across representative benchmarks. Their shapes and empirical modes agree with theory even when only a small number of models is available. Blue histograms show empirical distributions, orange curves show theoretical distributions, and the solid blue and dashed orange lines indicate the empirical and theoretical modes. The three panels in each row use the same horizontal scale.}
\label{fig:app-mode-validation}\end{figure}

\medskip
The leading-order relative gap $0.578/\sqrt M$ is universal, giving approximately $5.8\%$ at $M=100$, $1.8\%$ at $M=1000$, and $0.9\%$ at $M=4000$. These estimates describe how the mode approaches the limiting lower endpoint $y_0$. Figure~\ref{fig:app-mode-validation} provides validation on real benchmarks. The empirical MAE distributions closely match the theoretical predictions in both shape and mode even with only five models, and this agreement persists as $M$ increases. Together with Figure~\ref{fig:zzzz}, these results show that the behavior predicted by Theorem~\ref{thm:2} is already visible with a small set of models.

A non-vanishing mean--mode gap is worth noting. Within the Gaussian model, \theoryeqref{eq:proof-D.3} gives $\mathbb E[\mathrm{MAE}_M]=\sqrt{2/\pi}\,c\sigma$, independent of $M$ and $\rho$, whereas the mode tends to $y_0=\sqrt{2/\pi}\,c\sigma\sqrt{1-\rho}$. Thus $(\mathbb E[\mathrm{MAE}_M]-\hat y(M))/\mathbb E[\mathrm{MAE}_M]\to1-\sqrt{1-\rho}>0$. This separation is consistent with the low-error mode and persistent high-error draws discussed in the main text. A mode near the lower endpoint does not, by itself, specify the probability of meeting a given error tolerance; that probability must be obtained from the distribution.

\subsection{Proof of Theorem \ref{thm:3}}\label{app:thm3}

\noindent\textbf{Theorem \ref{thm:3}.}
Under Assumptions~\ref{assump:1} and~\ref{assump:2}, least-squares orthogonality decomposes the common shift into $\bar E=\bar E_\parallel+\bar E_\perp$ with $\bar E_\parallel=\beta^\top(\bar x_{\mathcal S}-\bar x_N)$ and $\bar E_\perp=\bar e_{\mathcal S}$, uncorrelated with variances $R^2s^2$ and $(1-R^2)s^2$. For the full-benchmark-reference rule $S^{*}=\arg\min_{j\le m}\|\bar x_{S_j}-\bar x_N\|$, write $D^{*}=\|\bar x_{S^{*}}-\bar x_N\|$. Within the joint Gaussian approximation specified below, the selected scale satisfies $s_{S^{*}}^2=\mathrm{Var}(\bar E_\parallel(S^{*}))+(1-R^2)s^2$, with $0\le\mathrm{Var}(\bar E_\parallel(S^{*}))\le\|\beta\|^2\mathbb E[D^{*2}]$. Consequently, if $\mathbb E[D^{*2}]\to0$, the scale tends to $\sqrt{1-R^2}\,s<s$.

Throughout write $c:=c(k)$. The orthogonal decomposition and the pre-selection variance identities are exact under uniform sampling without replacement and do not require Assumption~\ref{assump:1}. The post-selection residual-variance identity additionally uses a joint Gaussian approximation for the probe and residual imbalances, together with independent candidate draws from the full benchmark. Least-squares orthogonality alone gives zero covariance, not independence. This distinction specifies the scope of the selection formula in Theorem~\ref{thm:3}; the pool-based implementation is discussed below.

\clearpage

\subsubsection{Normal equations and in-sample orthogonality}

Assumption~\ref{assump:2} regresses difficulty on the $p$ probe scores by least squares, $g_s=\alpha+\beta^\top x_s+e_s$, where $\alpha,\beta$ minimize $\sum_s(g_s-\alpha-\beta^\top x_s)^2$ and $e_s$ is the residual.

\medskip
\noindent\theorylabel{lem:F.1}{F.1}\textbf{Lemma F.1.}
The least-squares solution satisfies the normal equations
\begin{equation}
\bar e_N=\frac1N\sum_{s=1}^N e_s=0,\qquad
\mathrm{Cov}_N(x,e)=\frac1N\sum_{s=1}^N x_s e_s-\bar x_N\bar e_N=0 . \tag{F.2}\label{eq:proof-F.2}
\end{equation}

\noindent\textbf{Proof.}
At the optimum the partial derivatives of the residual sum of squares vanish. Differentiating with respect to $\alpha$,
\begin{equation*}
\frac{\partial}{\partial\alpha}\sum_s(g_s-\alpha-\beta^\top x_s)^2=-2\sum_s e_s=0,
\end{equation*}
so $\sum_s e_s=0$ and $\bar e_N=0$. Differentiating with respect to the $i$-th component of $\beta$,
\begin{equation*}
\frac{\partial}{\partial\beta_i}\sum_s(g_s-\alpha-\beta^\top x_s)^2=-2\sum_s x_{s,i}e_s=0,
\end{equation*}
so $\sum_s x_{s,i}e_s=0$ for every $i$, that is $\sum_s x_se_s=0$. With $\bar e_N=0$ this gives $\mathrm{Cov}_N(x,e)=\frac1N\sum_s x_se_s-\bar x_N\bar e_N=0$. \hfill$\blacksquare$

\medskip
The orthogonality \theoryeqref{eq:proof-F.2} is an algebraic identity of least squares and holds for any data.

\subsubsection{The Pythagorean split of difficulty variance and $R^2$}

\noindent\theorylabel{lem:F.2}{F.2}\textbf{Lemma F.2.}
The empirical difficulty variance splits orthogonally,
\begin{equation}
\mathrm{Var}_N(g)=\mathrm{Var}_N(\beta^\top x)+\mathrm{Var}_N(e), \tag{F.3}\label{eq:proof-F.3}
\end{equation}
so $R^2:=\mathrm{Var}_N(\beta^\top x)/\mathrm{Var}_N(g)\in[0,1]$ and $\mathrm{Var}_N(e)=(1-R^2)\mathrm{Var}_N(g)$. Under Assumption~\ref{assump:2}, $R^2\in(0,1]$.

\noindent\textbf{Proof.}
From $g_s=\alpha+\beta^\top x_s+e_s$, and since a constant contributes no variance,
\begin{equation*}
\mathrm{Var}_N(g)=\mathrm{Var}_N(\beta^\top x)+2\,\mathrm{Cov}_N(\beta^\top x,e)+\mathrm{Var}_N(e).
\end{equation*}
The cross term is $\mathrm{Cov}_N(\beta^\top x,e)=\beta^\top\mathrm{Cov}_N(x,e)=0$ by Lemma \ref{lem:F.1}, so only two terms remain, giving \theoryeqref{eq:proof-F.3}. Dividing by $\mathrm{Var}_N(g)>0$ gives $R^2+\mathrm{Var}_N(e)/\mathrm{Var}_N(g)=1$, hence $\mathrm{Var}_N(e)=(1-R^2)\mathrm{Var}_N(g)$ and $R^2\in[0,1]$. The case $R^2=1$ is allowed: it occurs when the probes fit difficulty exactly, leaving zero residual variance. Assumption~\ref{assump:2} excludes only $R^2=0$. \hfill$\blacksquare$

\medskip
Here $R^2$ measures how strongly items that are hard overall are also hard for the probes, and a larger value means the probes see more of the difficulty shift.

\subsubsection{The visible and invisible parts of the common shift}

\noindent\theorylabel{prop:F.3}{F.3}\textbf{Proposition F.3.}
The common shift decomposes as
\begin{equation}
\bar E=\bar E_\parallel+\bar E_\perp,\qquad
\bar E_\parallel=\beta^\top(\bar x_{\mathcal S}-\bar x_N),\quad
\bar E_\perp=\bar e_{\mathcal S}, \tag{F.4}\label{eq:proof-F.4}
\end{equation}
with uncorrelated parts of variance
\begin{equation}
\mathrm{Var}(\bar E_\parallel)=R^2 s^2,\qquad
\mathrm{Var}(\bar E_\perp)=(1-R^2)s^2,\qquad
\mathrm{Cov}(\bar E_\parallel,\bar E_\perp)=0 . \tag{F.5}\label{eq:proof-F.5}
\end{equation}

\noindent\textbf{Proof.}
By Proposition \ref{prop:C.3}, $\bar E=\bar g_{\mathcal S}-\bar g_N$. Substituting $g_s=\alpha+\beta^\top x_s+e_s$ and cancelling the constant $\alpha$ between the two averages,
\begin{equation*}
\bar E=\big(\alpha+\beta^\top\bar x_{\mathcal S}+\bar e_{\mathcal S}\big)-\big(\alpha+\beta^\top\bar x_N+\bar e_N\big)
=\beta^\top(\bar x_{\mathcal S}-\bar x_N)+(\bar e_{\mathcal S}-\bar e_N),
\end{equation*}
and $\bar e_N=0$ from Lemma \ref{lem:F.1} gives $\bar e_{\mathcal S}-\bar e_N=\bar e_{\mathcal S}$, which is \theoryeqref{eq:proof-F.4}. For the variances, $\bar E_\parallel=\beta^\top(\bar x_{\mathcal S}-\bar x_N)$ is the sample-mean deviation of the per-item quantity $\beta^\top x_s$, so Lemma \ref{lem:B.2} with $h=h'=\beta^\top x$ gives
\begin{equation*}
\mathrm{Var}(\bar E_\parallel)=c^2\,\mathrm{Var}_N(\beta^\top x)=c^2\,R^2\,\mathrm{Var}_N(g)=R^2 s^2,
\end{equation*}
using Lemma \ref{lem:F.2} and $s^2=c^2\mathrm{Var}_N(g)$ from \theoryeqref{eq:proof-C.8}. Likewise $\bar E_\perp=\bar e_{\mathcal S}=\Delta\bar e$ since $\bar e_N=0$, so
\begin{equation*}
\mathrm{Var}(\bar E_\perp)=c^2\,\mathrm{Var}_N(e)=c^2(1-R^2)\mathrm{Var}_N(g)=(1-R^2)s^2 .
\end{equation*}
The cross term uses Lemma \ref{lem:B.2} with $h=\beta^\top x$, $h'=e$ and the normal equation $\mathrm{Cov}_N(x,e)=0$,
\begin{equation*}
\mathrm{Cov}(\bar E_\parallel,\bar E_\perp)=c^2\,\mathrm{Cov}_N(\beta^\top x,e)=c^2\,\beta^\top\mathrm{Cov}_N(x,e)=0 .
\end{equation*}
\hfill$\blacksquare$

\medskip
Before selection, the common-shift variance splits into the probe-explained block $R^2s^2$ and the residual block $(1-R^2)s^2$. Probe matching directly controls the former. Whether selection leaves the latter unchanged requires the additional distributional argument below.

\subsubsection{Selecting the probe-matched subset contracts the visible block}

Building on the decomposition above, we analyze an idealized full-reference selection rule for benchmark compression and show that probe matching can reduce the probe-explained component of common-shift variance. Covariate balance has been studied in other settings \citep{morgan2012rerandomization,li2018asymptotic,wang2025asymptotic,deville2004efficient}.
This rule draws $S_1,\dots,S_m$ independently, each uniformly among the $k$-subsets of the full benchmark, and selects $S^{*}=\arg\min_{j\le m}\|\bar x_{S_j}-\bar x_N\|$ with $D^{*}=\|\bar x_{S^{*}}-\bar x_N\|$. Sampling is without replacement within each candidate, but candidates may overlap or repeat. Ties are broken by a fixed index rule that does not inspect residuals. Write $\bar E_\parallel(S^{*})=\beta^\top(\bar x_{S^{*}}-\bar x_N)$ and $\bar E_\perp(S^{*})=\bar e_{S^{*}}$.

\medskip
\noindent\theorylabel{lem:F.4}{F.4}\textbf{Lemma F.4.}
Under Assumptions~\ref{assump:1} and~\ref{assump:2}, and within the joint Gaussian approximation for independent candidate probe and residual imbalances, the common-shift scale on the selected subset is
\begin{equation}
s_{S^{*}}^2=\mathrm{Var}\big(\bar E_\parallel(S^{*})\big)+(1-R^2)s^2,\qquad
0\le\mathrm{Var}\big(\bar E_\parallel(S^{*})\big)\le\|\beta\|^2\,\mathbb E\big[D^{*2}\big], \tag{F.7}\label{eq:proof-F.7}
\end{equation}
If $\mathbb E[D^{*2}]\to0$, then $s_{S^{*}}\to s\sqrt{1-R^2}<s$. The variance bound on the probe-explained part is exact even without the Gaussian approximation; invariance of the residual variance is a Gaussian-model conclusion.

\noindent\textbf{Proof.}
For a candidate subset $S$, write $\Delta\bar x:=\bar x_S-\bar x_N\in\mathbb R^p$ and $\Delta\bar e:=\bar e_S-\bar e_N=\bar e_S$ using $\bar e_N=0$, so the two parts are $\bar E_\parallel=\beta^\top\Delta\bar x$ and $\bar E_\perp=\Delta\bar e$. The proof is in three steps.

First, the visible block is squeezed by the probe imbalance, and the selection presses on it directly. By the Cauchy--Schwarz inequality, $|\bar E_\parallel|=|\beta^\top\Delta\bar x|\le\|\beta\|\,\|\Delta\bar x\|=\|\beta\|\,D$ with $D=\|\Delta\bar x\|$. The selection rule takes $\min_j\|\Delta\bar x_j\|$, driving $D$ to $D^{*}$, so the selected visible block obeys $|\bar E_\parallel(S^{*})|\le\|\beta\|D^{*}$. Squaring and taking expectation,
\begin{equation*}
\mathrm{Var}\big(\bar E_\parallel(S^{*})\big)\le\mathbb E\big[\bar E_\parallel(S^{*})^2\big]\le\|\beta\|^2\,\mathbb E\big[D^{*2}\big]\ \xrightarrow{\ D^{*}\to0\ }\ 0 .
\end{equation*}
The bound holds for any $m$ and rests only on Cauchy--Schwarz. Increasing the candidate count makes $D^{*}$ and this upper bound non-increasing under a coupling that retains previous candidates. For a fixed finite benchmark, however, the attainable minimum is $\min_{|S|=k}\|\bar x_S-\bar x_N\|$, which need not equal zero. Thus the limit $D^{*}\to0$ is conditional, not guaranteed merely by taking $m\to\infty$. The sufficient condition used here is $\mathbb E[D^{*2}]\to0$; for bounded probe scores, convergence in probability to zero also implies this condition.

Second, the invisible block is untouched by the selection. The rule reads only the probe imbalances $\{\Delta\bar x_j\}$ and never the residual imbalances $\{\Delta\bar e_j\}$. By Lemma \ref{lem:B.2} and the normal equation $\mathrm{Cov}_N(x,e)=0$,
\begin{equation*}
\mathrm{Cov}(\Delta\bar x,\Delta\bar e)=c^2\,\mathrm{Cov}_N(x,e)=\mathbf 0 .
\end{equation*}
Apply the finite-population central limit argument \citep{li2017general} to the augmented per-item vector $(x_s,e_s)$, with fixed probe dimension and the corresponding boundedness and covariance-limit conditions. This motivates a joint Gaussian approximation for $(\Delta\bar x,\Delta\bar e)$. Within that model, zero cross-covariance implies independence. Independence across candidate draws then makes the entire residual vector independent of the entire collection of probe imbalances. Conditional on all probe imbalances, the selected index is fixed and its residual retains the same zero-mean Gaussian marginal as every candidate. Therefore, within this approximation,
\begin{equation*}
\mathrm{Var}\big(\bar E_\perp(S^{*})\big)=\mathrm{Var}(\Delta\bar e)=c^2\,\mathrm{Var}_N(e)=(1-R^2)s^2 .
\end{equation*}

Third, combine the two blocks. Since $\bar E_\parallel(S^{*})$ is a function of $\{\Delta\bar x_j\}$ while $\bar E_\perp(S^{*})$ is independent of them with zero mean,
\begin{equation*}
\mathbb E\big[\bar E_\parallel(S^{*})\,\bar E_\perp(S^{*})\big]=\mathbb E\big[\bar E_\parallel(S^{*})\big]\cdot0=0,
\end{equation*}
so the two variances add,
\begin{equation*}
s_{S^{*}}^2=\mathrm{Var}\big(\bar E_\parallel(S^{*})\big)+(1-R^2)s^2 .
\end{equation*}
The first term lies in $[0,\|\beta\|^2\mathbb E[D^{*2}]]$ and vanishes if $\mathbb E[D^{*2}]\to0$, giving $s_{S^{*}}\to s\sqrt{1-R^2}$. For the original finite-population distribution, zero covariance alone does not ensure that selection preserves the residual variance or the cross-covariance, so the displayed post-selection equality is approximate rather than an exact sampling identity. Nor does a fixed-candidate Gaussian approximation by itself justify arbitrarily increasing $m$. \hfill$\blacksquare$

\medskip
The bound on the probe-explained variance shrinks with $\mathbb E[D^{*2}]$, while the residual contribution stays unchanged within the Gaussian selection model. Here $\beta$ is the coefficient vector for the panel-average difficulty $g_s$, not a coefficient separately defined for each target model. Since $|\bar E_\parallel|\le\|\beta\|D$, reducing $D$ provides a selection criterion that does not require knowing $\beta$. This motivates BestRand without asserting that every finite candidate count achieves the limiting contraction factor.




\medskip
In the centered Gaussian selection model, simultaneous sign reversal of all candidate probe imbalances leaves the selected index unchanged and reverses the selected probe vector. Its mean is therefore zero. Together with the independent zero-mean residual, this gives $\mathbb E[\bar E(S^{*})]=0$ within that model. This is not an exact unbiasedness guarantee for every target model under finite-population or pool-based selection.

\begin{figure}[t]
    \centering
    \includegraphics[width=\columnwidth]
    {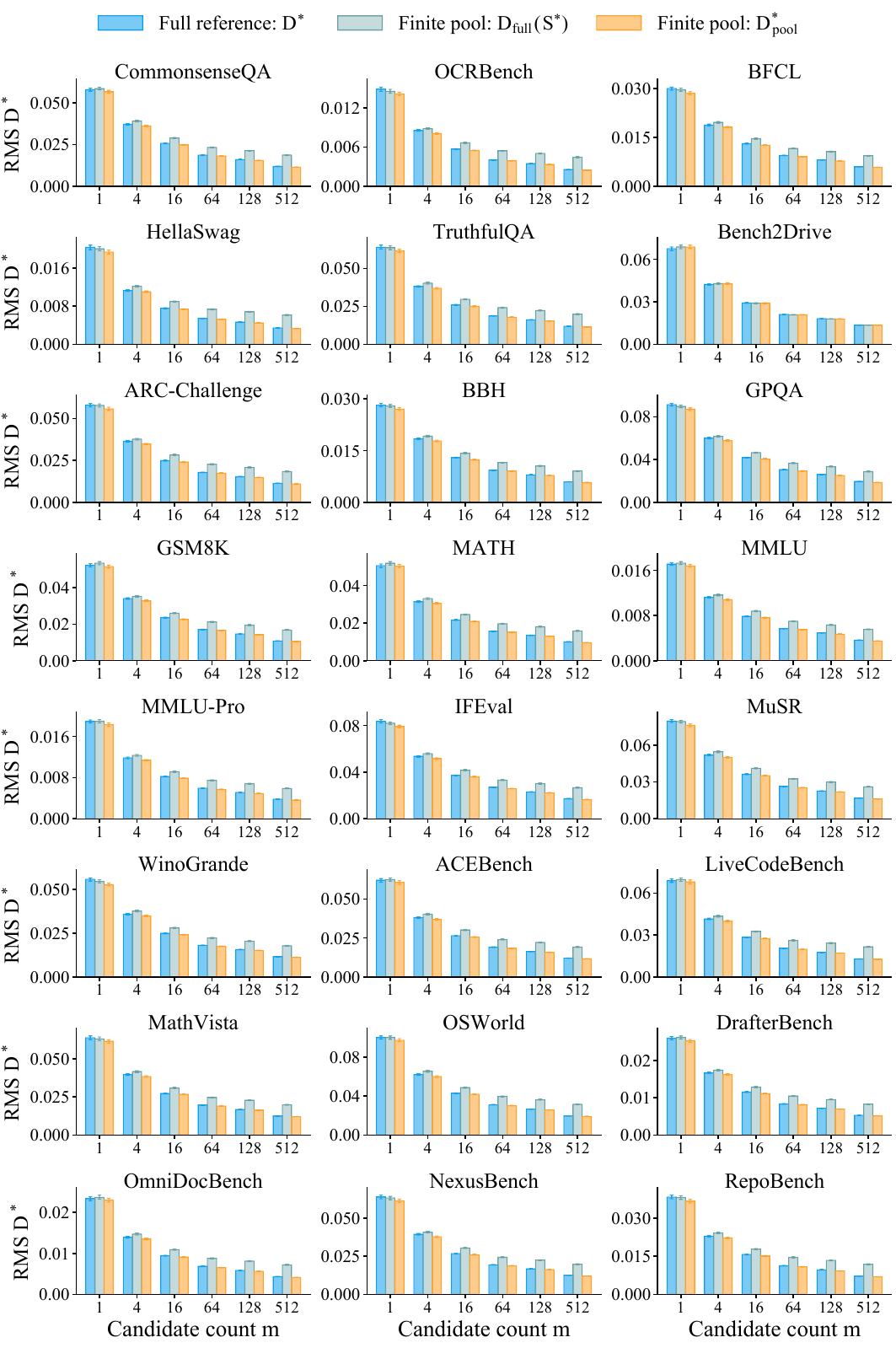}
    \caption{RMS probe imbalance as the candidate count increases. Full-reference selection reports $\sqrt{\mathbb{E}[D^{*2}]}$ for the rule in Theorem~\ref{thm:3}. For finite-pool BestRand, $D_{\mathrm{pool}}^*$ is the optimized imbalance and $D_{\mathrm{full}}(S^*)$ is the imbalance relative to the full benchmark probe mean.}
    \label{fig:bestrand-probe-imbalance}
\end{figure}

\begin{figure}[t]
    \centering
    \includegraphics[width=\columnwidth]
    {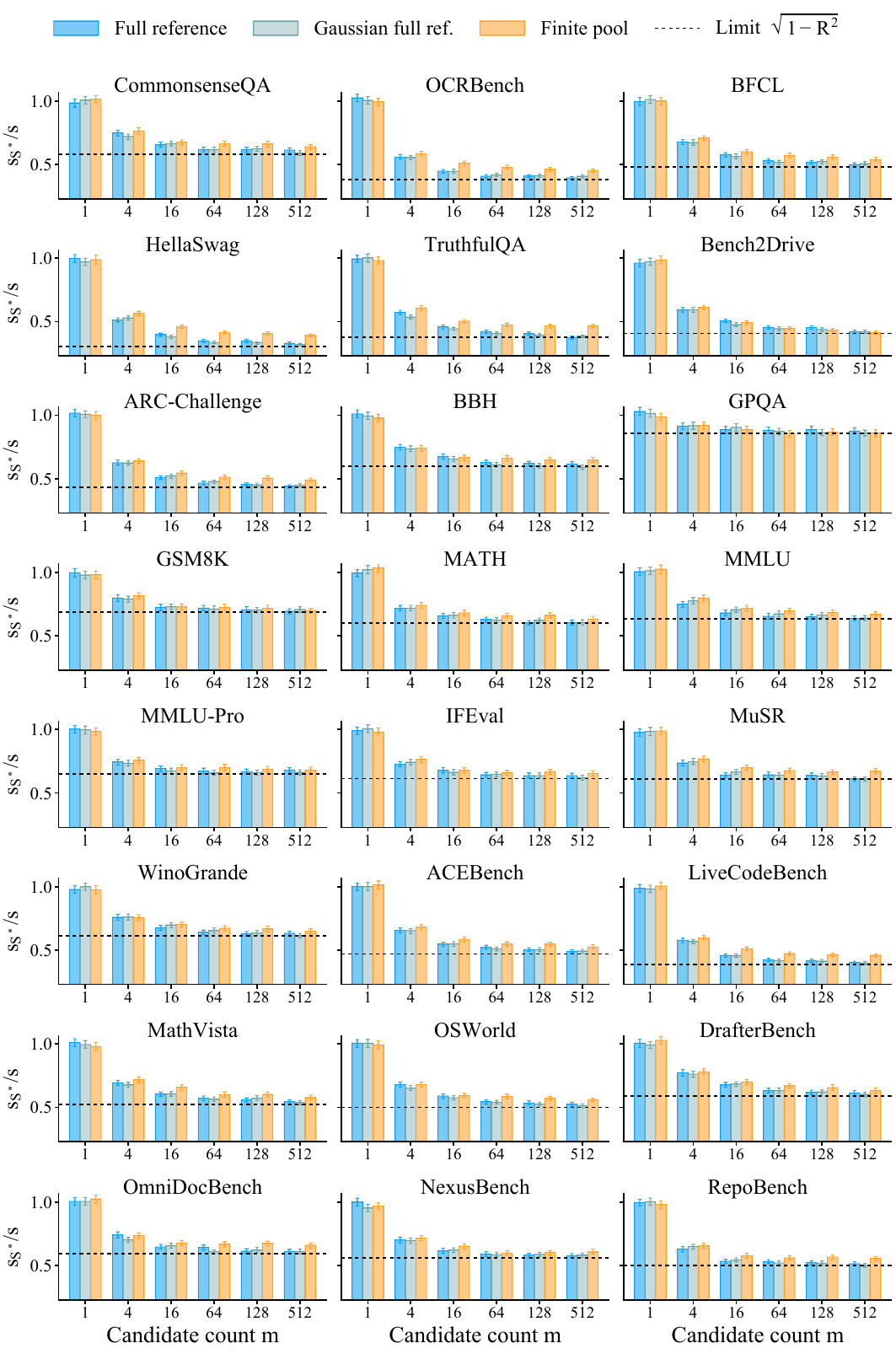}
    \caption{Common-shift scale ratio $s_{S^*}/s$ as the candidate count increases. Curves compare the joint Gaussian full-reference model, empirical full-reference selection, and finite-pool BestRand. The horizontal line marks the Gaussian full-reference limit $\sqrt{1-R^2}$ as $\mathbb{E}[D^{*2}]\to0$.}
    \label{fig:bestrand-longtail}
\end{figure}

Figures~\ref{fig:bestrand-probe-imbalance} and~\ref{fig:bestrand-longtail} compare the predictions of Lemma~\ref{lem:F.4} with experimental results. As $m$ increases, $\sqrt{\mathbb E[D^{*2}]}$ decreases under full-reference selection, while finite-pool BestRand reduces both its pool-reference objective and its full-reference imbalance. The common-shift scale ratio $s_{S^*}/s$ decreases accordingly, and the Gaussian and empirical full-reference results approach the predicted limit $\sqrt{1-R^2}$. Finite-pool results follow the same trend, with small deviations because $H$ only approximates $\bar x_N$. Overall, the experimental results closely match the predictions of Theorem~\ref{thm:3}.

\subsubsection{Implications for high-error risk}

For every selection rule, the pointwise inequality from Proposition~\ref{prop:C.3} remains valid:
\begin{equation}
\Pr\big(\mathrm{MAE}(S^{*})>t\big)\ge\Pr\big(|\bar E(S^{*})|>t\big).
\end{equation}
Within the independent-candidate Gaussian model of Lemma~\ref{lem:F.4}, if $\mathbb E[D^{*2}]\to0$ and $0<R^2<1$, the probe-explained part vanishes in mean square while the residual remains $\mathcal N(0,(1-R^2)s^2)$. Thus, for fixed $t>0$,
\begin{equation*}
\Pr\big(|\bar E(S^{*})|>t\big)\longrightarrow2\Phi\!\Big(-\frac{t}{s\sqrt{1-R^2}}\Big).
\end{equation*}
For $R^2=1$, the limiting common shift is zero and this probability tends to zero. At finite probe imbalance, the selected common shift need not be Gaussian, so replacing its law by a normal distribution solely on the basis of its variance is not justified.

The contraction reduces the common-shift contribution to high-error risk. A smaller lower bound alone does not prove that the actual $\mathrm{MAE}$ tail probability decreases, nor that its variance or quantiles contract by the same factor, or that its mode stays fixed. The empirical reduction in high-error outcomes is reported in Figure~\ref{fig:long-tail-distribution-main}; the theorem provides the common-shift mechanism motivating the search.

\clearpage

\section{Related Work}
\label{app:related-work}

\subsection{LLM, VLM, and VLA Benchmarks}
\label{app:llm-vlm-vla-benchmarks}

Benchmarks for LLMs, VLMs, and VLAs evaluate a broad range of capabilities across text, multimodal, and interactive tasks \citep{hendrycks2020measuring,chen2021evaluating,lu2024mathvista,dauner2024navsim, ling2026blue, zhong2024let, liu2025associam}.
Examples include LongBench v2, MMMU, and Bench2Drive \citep{bai2025longbench,yue2024mmmu,jia2024bench2drive}.
Evaluating modern foundation models can be expensive even for individual items, particularly when models process long contexts, generate extended reasoning chains, or perform repeated inference during interaction with an environment \citep{huang2025minilongbench,bai2025longbench,jia2024bench2drive}.
This cost accumulates across benchmarks containing hundreds to tens of thousands of evaluation items \citep{polo2024tinybenchmarks,perlitz2024efficient}.
Many benchmarks also contain redundancy, as different items can provide overlapping information about model capabilities \citep{zhang2025redundancy,vivek2024anchor}.
Prior studies show that small subsets can closely approximate full-benchmark scores or preserve model rankings, suggesting that much of the information needed for aggregate evaluation can be retained with fewer items \citep{vivek2024anchor,polo2024tinybenchmarks,huang2025minilongbench,saranathan2025sublime, zhao2024flasheval, wang2026rethinking}.
BestRand exploits this redundancy to reduce evaluation cost across these settings while improving the accuracy and stability of evaluation on compact subsets.

\subsection{Low-Cost Deep Learning}
\label{app:low-cost-deep-learning}

Improving efficiency and reducing training and inference costs have long been central goals in deep learning, with methods operating on both models and data \citep{wan2023efficient}.
On the model side, pruning and quantization improve inference efficiency \citep{han2015deep,frantar2023sparsegpt,xiao2023smoothquant}, while parameter-efficient fine-tuning reduces the resources required to adapt pretrained models \citep{houlsby2019parameter, hu2021lora}.
On the data side, coreset selection and data pruning retain informative training examples \citep{mirzasoleiman2020coresets,paul2021deep}, while dataset condensation synthesizes compact training sets that approximate the learning signal of larger datasets \citep{zhao2020dataset,cazenavette2022dataset}.
BestRand applies this data-efficiency perspective to model evaluation by selecting a reusable subset of existing test items that closely preserves full-benchmark results at low construction cost.
By reducing the number of evaluated items, BestRand complements methods that reduce the cost of processing each item.

\subsection{Benchmark Compression Methods}
\label{app:benchmark-compression-methods}

Benchmark compression reduces the number of evaluated items while preserving full-benchmark scores or model rankings. \citep{ling2026testingmuchsurveybenchmark, lin2026holocosmsinglemetricbenchmarkcompression}
For constructing reusable subsets, a useful distinction is whether methods use historical item-level model responses.
Methods without such records use random sampling or select representative items through prompt embeddings and other content-derived features, typically using clustering or diversity-based selection \citep{song2025less,bean2025scales++}.
Response-based methods instead exploit evaluation results from previously tested models to identify informative items through response-pattern clustering, item response modeling, or learned score prediction \citep{vivek2024anchor,polo2024tinybenchmarks,zhang2026sparseeval}.
Many such methods rely on large collections of model evaluations, making subset construction expensive for new benchmarks without public evaluation logs \citep{bean2025scales++, huang2026zipbenchlowcostframeworkcompressing}.
Efforts to reduce this dependence include learning from small anchor sets, augmenting observed responses with synthetic records, and calibrating random subset sizes using a few models \citep{saranathan2025sublime,huang2026zipbenchlowcostframeworkcompressing,das2026mince}.

Beyond method development, prior analyses examine benchmark redundancy \citep{zhang2025redundancy,sha2026benchscope} and the statistical reliability of model comparisons \citep{miller2024adding,heineman2026signal}.
Studies of compressed evaluation further highlight sensitivity to calibration data and model shifts, the limitations of very small subsets, and the competitiveness of random sampling \citep{zhang2026benchmark,yauney2026reliable}.
Our theory complements these perspectives by characterizing the distribution of compression errors, explaining persistent high-error risk and providing a principled basis for reducing it.
BestRand offers improved accuracy and stability over random sampling with lower construction cost than methods that rely on extensive historical evaluation records.
Detailed descriptions of the baselines considered in our experiments are provided in Appendix~\ref{app:baseline-details}.

\addtocontents{toc}{\protect\setcounter{tocdepth}{1}}
\section{The Novelty and Contribution of BestRand}
\label{app:novelty-contribution}

\subsection{Novelty of BestRand}
\label{app:novelty-of-bestrand}

Our novelty follows a clear progression. We first derive a theoretical characterization of benchmark compression and explain its behavior, then use these insights to improve its reliability and stability.

\paragraph{An accurate characterization of the error distribution in benchmark compression.}
We derive an accurate analytical approximation to the MAE distribution when a benchmark subset is used to estimate full-benchmark scores. This distribution-level view goes beyond average error and ranking fidelity by describing both typical outcomes and high-error outcomes. It therefore provides a principled basis for understanding the accuracy and reliability of benchmark compression. Numerical results across LLM, VLM, and VLA benchmarks closely match the predicted distributions.

\paragraph{An explanation of why random sampling is almost enough.}
Our analysis shows that the MAE distribution has a dominant low-error region. As the number of evaluated models grows, its mode approaches the lower endpoint of the limiting distribution at a universal rate. Together with the observed concentration around this region, this result explains why random sampling produces accurate compressed benchmarks with high probability and is often a strong baseline. The locations of the empirical modes are consistent with the theoretical prediction.

\paragraph{An explanation of the persistent instability of random sampling.}
We identify the common shift, caused by a subset being easier or harder than the full benchmark, as a key source of the upper error tail. Its scale grows with behavioral correlation across models and remains positive as model count grows under positive correlation. Thus, evaluating more models cannot average away shared errors. This persistent instability reflects correlated behavior among foundation models, consistent with relationships among model correlation, common-shift scale, and high-error risk.

\paragraph{A theoretically grounded, stable, and low-cost method.}
This analysis motivates BestRand. We decompose the common shift into probe-explained and residual components. BestRand draws multiple random subsets and selects the one whose mean probe scores best match a pool reference to control the explained component. Under our selection model, theory guarantees that the common-shift variance converges to a value strictly below that of random sampling as the probe imbalance vanishes, thereby improving stability. Experiments consistently confirm this theoretical prediction.

\subsection{Contributions of BestRand}
\label{app:contributions-of-bestrand}
Together, these theoretical insights and methodological advances yield three main contributions.

\paragraph{A theoretical framework for benchmark compression.}
We derive exact sampling moments and an accurate analytical approximation to the MAE distribution. The framework connects compression error and high-error risk to the subset size, item difficulty variation, and model correlation. The theory further shows that matching probe-model scores can reduce common-shift variance. Extensive numerical validation shows that the predicted distributions, modes, and tail-related quantities closely match empirical measurements, supporting both the validity and practical value of the theory.

\paragraph{A stable and low-cost compression method.}
We introduce BestRand, which uses a few probe models to select from random subsets. These probe models evaluate only a small subset of benchmark items rather than the full benchmark, so construction cost is low. Our theory shows that sufficiently accurate probe matching gives BestRand lower common-shift variance than Random and thus greater stability. Experiments consistently confirm this stability gain.

\paragraph{Extensive empirical validation and strong results.}
We evaluate BestRand on 70 text, multimodal, and agent benchmarks across various retention ratios, and further extend it to VLA evaluation. The empirical error distributions closely match our theoretical analysis, and BestRand consistently suppresses the high-error region predicted by the theory. It improves estimation accuracy at low construction cost while maintaining strong ranking fidelity. These gains remain consistent under score-range, model-family, and temporal shifts, supporting its robustness to unseen target models.

\addtocontents{toc}{\protect\setcounter{tocdepth}{2}}
\end{document}